\documentclass{article}

\usepackage[final]{neurips_2023}

\usepackage[utf8]{inputenc} 
\usepackage[T1]{fontenc}    
\usepackage{hyperref}       
\usepackage{url}            
\usepackage{booktabs}       
\usepackage{amsfonts}       
\usepackage{nicefrac}       
\usepackage{microtype}      
\usepackage{xcolor}         

\usepackage[pdftex]{graphicx}
\usepackage{rotating}
\usepackage{caption}
\usepackage{subcaption}
\usepackage{bm}
\newcommand{\BR}{\mathbb{R}}

\usepackage{amsmath}
\usepackage{amssymb}
\usepackage{mathtools}
\usepackage{amsthm}

\theoremstyle{plain}
\newtheorem{theorem}{Theorem}[section]

\newtheorem{corollary}[theorem]{Corollary}
\theoremstyle{definition}
\newtheorem{definition}[theorem]{Definition}

\theoremstyle{remark}

\DeclareMathOperator{\IG}{IG}

\title{Understanding the Vulnerability of CLIP to Image Compression}

\author{%
  Cangxiong Chen
  \\
  Institute for Mathematical Innovation\\
  University of Bath\\
  \texttt{cc2458@bath.ac.uk} \\
  \And
  Vinay P.~Namboodiri \\
  Department of Computer Science \\
  University of Bath \\
  \texttt{vpn22@bath.ac.uk} \\
  \AND
  Julian Padget \\
  Department of Computer Science \\
  University of Bath \\
  \texttt{J.A.Padget@bath.ac.uk} \\
}

\begin{document}

\maketitle

\begin{abstract}

CLIP is a widely used foundational vision-language model that is used for zero-shot image recognition and other image-text alignment tasks. We demonstrate that CLIP is vulnerable to change in image quality under compression. This surprising result is further analysed using an attribution method-Integrated Gradients. Using this attribution method, we are able to better understand both quantitatively and qualitatively exactly the nature in which the compression affects the zero-shot recognition accuracy of this model. We evaluate this extensively on CIFAR-10 and STL-10. Our work provides the basis to understand this vulnerability of CLIP and can help us develop more effective methods to improve the robustness of CLIP and other vision-language models.
\end{abstract}

\section{Introduction}
CLIP (Contrastive Language-Image Pretraining) \citep{radford21LearningTransferable} is a foundation model that is trained to find the best pairing between given images and texts using contrastive learning. Benefitting from the variety and quantity of training data, CLIP can be used in many applications. We are interested in using CLIP as a zero-shot image classifier with a given fixed text prompt. The work \citet{radford21LearningTransferable} has claimed through extensive experiments with datasets with different natural distribution shift that zero-shot CLIP is much more robust to distribution shift than standard ImageNet models (for example, natural images, sketches and renditions of objects of a particular class). However, we find that CLIP's zero-shot prediction is sensitive to the quality of the input images. For example, the predicted text label for the same image can differ significantly when the image has been compressed using the discrete cosine transform (see Figures \ref{fig:clip prec cifar10 test}). This is surprising because CLIP has been trained on over 400 million image-text pairs with images of various qualities and we would therefore expect it to be robust against degradation of the quality of the input images. To better understand the vulnerability of CLIP to image compression, we use Integrated Gradients \citep{sundararajan17aAxiomaticAttribution} on CLIP with input images of different qualities to probe how the change of quality affects predictions. We found that the method of Integrated Gradients provides an effective way to quantify the source of impact on CLIP's predictions at the pixel level of the input. Furthermore, the method can be applied to any foundation model subject to the assumption that the model defines a function that is differentiable almost everywhere. Unlike most existing attribution methods, Integrated Gradients satisfies 
\textit{sensitivity} and \textit{implementation invariance} which are two axioms characterisable mathematically.  We believe this work can help us monitor the distribution shift of the dataset and the model. We also hope it can inspire the community to develop mitigation strategies to improve the robustness of CLIP and other foundational models.

\paragraph{Our contributions}
\begin{enumerate}
    \item We demonstrate through CIFAR-10 \citep{krizhevsky2009learning}\citep{cifar10license} and STL-10 \citep{Coates11Analysis} that CLIP is sensitive to the quality of the input image when it performs a zero-shot image recognition task with fixed text prompts. 
    \item We investigate the above behaviour through an attribution method of Integrated Gradients and provide numerical estimates and visualisations to explain how change in the qualities of the inputs affects the predictions.  
\end{enumerate}
In section \ref{CLIP_summary}, we give a brief summary of CLIP. We summarise the Integrated Gradients method in our notation in Section \ref{sec:integrated gradients}. In Section \ref{sec:probing CLIP}, we present our method probing CLIP using Integrated Gradients and its mathematical formulation. In Section \ref{sec:experiments}, we evaluate the robustness of CLIP on CIFAR-10 and STL-10 and demonstrate our method using CIFAR-10. Section \ref{sec:related work} reviews related works and we conclude in Section \ref{sec:conclusions} with suggestions for future work.

\section{CLIP}\label{CLIP_summary}
CLIP is a learning method from natural language supervision that predicts the best text description that pairs with any given image. At training time, CLIP jointly trains an image encoder and a text encoder to predict the correct image-text pairs. At test time, CLIP can be used as a zero-shot image classifier by predicting which text label goes best with the given image. This can be done by providing CLIP with a text prompt similar to ``This is an image of \{CLS\}'' with ``CLS'' being the labels. Then for each given image, we use CLIP to compute the dot product between the encoding of the image and encoding of the prompt with each label. Finally, we choose the label that maximises the dot product. Using CLIP for zero-shot image recognition has also been investigated in works such as \citet{radford21LearningTransferable}, \citet{zhou2022coop} and \citet{wu2023WhyIsPromptTuning}. As described by \citet{radford21LearningTransferable}, the model has been trained with a large collection of around 400 million image-text pairs. This would imply that the method should have been trained with a variety of different image compression ratios and should naturally be robust to any issues related to compression. However, as we go on to show, the model is vulnerable to image compression ratio. We next consider an attribution method that can be used to thoroughly analyse this phenomenon.

\section{Integrated Gradients}\label{sec:integrated gradients}
Integrated Gradients \citep{sundararajan17aAxiomaticAttribution} is an attribution method that analyses what features can (and how they) affect the prediction of a deep network. The Integrated Gradients method satisfies two axioms `sensitivity' and `implementation invariance' and can be used for most deep learning models. `Sensitivity' refers to the property that when the outputs of the network are different at two features, the attributes should also be different. `Implementation invariance' means that the attributes are the same for two functionally equivalent networks, that is networks having the same outputs given the same inputs. Integrated gradients satisfy these axioms because it is defined as a path integral of the deep network (as a function) from the baseline input to the target. The axioms are guaranteed due to the independence of parameterisations of path integrals from the fundamental theorem of Calculus. To facilitate our discussion in Section \ref{sec:probing CLIP}, we re-state the definition of integrated gradients and its properties in the general context. 

Let $F \colon \BR^n \rightarrow \BR^m$ be a neural network. Let $\bm{x}_0, \bm{x}_1$ be two inputs in $\BR^n$ which do not have to be distinct. Let $\gamma \colon [0,1] \rightarrow \BR^n$ be a continuously differentiable path connecting $\bm{x}_0$ and $\bm{x}_1$, so that $\gamma(0) = \bm{x}_0$, $\gamma(1) = \bm{x}_1$ and $\gamma^\prime (t)$ is continuous for all $t \in (0,1)$.
\begin{definition}\label{def:IG}
Assuming $F$ is differentiable almost everywhere, we define the integrated gradients as the path integral of $F$ from $\bm{x}_0$ to $\bm{x}_1$:
\begin{equation}
    \IG(F,\bm{x}_0,\bm{x}_1) := \int_{0}^{1}\frac{\partial F(\gamma(t))}{\partial t}dt. 
\end{equation}
\end{definition}
From the fundamental theorem of Calculus, we know the following fact as a result of Definition \ref{def:IG}:
\begin{corollary}
Under the same assumptions in Definition \ref{def:IG}, we have 
\begin{equation}\label{eq:FTC}
    \IG(F,\bm{x}_0,\bm{x}_1) = F(\bm{x}_1) - F(\bm{x}_0).
\end{equation}
\end{corollary}
This shows that the integral in \eqref{def:IG} is in fact path independent, justifying that there is no $\gamma$ in the notation $\IG(F,\bm{x}_0,\bm{x}_1)$. In the context of attribution analysis, integrated gradients defines an attribute that satisfies sensitivity \citep[2.1]{sundararajan17aAxiomaticAttribution} which essentially says for all $\bm{x}_0,\bm{x}_1$ such that $F(\bm{x}_1) \neq F(\bm{x}_0)$, we should have $\IG(F,\bm{x}_0,\bm{x}_1) \neq 0$. Note that this is guaranteed by \eqref{eq:FTC}. This also explains why Integrated Gradients satisfies implementation invariance.

\section{Probing CLIP using Integrated Gradients}\label{sec:probing CLIP}
CLIP is built on transformers and ResNets as image and text encoders with activations that are differentiable almost everywhere. It therefore is appropriate to apply attribution analysis using integrated gradients, as we now show.  
Following the same notation as Section \ref{sec:integrated gradients}, we denote the CLIP model as $F \colon (\bm{x},\bm{z}) \mapsto F(\bm{x},\bm{z}) \in \BR^C$, where $\bm{x},\bm{z}$ are the image and text inputs respectively and $C$ is the total number of classes. Since our goal is to investigate the impact of image quality on CLIP, we fix the text input and its representation given by the text encoder. In other words, we consider $F(\bm{x}|\bm{z})$. In subsequent discussion, we will write $F(\bm{x})$ instead of $F(\bm{x}|\bm{z})$ unless specified otherwise. 

We consider integrated gradients $\IG[l(F(\bm{x}),k_{\bm{x}}),\bm{x}_0,\bm{x}_1]$ of the cross-entropy loss $l(F(\bm{x}), k_{\bm{x}})$ with respect to the baseline $\bm{x}_0$ and the target $\bm{x}_1$ where 
\begin{equation}
l(F(\bm{x}), k_{\bm{x}}) = - \log \frac{e^{f_{k_{\bm{x}}}(\bm{x})}}{\sum_j e^{f_{j}(\bm{x})}}.
\end{equation}
and $k_{\bm{x}}$ is the true label of $\bm{x}$. Here we write $F(\bm{x}) = (f_j (\bm{x})), 1 \leq j \leq C, f_j (\bm{x}) \in \BR$. 

In order to compute the integral $\IG[l(F(\bm{x}),k_{\bm{x}}),\bm{x}_0,\bm{x}_1]$ numerically, we make further simplifications. Following \citet{sundararajan17aAxiomaticAttribution}, we take $\gamma$ to be the line path and apply the trapezoidal approximation to the integral: 
\begin{equation}\label{eq:approx the integral}
    \IG[l(F(\bm{x}),k_{\bm{x}}),\bm{x}_0,\bm{x}_1] \approx 
    \sum_{i=1}^{n} \frac{x_0^{(i)} - x_1^{(i)}}{m} \sum_{s=1} ^N \frac{\partial l(\bm{x})}{\partial x^{(i)}} \Bigg \vert_{x^{(i)} = x_0^{(i)} + \frac{s}{N}\left (x_1^{(i)} - x_0^{(i)} \right )},
\end{equation}
where $N$ is the number of steps in the approximation and the notations $x^{(i)}, x_0^{(i)}, x_1^{(i)}$ denote the $i$-th component of the vector $\bm{x}, \bm{x}_0, \bm{x}_1$ respectively.

\section{Experiments}\label{sec:experiments}
\subsection{CLIP is sensitive to the quality of the input image}\label{sec:sensitive to the quality of image}
\begin{figure*}[!ht]
  \centering
  \begin{minipage}{\textwidth}
    \centering
    \scalebox{1}{
    \begin{tabular}{lccccc}
    \toprule
        Image encoder & Original & Quality 75 & Quality 50 & Quality 25  \\
         \midrule
            ResNet50 & 0.7141 & 0.5457 & 0.4689 & 0.3562 \\ 
                ResNet101 & 0.7934 & 0.6179 & 0.4945 & 0.3441 \\ 
                ResNet50x4 & 0.7674 & 0.6007 & 0.4948 & 0.3524 \\ 
                ResNet50x16 & 0.8125 & 0.6632 & 0.5607 & 0.4050 \\
                ResNet50x64 & 0.8346 & 0.6782 & 0.5870 & 0.4535 \\
              \midrule
                ViT-B/32 & 0.8831 & 0.7214 & 0.6162 & 0.4763 \\
                ViT-B/16 & 0.9052 & 0.7696 & 0.6574 & 0.4854 \\
                ViT-L/14 & 0.9538 & 0.8600 & 0.7735 & 0.6129 \\
          ViT-L/14@336px & 0.9493 & 0.8466 & 0.7571 & 0.5990 \\
         \bottomrule
    \end{tabular}
    }
  \end{minipage}
  
  \begin{minipage}{\textwidth}
    \centering
    \includegraphics[width=\linewidth]{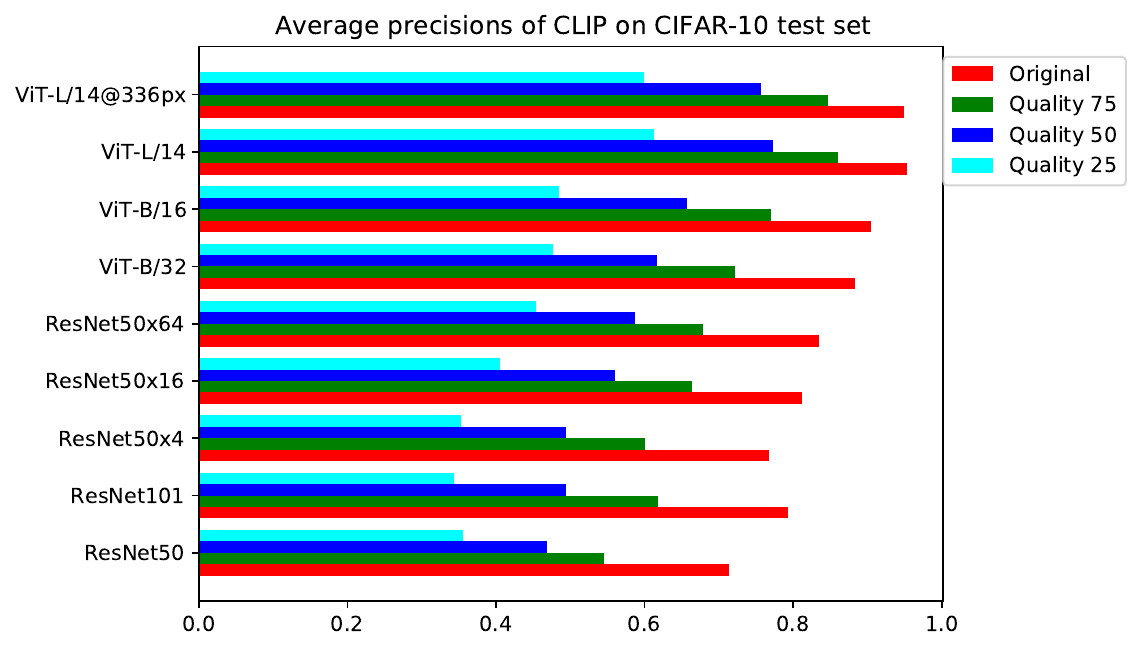}
    \label{fig:plot clip prec cifar10 test}
  \end{minipage}%
  \caption{Average precision of CLIP predictions over the test dataset from CIFAR-10 across different image qualities. The number ``*'' in ``Quality *'' refers to the setting of the ``quality'' parameter in PIL's ``Image.save'' function. The larger the value, the better the quality (or equivalent less compression) is. Precision scores decrease when the image quality degrades whichever image encoder we use in CLIP. The plot is a visualisation of the table.}
  \label{fig:clip prec cifar10 test}
\end{figure*}

\begin{figure*}[!ht]
  \centering
  \begin{minipage}{\textwidth}
    \centering
    \scalebox{1}{
    \begin{tabular}{lccccc}
    \toprule
        Image encoder & Original & Quality 75 & Quality 50 & Quality 25  \\
         \midrule
            ResNet50 & 0.9479 & 0.9203 & 0.8826 & 0.8143 \\

                ResNet101 & 0.9624 & 0.9413 & 0.9180 & 0.8575 \\
                
                ResNet50x4 & 0.9623 & 0.9395 & 0.9144 & 0.8582 \\
                
                ResNet50x16 & 0.9755 & 0.9555 & 0.9346 & 0.8902 \\
                
                ResNet50x64 & 0.9851 & 0.9686 & 0.9547 & 0.9250 \\
                \midrule
                ViT-B/32 & 0.9719 & 0.9573 & 0.9340 & 0.8798 \\
                
                ViT-B/16 & 0.9832 & 0.9712 & 0.9537 & 0.9048 \\
                
                ViT-L/14 & 0.9924 & 0.9880 & 0.9800 & 0.9585 \\

                ViT-L/14@336px & 0.9925 & 0.9874 & 0.9765 & 0.9487 \\
         \bottomrule
    \end{tabular}
    }
  \end{minipage}
  
  \begin{minipage}{\textwidth}
    \centering
    \includegraphics[width=\linewidth]{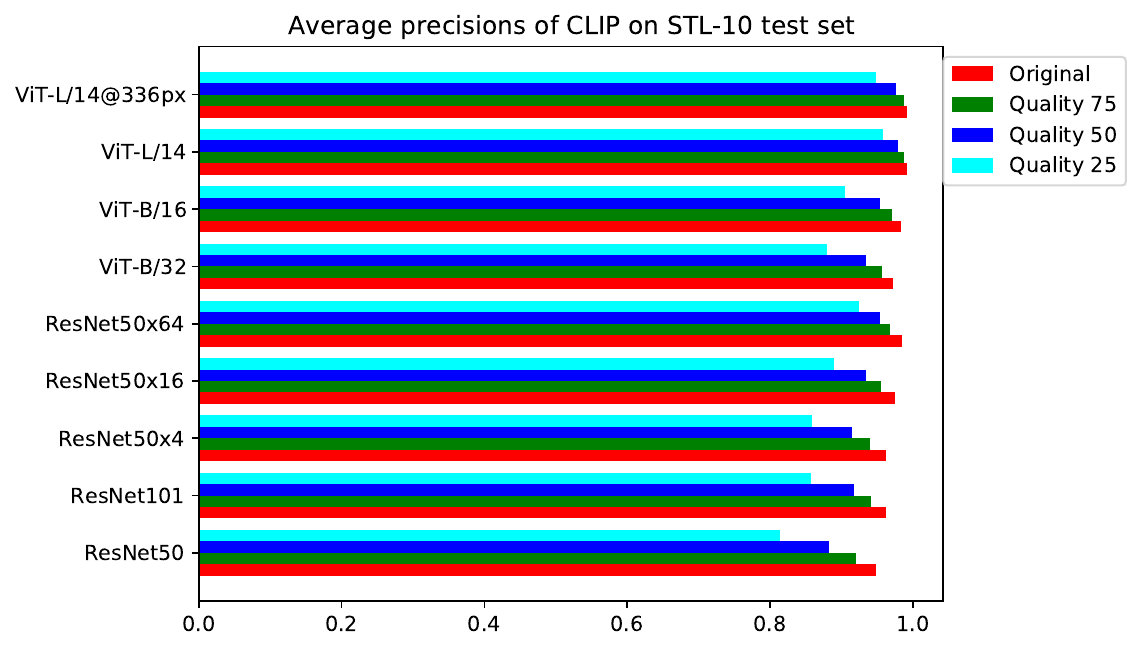}
    \label{fig:plot clip prec stl10 test}
  \end{minipage}%
  \caption{Average precision of CLIP predictions over the test dataset from STL-10 across different image qualities. The definitions of terms are provided in Figure \ref{fig:clip prec cifar10 test}. We observe a decrease in precision scores for each image encoder, although the relative drops of scores are much smaller. The plot is a visualisation of the table.}
  \label{fig:clip prec stl10 test}
\end{figure*}

We demonstrate that CLIP is sensitive to the quality of the input image when we use it for image classifications on CIFAR-10 and STL-10. As we run the same test on each dataset, we will only explain the testing procedure using CIFAR-10. We create four groups of the CIFAR-10 test dataset: the first one with the original quality; the second to the fourth have degraded quality given by JPEG compression implemented by the ``Image.save'' function in Python PIL library \citep{pillow}. As the quality factor is reduced, the JPEG algorithm aggressively increases the quantization and the image detail is lost. At the highest compression ratios one observes blocky artefacts. Each of the groups have 10k images. To allow CLIP to classify an image, we provide the text prompt \textit{``This is an image of a \{CLS\}.''} where CLS is replaced by the CIFAR-10 classes in words: \textit{``airplane'', ``automobile'', ``bird'', ``cat'', ``deer'', ``dog'
', ``frog'', ``horse'', ``ship'', ``truck''}. We have chosen this prompt because we think it provides the minimum information needed for CLIP to do zero-shot image recognition. To evaluate the accuracy, we compute average precision over all 10 classes. We test CLIP with every pretrained image encoder provided in \citet{radford21LearningTransferable} and implemented by \citet{clipGitRepo}. The results are summarised in Figure \ref{fig:clip prec cifar10 test} and \ref{fig:clip prec stl10 test}. We can observe that in the CIFAR-10 test, the precision scores decrease significantly as the image quality degrades in each case of the image encoder. In the STL-10 test, we also observe a decrease in precision scores for all image encoders, although the amount of  decrease is much smaller. We provide evaluations on the training datasets from CIFAR-10 and STL-10 in the Appendix.

                
                
                
                
                

\subsection{Probing CLIP using Integrated Gradients}
In this experiment, we use Integrated Gradients to probe how changes in the image quality affect the predictions of CLIP. It is worth noting that our objective in this experiment is to demonstrate how we can probe the robustness of CLIP using attribution method given by Integrated Gradients, not a comprehensive evaluation of the robustness of CLIP on different perturbations on benchmark datasets such as provided by \citet{hendrycks2019Benchmarking}. 

We set the baseline to be the image with original quality. Note that CLIP only accepts input image size 224$\times$224$\times$3 so the CIFAR-10 images have been resized in the experiments using bicubic interpolation implemented by \texttt{torchvision.transforms} from PyTorch \citep{Paszke19PyTorch}. We compute integrated gradients between the baseline and the target image with various degrees of compressions using the approximation given by \eqref{eq:approx the integral} with number of steps $N=50$. We also visualise the impact of the changes by overlaying integrated gradients with the target images. To facilitate the visualisation, we compute the overlay to be a weighted average of the image and the integrated gradients. Following \citet{sundararajan17aAxiomaticAttribution}, we clip the integrated gradients into two ranges $[-1, 0]$ and $[0, 1]$ which are referred to as negative and positive polarity respectively. We plot gradients with negative polarity using the red channel and those with the positive one using the green channel. Thus, red at a pixel location means that the reduction in the quality increases loss at this location, and similarly green means reduction in loss at this location. The intensity of the colour is determined by the absolute value of the gradients. In summary, the scale and intensity of the colour shows how the difference between the target and the baseline changes the prediction of the model.

We experiment with CLIP with ResNet50 and ViT-B/32 image encoders to illustrate how the method works. The method can be applied to all image encoders shown in Figure \ref{fig:clip prec cifar10 test} but we keep to the above cases for sake of space. For each image encoder, we present two examples with different baselines and plot integrated gradients with negative, positive and both polarities. We also provide predicted labels, their scores and values of the integrated gradients in tables \ref{tbl:CIFAR-10 RN50 comparisons} and \ref{tbl:CIFAR-10 ViT comparisons}.  

\begin{figure*}[!ht]
		\centering
		%
		%
		\begin{subfigure}[h]{0.49\textwidth}
		\begin{minipage}[c]{0.05\linewidth}
		      \rotcaption{\parbox{1.75cm}{\fontsize{8pt}{8pt}\selectfont RN50 negative}}\label{fig:CIFAR10_IG_RN50_negative}  
		    \end{minipage}
		    \begin{minipage}[c]{0.85\linewidth}
		    \includegraphics[width=1.1\linewidth]{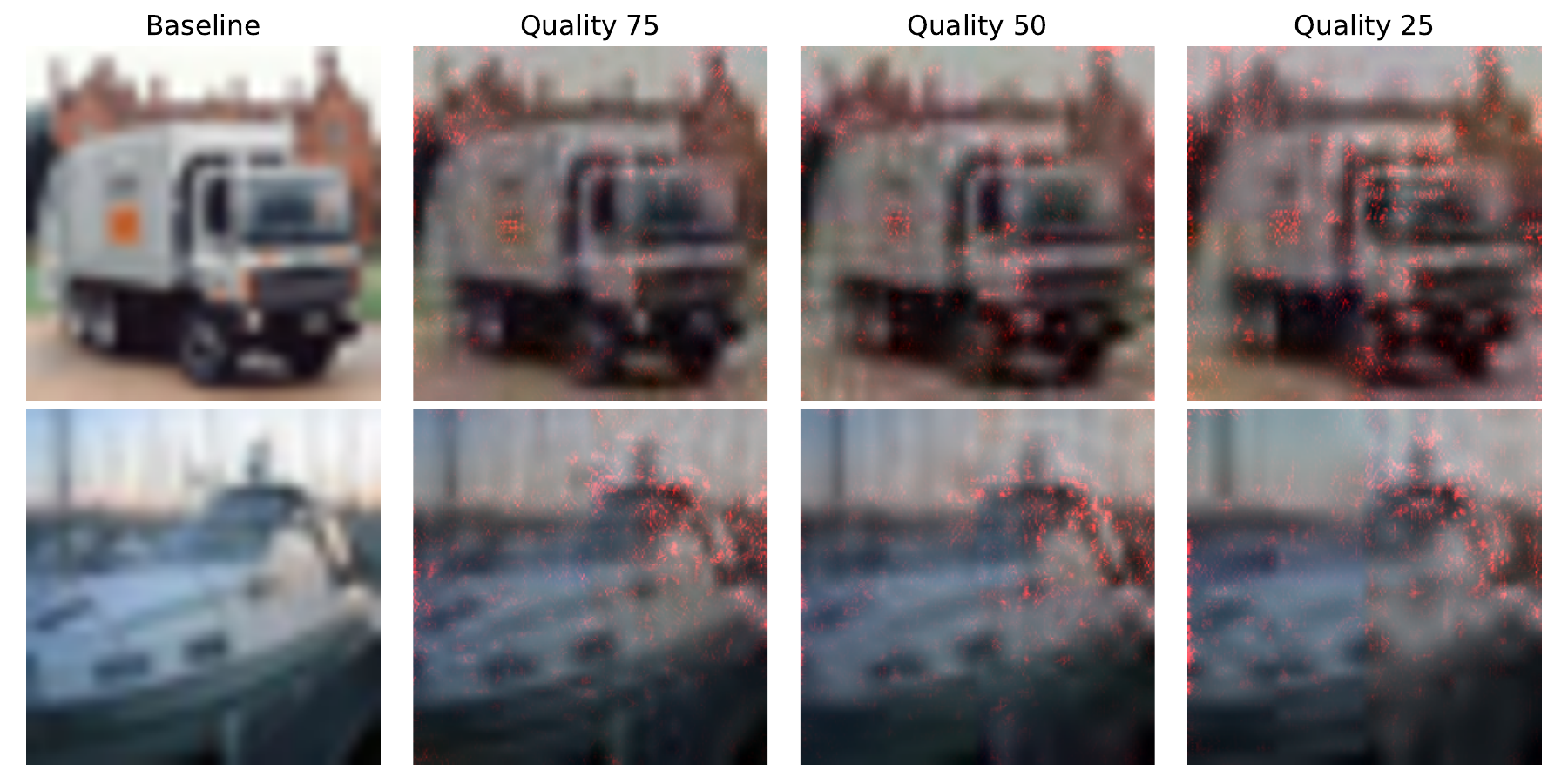}\\
		    \end{minipage}
		\end{subfigure}\hfill
		\begin{subfigure}[h]{0.49\textwidth}
		\begin{minipage}[c]{0.05\linewidth}
		      \rotcaption{\parbox{1.75cm}{\fontsize{8pt}{8pt}\selectfont RN50 positive}}\label{fig:CIFAR10_IG_RN50_positive}  
		    \end{minipage}
		    \begin{minipage}[c]{0.85\linewidth}
		    \includegraphics[width=1.1\linewidth]{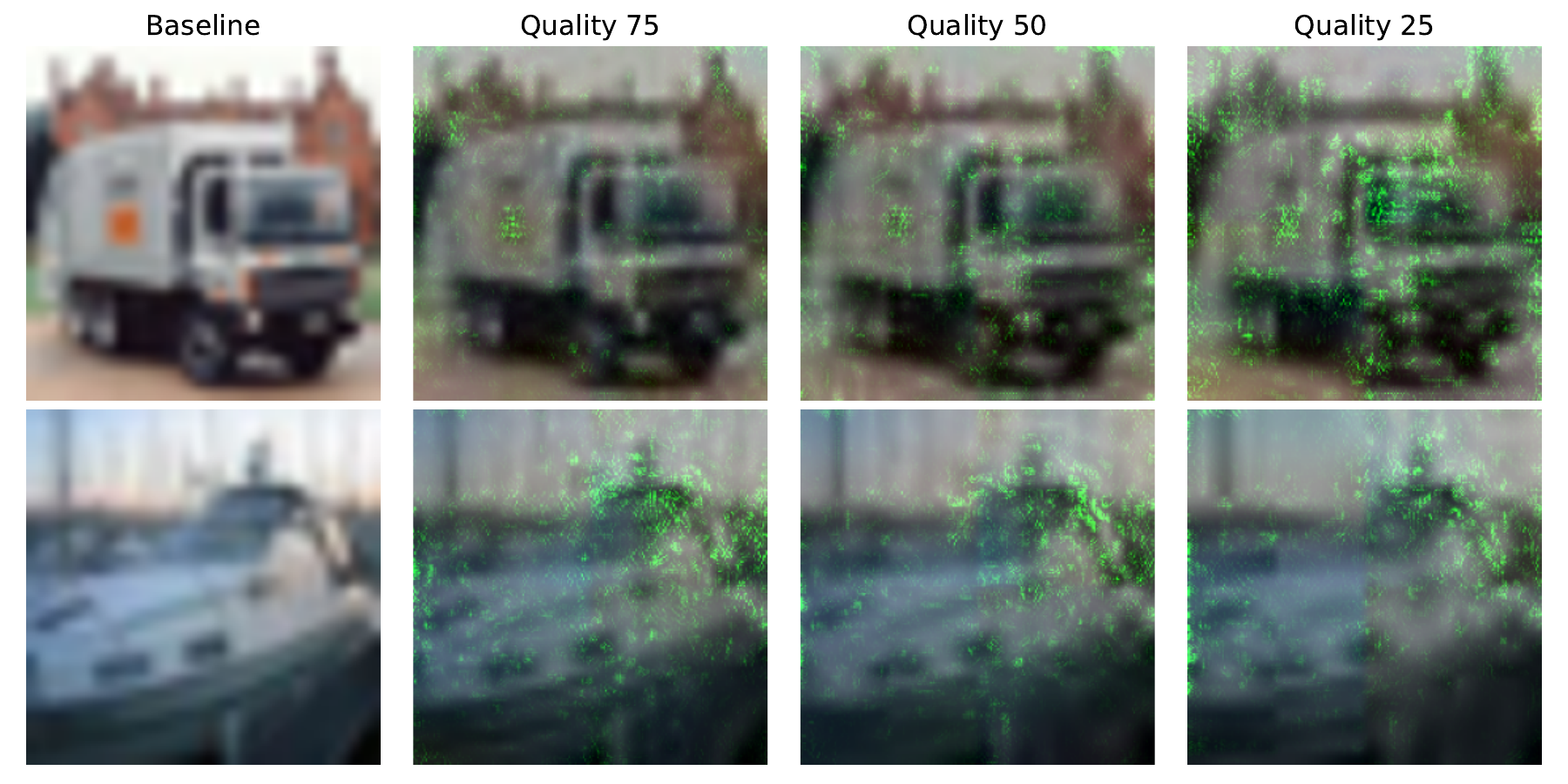}\\
		    \end{minipage}
		\end{subfigure}\hfill
		\begin{subfigure}[h]{0.49\textwidth}
		\begin{minipage}[c]{0.05\linewidth}
		      \rotcaption{\parbox{1.75cm}{\fontsize{8pt}{8pt}\selectfont RN50 both}}\label{fig:CIFAR10_IG_RN50_both}  
		    \end{minipage}
		    \begin{minipage}[c]{0.85\linewidth}
		    \includegraphics[width=1.1\linewidth]{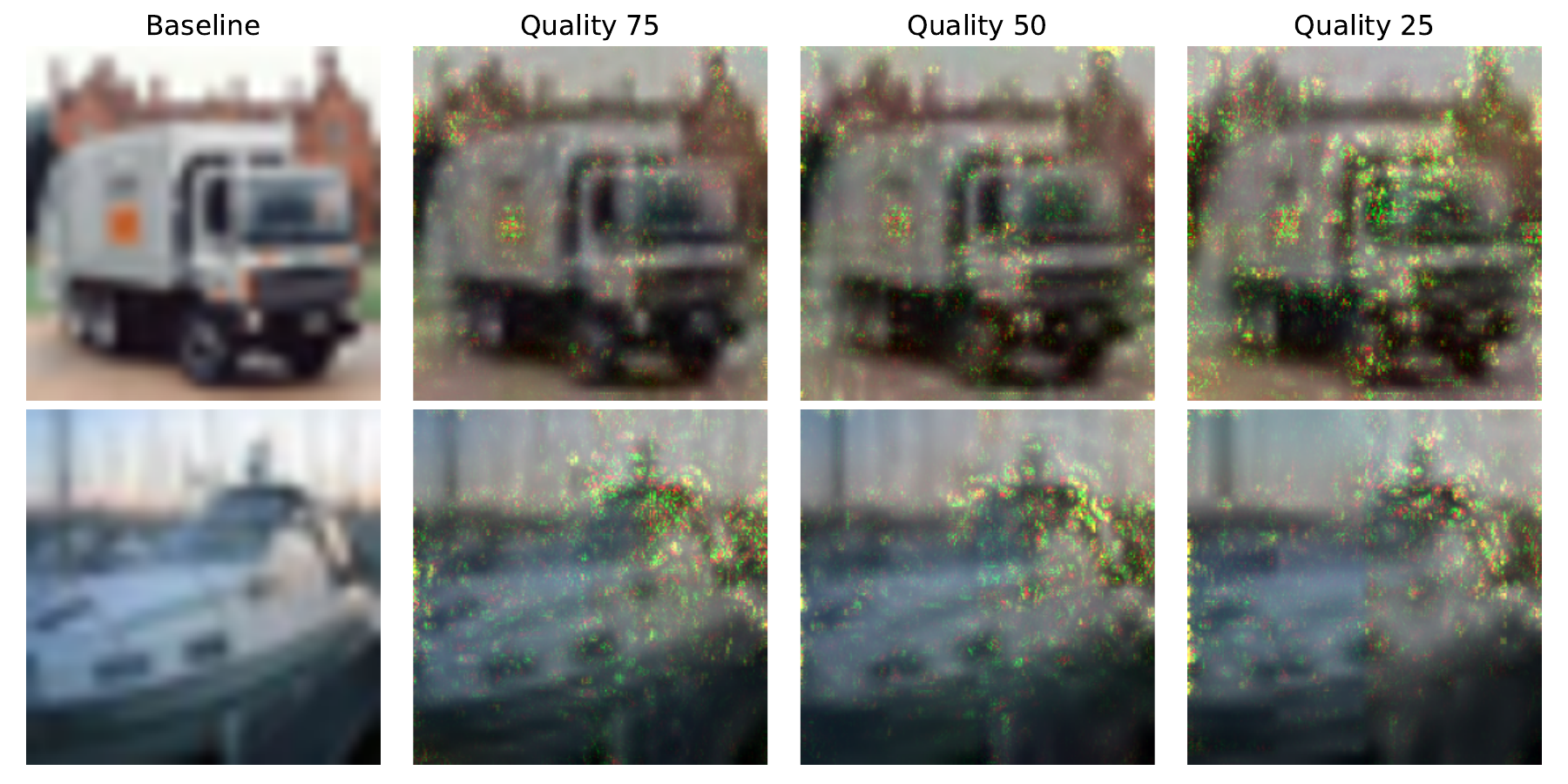}\\
		    \end{minipage}
		\end{subfigure} 
		\vspace{1mm}
		\caption{Visualisation of integrated gradients for CLIP with ResNet50 as the image encoder and different compressions of the baselines. We plot all three polarities of gradients, negative \ref{fig:CIFAR10_IG_RN50_negative}, positive \ref{fig:CIFAR10_IG_RN50_positive} and both \ref{fig:CIFAR10_IG_RN50_both}. For each the compressed images from quality 75 to 25, the overlay is computed as 0.7*image + 1.5*IG. Table \ref{tbl:CIFAR-10 RN50 comparisons} contains more details of the visual outputs.}
		\label{fig:IG plots RN50}
\end{figure*}

\begin{figure*}[!ht]
		\centering
		%
		%
		\begin{subfigure}[h]{0.49\textwidth}
		\begin{minipage}[c]{0.05\linewidth}
		      \rotcaption{\parbox{1.75cm}{\fontsize{8pt}{14pt}\selectfont ViT negative}}\label{fig:CIFAR10_IG_ViT_negative}  
		    \end{minipage}
		    \begin{minipage}[c]{0.85\linewidth}
		    \includegraphics[width=1.1\linewidth]{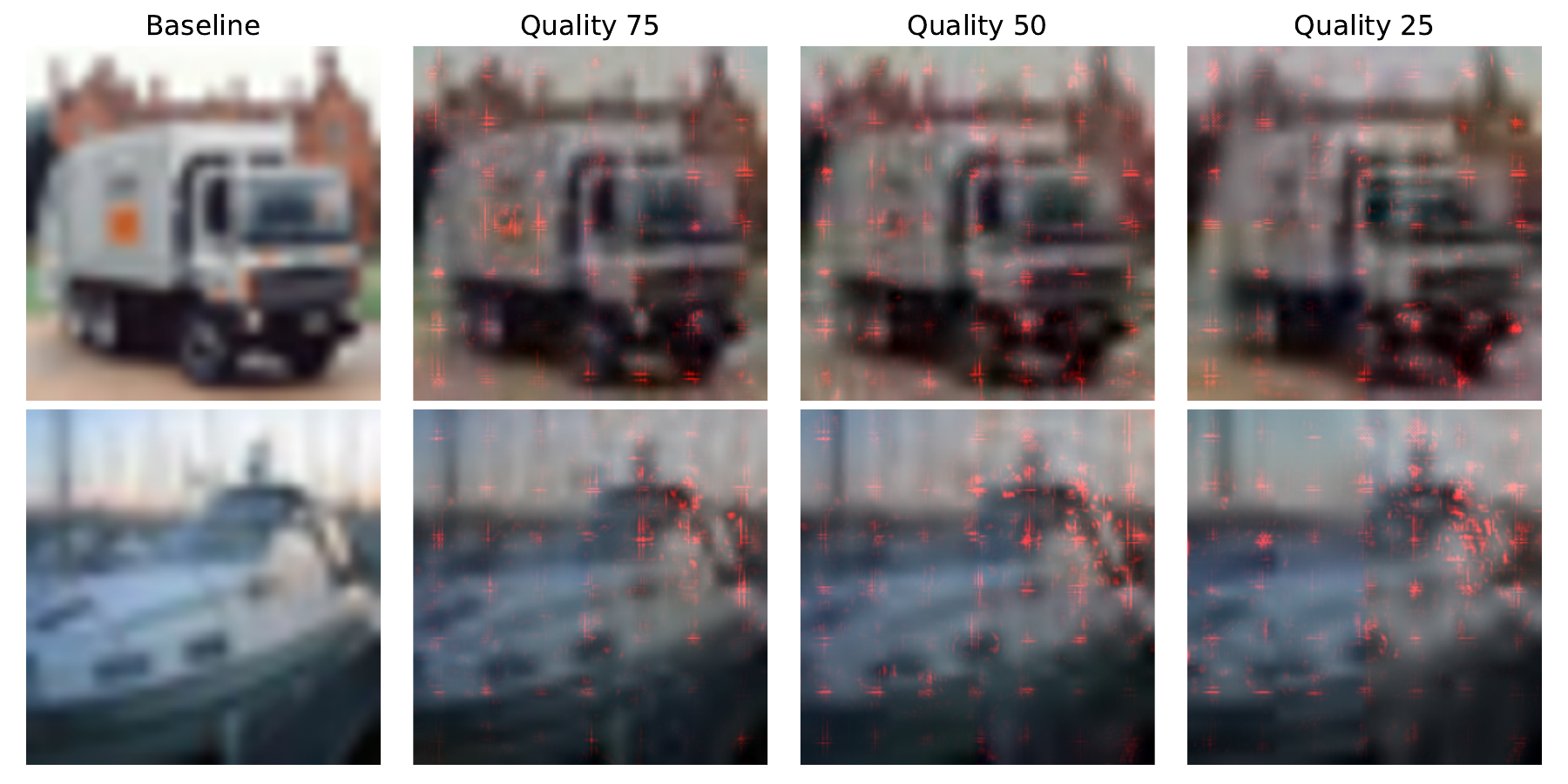}\\
		    \end{minipage}
		\end{subfigure}\hfill
		\begin{subfigure}[h]{0.49\textwidth}
		\begin{minipage}[c]{0.05\linewidth}
		      \rotcaption{\parbox{1.75cm}{\fontsize{8pt}{14pt}\selectfont ViT positive}}\label{fig:CIFAR10_IG_ViT_positive}  
		    \end{minipage}
		    \begin{minipage}[c]{0.85\linewidth}
		    \includegraphics[width=1.1\linewidth]{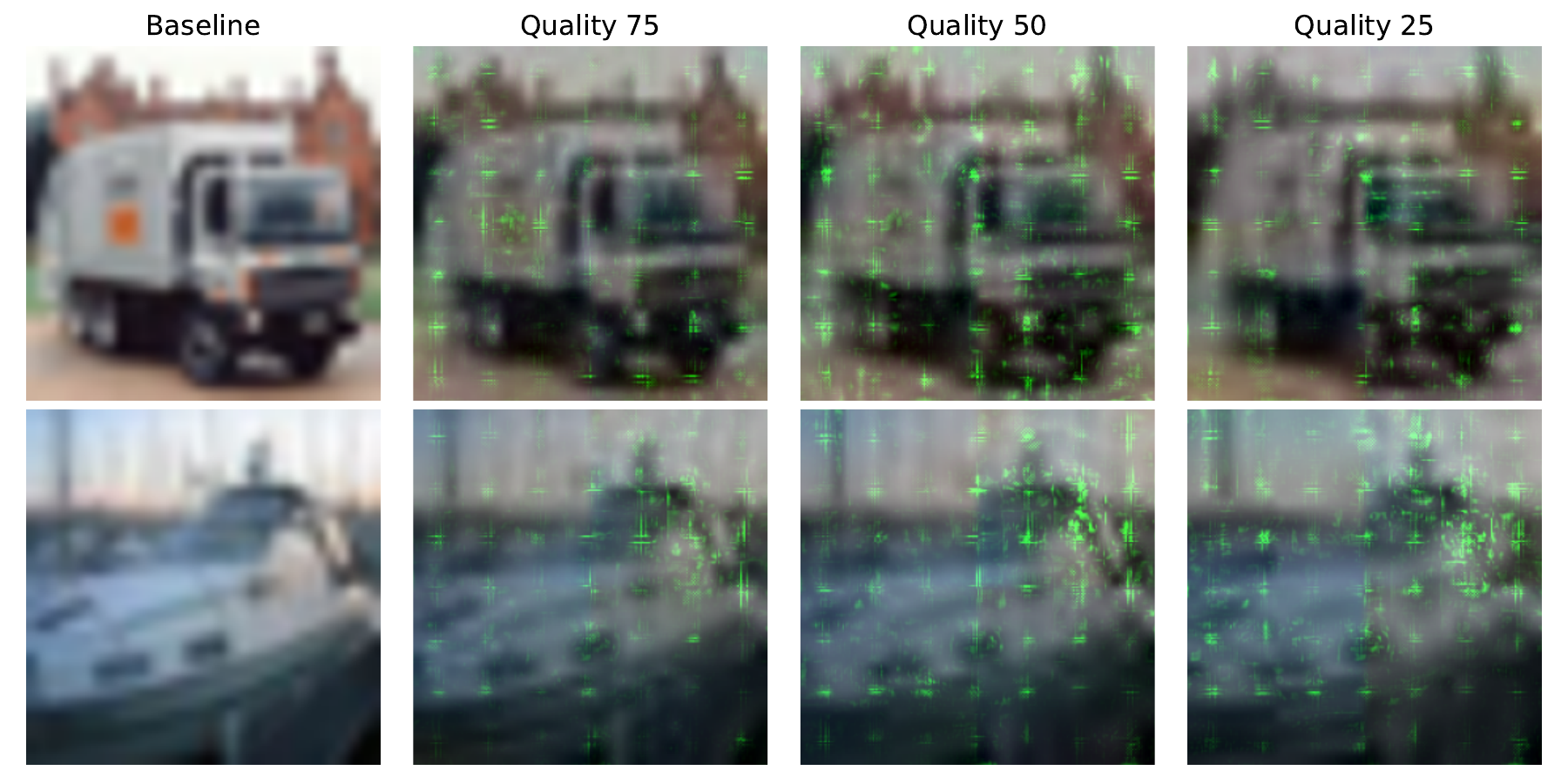}\\
		    \end{minipage}
		\end{subfigure}\hfill
		\begin{subfigure}[h]{0.49\textwidth}
		\begin{minipage}[c]{0.05\linewidth}
		      \rotcaption{\parbox{1.75cm}{\fontsize{8pt}{14pt}\selectfont ViT both}}\label{fig:CIFAR10_IG_ViT_both}  
		    \end{minipage}
		    \begin{minipage}[c]{0.85\linewidth}
		    \includegraphics[width=1.1\linewidth]{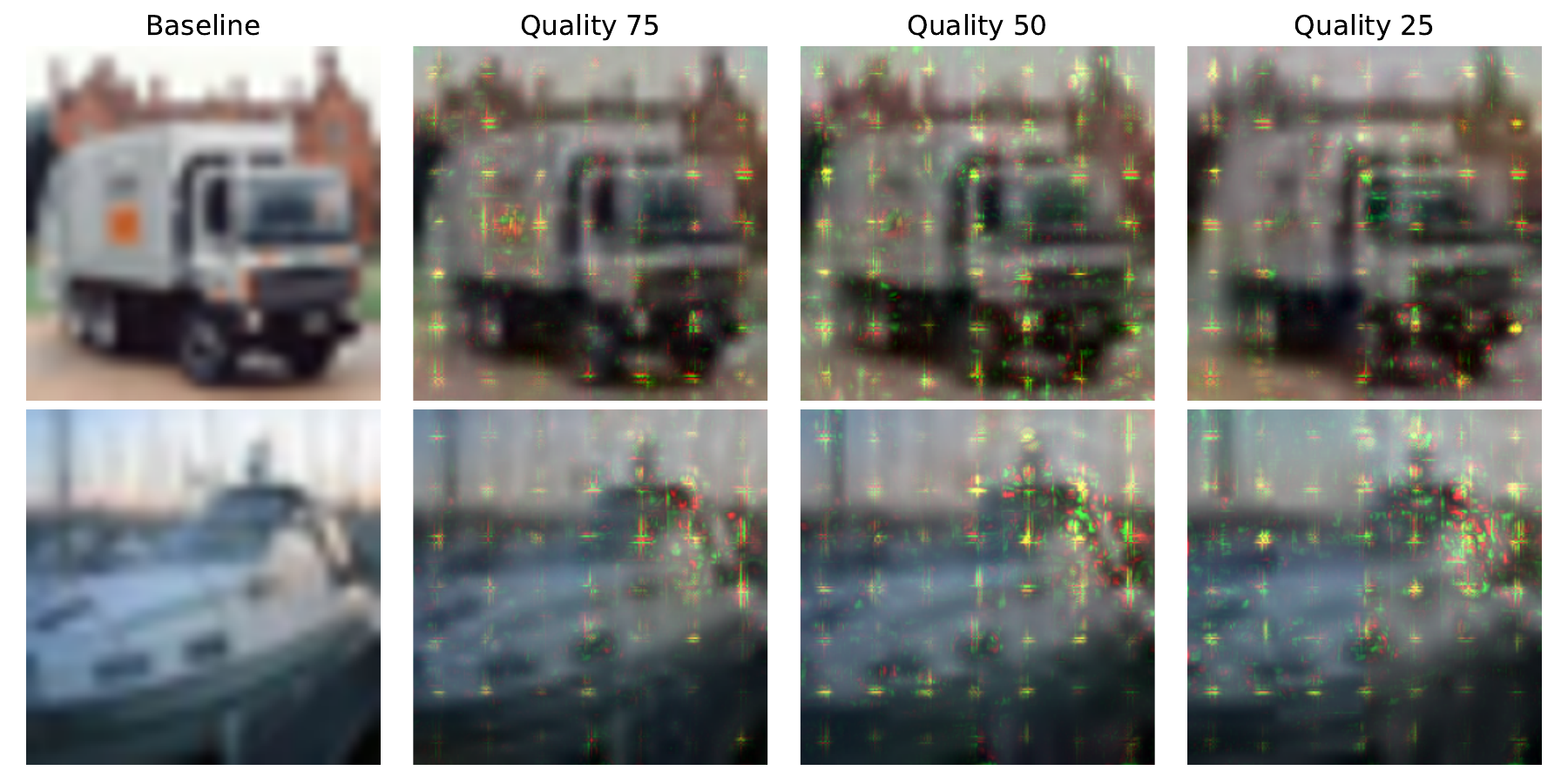}\\
		    \end{minipage}
		\end{subfigure} 
		\vspace{1mm}
		\caption{Visualisation of integrated gradients for CLIP with ViT-B/32 as the image encoder and different compressions of the baselines.}
		\label{fig:IG plots ViT}
\end{figure*}

By Definition \ref{def:IG}, the value of Integrated Gradients is equal to the difference in the loss function evaluated at the baseline and the target. The formula \eqref{eq:approx the integral} provides an approximation to the theoretic value of integrated gradients whose error is determined by the step size. We can observe from the values in Table \ref{tbl:CIFAR-10 RN50 comparisons} and \ref{tbl:CIFAR-10 ViT comparisons} that the integrated gradients provide accurate approximations to changes in the loss (which can be computed by taking difference of minus of the logarithm of the predicted scores). This shows integrated gradients serves as a good attribute for CLIP. The visualisations of integrated gradients in Figure \ref{fig:IG plots RN50} and \ref{fig:IG plots ViT} can help us probe the locations and the degree of impact on the predictions resulting from the changes in quality between the baselines and the targets. Another interesting observation we can make here is that our visualisations show the inductive bias brought by the two encoders. In the case of ResNet-50, we can see coloured areas typically concentrated in local regions. However, there is no such locality from ViT-B/32 and the coloured areas typically spread throughout the image in a grid pattern. We provide more visualised examples in the Appendix.
\begin{table*}[!ht]
		\centering
		\vspace{-4mm}
		\scalebox{0.85}{%
		\begin{tabular}{lccc}
		\toprule
		      True label & Predicted label & Predicted score & IG \\
		     \midrule
		     truck & automobile, airplane, airplane, airplane & 0.0900, 0.0401, 0.0371, 0.0300 & 0.8109, 0.8589, 1.0546 \\
              ship & airplane, airplane, airplane, airplane & 0.0925, 0.1322, 0.1549, 0.2499 & -0.3539, -0.5276, -0.9962 \\
		     \bottomrule
		\end{tabular}
		}
        \caption{Detailed information on the visual outputs in Figure \ref{fig:IG plots RN50}. The first row corresponds to that in Figure \ref{fig:IG plots RN50}, and the order in the labels and the numbers also follow that from the same figure. Predicted scores are the softmax of the logits of the model output at the index of the true label. The `IG' column contains values of the integrated gradients.}\label{tbl:CIFAR-10 RN50 comparisons}
\end{table*}
\begin{table*}[!ht]
		\centering
		\vspace{-4mm}
		\scalebox{0.85}{%
		\begin{tabular}{lccc}
		\toprule
		      True label & Predicted label & Predicted score & IG \\
		     \midrule
		     truck & truck, truck, automobile, ship & 0.0935, 0.7420, 0.3092, 0.0639 & 0.1936, 1.0680, 2.6468 \\
              ship & automobile, automobile, automobile, ship & 0.0773, 0.1745, 0.1396, 0.3113 & -0.8144, -0.5819, -1.3879 \\
		     \bottomrule
		\end{tabular}
		}
        \caption{Detailed information on the visual outputs in Figure \ref{fig:IG plots ViT}. Explanation of the table is given in the caption under Table \ref{tbl:CIFAR-10 RN50 comparisons}.}\label{tbl:CIFAR-10 ViT comparisons}
\end{table*}

\section{Related work}\label{sec:related work}
\paragraph{Robustness of Vision-Language models}
Much existing work on the robustness of deep learning models focusses on single-modality models. For example, \citet{BhojanapalliUnderstandingRobustness2021}, \citet{hendrycks2019Benchmarking}, \citet{Hendrycks21CriticalAnalysis} consider image recognition models. For language models, \citet{Wang22MeasureandImprove} provide a comprehensive survey of works in robustness of NLP models. 

\citep{Schiappa22RobustnessAnalysis} investigates robustness of video-language models for text-to-video retrieval. The authors use JPEG compression as a test for robustness, but they do not probe further into how compression changes the model behaviour. \citet{Yang21DefendingMultimodal} investigates adversarial robustness of Vision-Language models on any single modality. \citet{liang2021multibench} establishes a benchmark for evaluating robustness of multi-modal models but does not discuss how to probe the model to understand the cause of robustness issues.  

We have noticed that the classification accuracy of CLIP for zero-shot image classification is heavily influenced by the text prompt we provide surrounding the label. This has also been observed in \citet{zhou2022coop}. Prompt tuning has been shown to increase the robustness against variations in prompts (\citep{zhou2022coop}, \citep{wu2023WhyIsPromptTuning}). Since our focus in this work is robustness with respect to the image modality, we have fixed the prompt throughout our investigation. 
 
\paragraph{Attribution methods}
Compared with previous approaches (\citet{Shrikumar17LearningImportantFeatures}, \citet{Binder16LayerwiseRelevance}, \citet{Zeiler14Visualizing}, \citet{Springenberg15StrivingforSimplicity}), the Integrated Gradients method proposed in \citet{sundararajan17aAxiomaticAttribution} is the only attribution method satisfying both the axiom on sensitivity and on implementation invariance which are formulated in the same work. In a follow-up work, \citet{Hess21FastAxiomatic} improves the computational efficiency of the Integrated Gradients method but adds a strong assumption that the model must define a homogeneous function and thus the neural networks cannot have non-zero bias terms. We consider this too restrictive so we have followed the original method in \citet{sundararajan17aAxiomaticAttribution}.

To our best knowledge, our work is the first to use Integrated Gradients as an attribution method to probe CLIP under image-quality degradation due to compression.

\section{Conclusion}\label{sec:conclusions}
In this work, we demonstrate that CLIP is sensitive to image quality degradation from compression in the task of zero-shot image classification with a fixed text prompt. To help us understand the source of this vulnerability, we probe the model by using the attribution method of Integrated Gradients which offers us insight on how the change in the image quality can affect the value of the loss function and the prediction. We demonstrate how we can visualise attributes given by integrated gradients and their approximated values for quantitative comparisons. As for future work, we will investigate methods such as data augmentation to improve the robustness of CLIP with respect to input-image quality.
The code for this work will be made available.

\paragraph{Acknowledgements} Cangxiong Chen would like to thank the Institute for Mathematical Innovation for its support during preparation of this paper and for presenting it at the workshop. This work was partially supported by UKRI project \#10029108 (\href{https://gtr.ukri.org/projects?ref=10029108}{Novel AI video content moderation system}).

\bibliographystyle{abbrvnat}
\bibliography{reference}

\newpage
\appendix
\begin{center}
\section*{Understanding the Vulnerability of CLIP to Image Compression: Supplementary Material}    
\end{center}
\section{Evaluation of CLIP on the training datasets on CIFAR-10 and STL-10}
We provide more results on the vulnerability of CLIP using the training dataset from CIFAR-10 and STL-10. We can observe similar drops of precision scores for each image encoder as quality degrades, compared to figures \ref{fig:clip prec cifar10 test} and \ref{fig:clip prec stl10 test}.







\begin{figure*}[!ht]
  \centering
  \begin{minipage}{\textwidth}
    \centering
    \scalebox{1}{
    \begin{tabular}{lccccc}
    \toprule
        Image encoder & Original & Quality 75 & Quality 50 & Quality 25  \\
         \midrule
            ResNet50 & 0.7162 & 0.5531 & 0.4684 & 0.3582 \\

                ResNet101 & 0.7911 & 0.6180 & 0.4916 & 0.3422 \\

                ResNet50x4 & 0.7671 & 0.6069 & 0.4951 & 0.3610 \\

                ResNet50x16 & 0.8152 & 0.6705 & 0.5627 & 0.4014 \\

                ResNet50x64 & 0.8378 & 0.6842 & 0.5871 & 0.4473 \\
            \midrule
                ViT-B/32 & 0.8817 & 0.7220 & 0.6138 & 0.4762 \\

                ViT-B/16 & 0.9090 & 0.7686 & 0.6557 & 0.4809 \\

                ViT-L/14 & 0.9549 & 0.8599 & 0.7691 & 0.6172 \\

                ViT-L/14@336px & 0.9524 & 0.8477 & 0.7543 & 0.5987 \\
         \bottomrule
    \end{tabular}
    }
  \end{minipage}
  \hspace{1cm}
  \begin{minipage}{\textwidth}
    \centering
    \includegraphics[width=\linewidth]{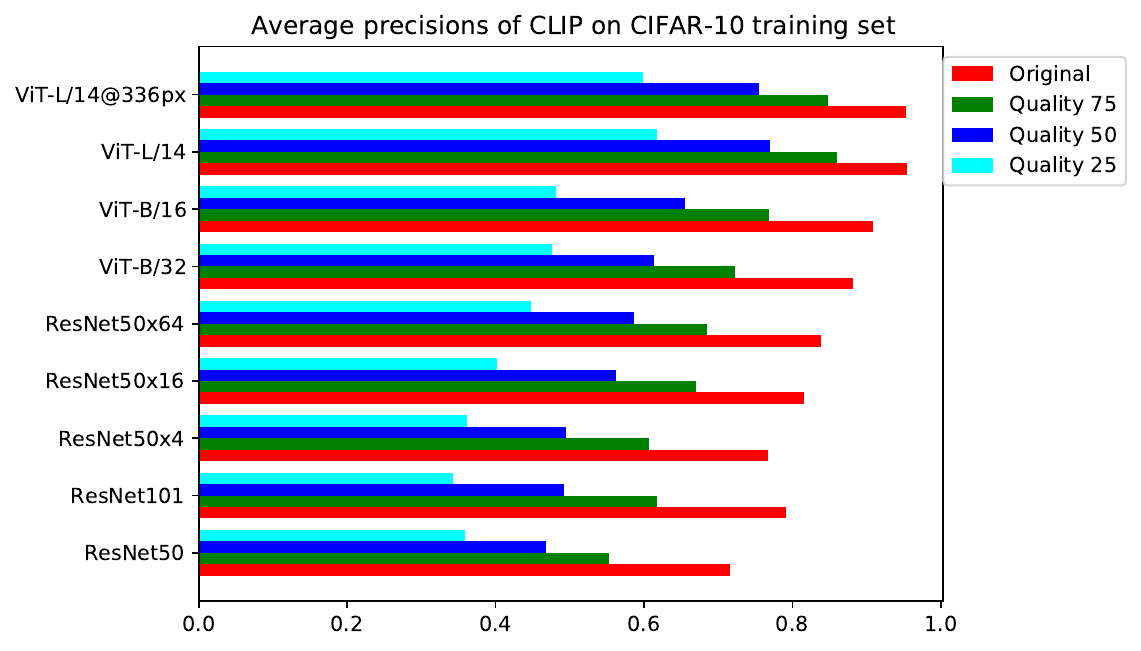}
    \label{fig:plot clip prec cifar10 train}
  \end{minipage}%
  \caption{Average precision of CLIP predictions over the training dataset from CIFAR-10 across different image qualities. The definitions of terms are provided in Figure \ref{fig:clip prec cifar10 test}.}\label{tbl:Supp CLIP on CIFAR-10 train}
\end{figure*}

                
                
                
                
                
                
\begin{figure*}[!ht]
  \centering
  \begin{minipage}{\textwidth}
    \centering
    \scalebox{1}{
    \begin{tabular}{lccccc}
    \toprule
        Image encoder & Original & Quality 75 & Quality 50 & Quality 25  \\
         \midrule
            ResNet50 & 0.9464 & 0.9132 & 0.8882 & 0.8130 \\

                ResNet101 & 0.9578 & 0.9436 & 0.9224 & 0.8640 \\
                
                ResNet50x4 & 0.9634 & 0.9360 & 0.9156 & 0.8668 \\
                
                ResNet50x16 & 0.9748 & 0.9540 & 0.9384 & 0.8934 \\
                
                ResNet50x64 & 0.9852 & 0.9674 & 0.9544 & 0.9314 \\
                \midrule
                ViT-B/32 & 0.9682 & 0.9578 & 0.9336 & 0.8760 \\
                
                ViT-B/16 & 0.9814 & 0.9690 & 0.9544 & 0.9092 \\
                
                ViT-L/14 & 0.9926 & 0.9844 & 0.9798 & 0.9584 \\
                
                ViT-L/14@336px & 0.9932 & 0.9848 & 0.9780 & 0.9506 \\
         \bottomrule
    \end{tabular}
    }
  \end{minipage}
  \hspace{1cm}
  \begin{minipage}{\textwidth}
    \centering
    \includegraphics[width=\linewidth]{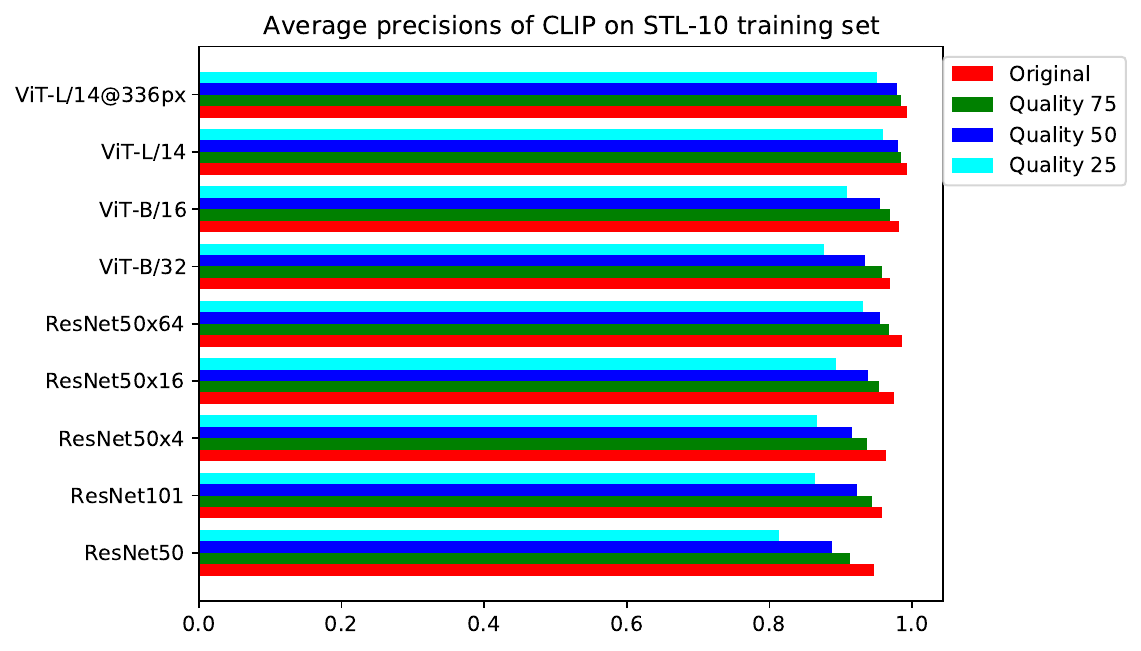}
    \label{fig:plot clip prec stl10 train}
  \end{minipage}%
  \caption{Average precision of CLIP predictions over the training dataset from STL-10 across different image qualities. The definitions of terms are provided in Figure \ref{fig:clip prec cifar10 test}.}\label{tbl:Supp CLIP on STL-10 train}
\end{figure*}

\newpage
\section{More examples of Integrated Gradients on CIFAR-10}
We provide more examples of visualised integrated gradients on CIFAR-10.
\begin{figure}[ht]
		\centering
		%
		%
		\begin{subfigure}[h]{0.45\textwidth}
			\includegraphics[width=\linewidth ]{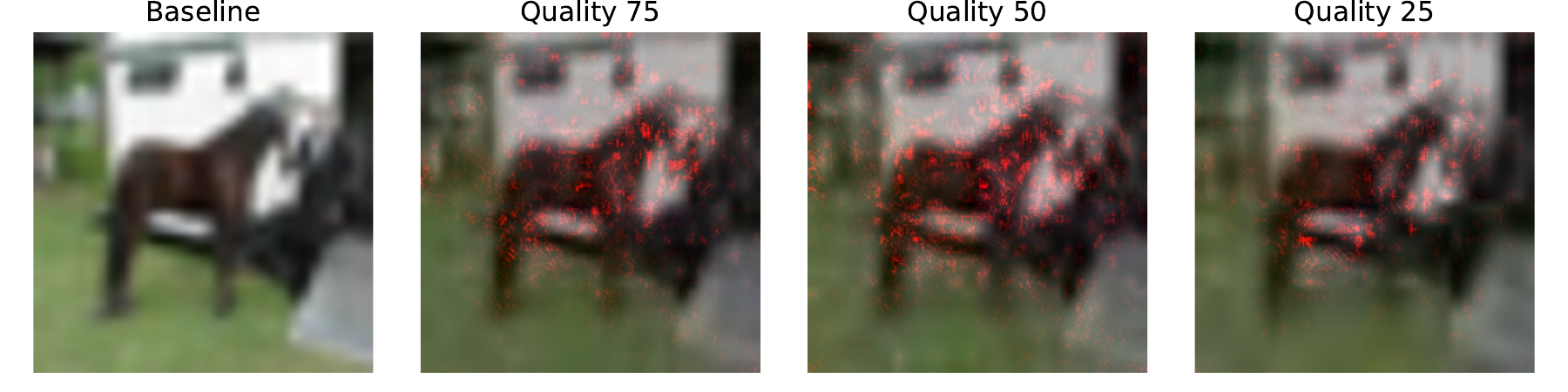}\\
			\includegraphics[width=\linewidth ]{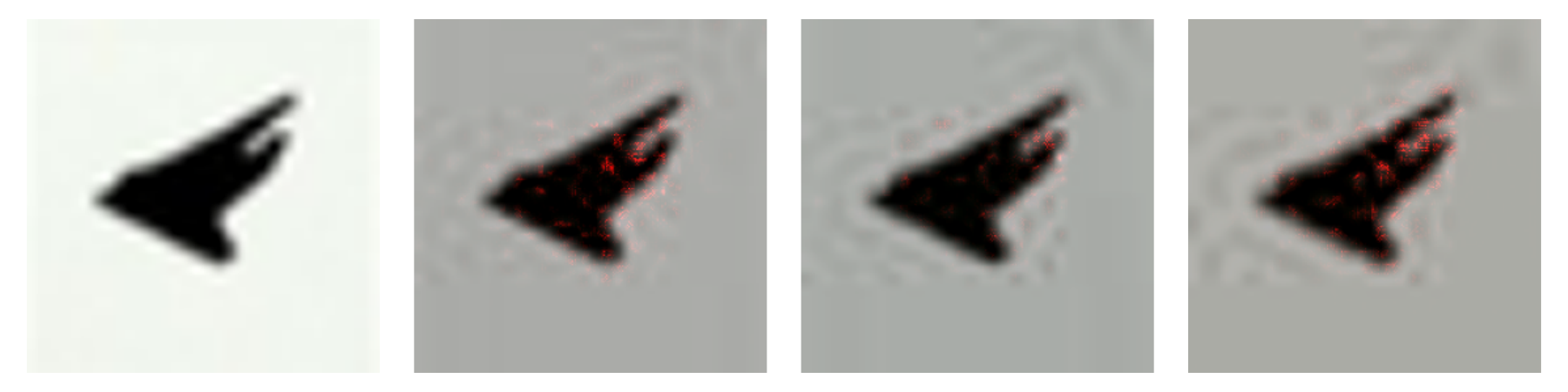}\\
			\includegraphics[width=\linewidth ]{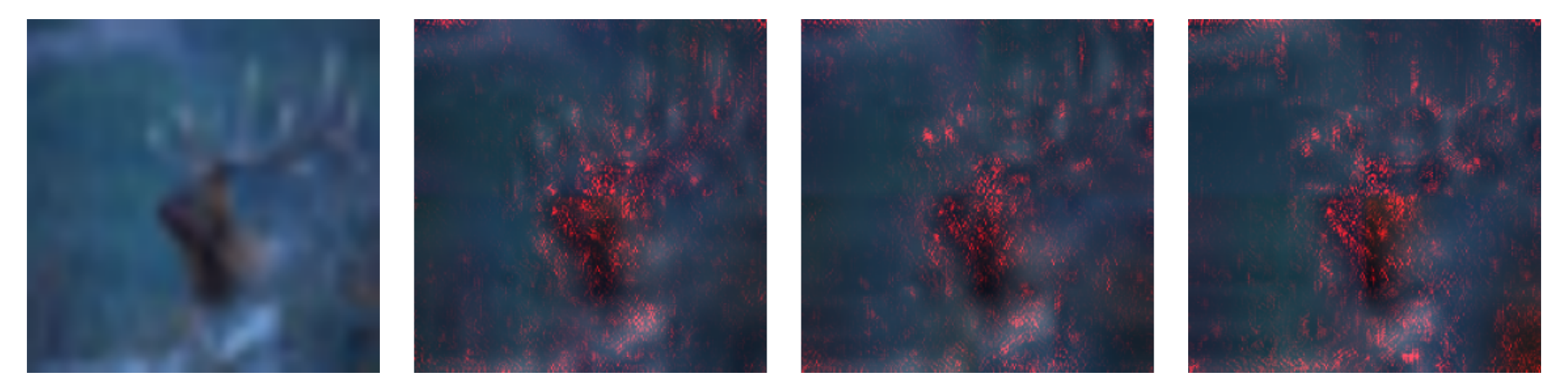}\\
			\includegraphics[width=\linewidth ]{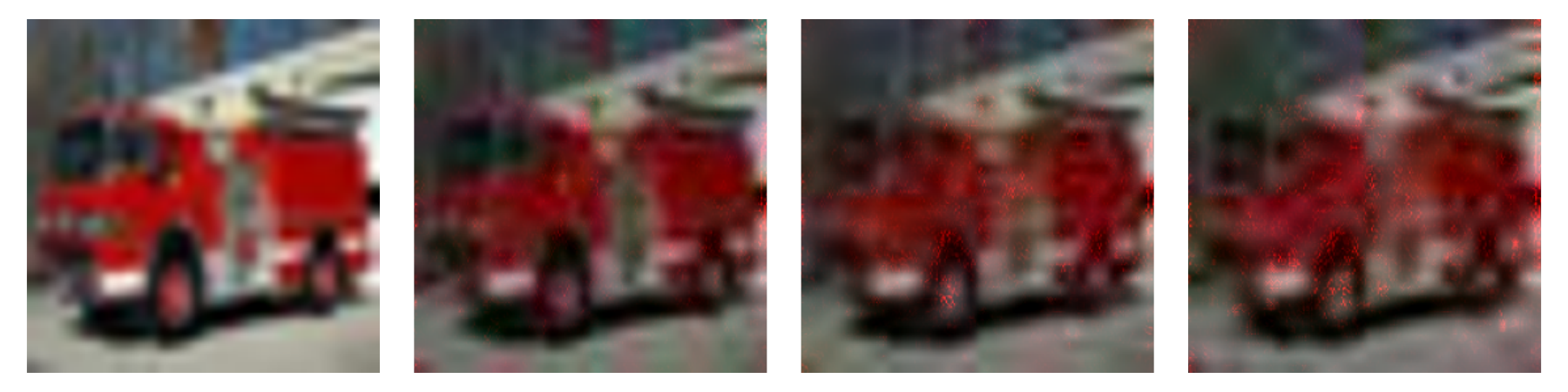}\\
			\includegraphics[width=\linewidth ]{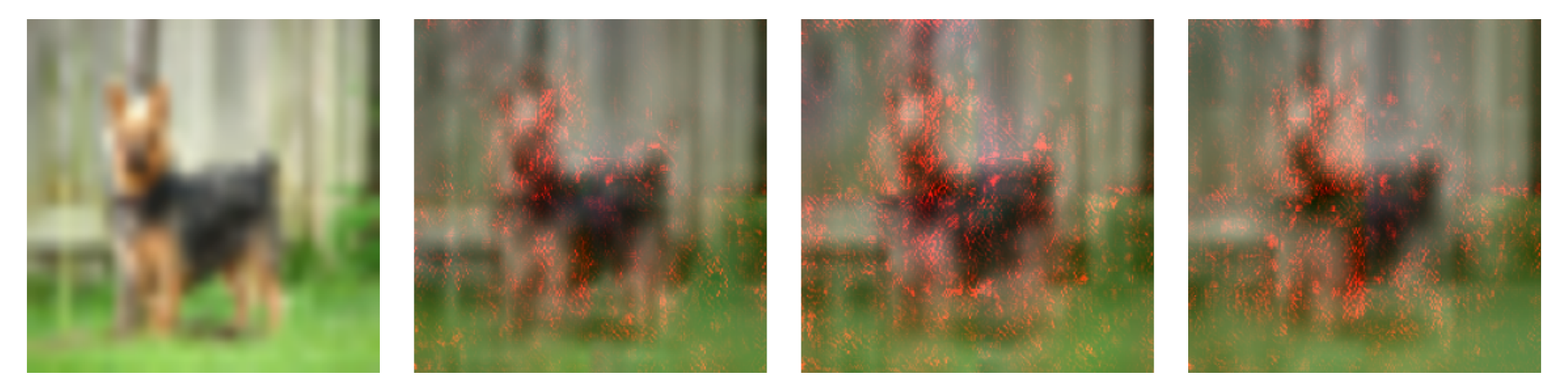}\\
			\includegraphics[width=\linewidth ]{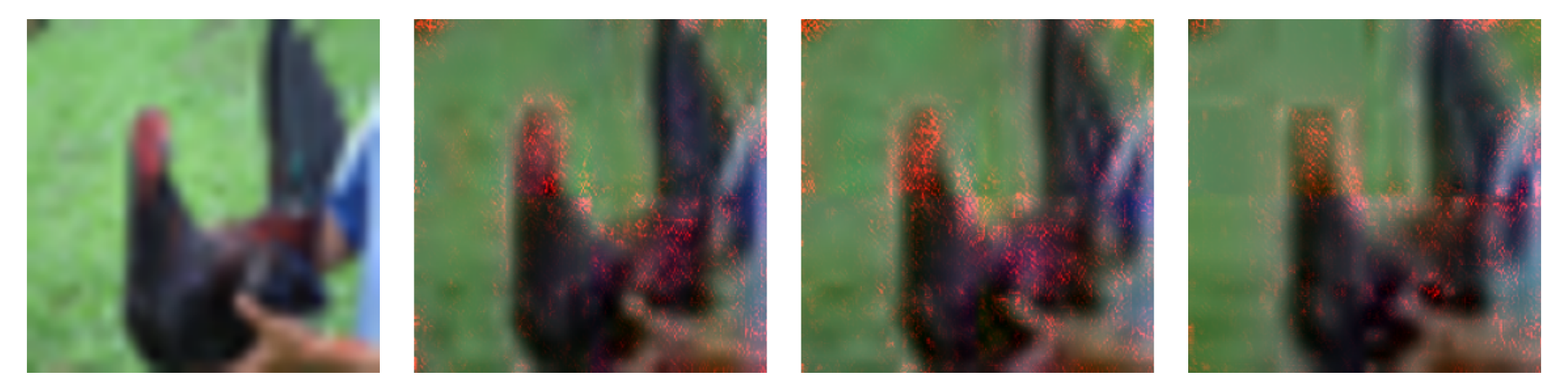}\\
			\includegraphics[width=\linewidth ]{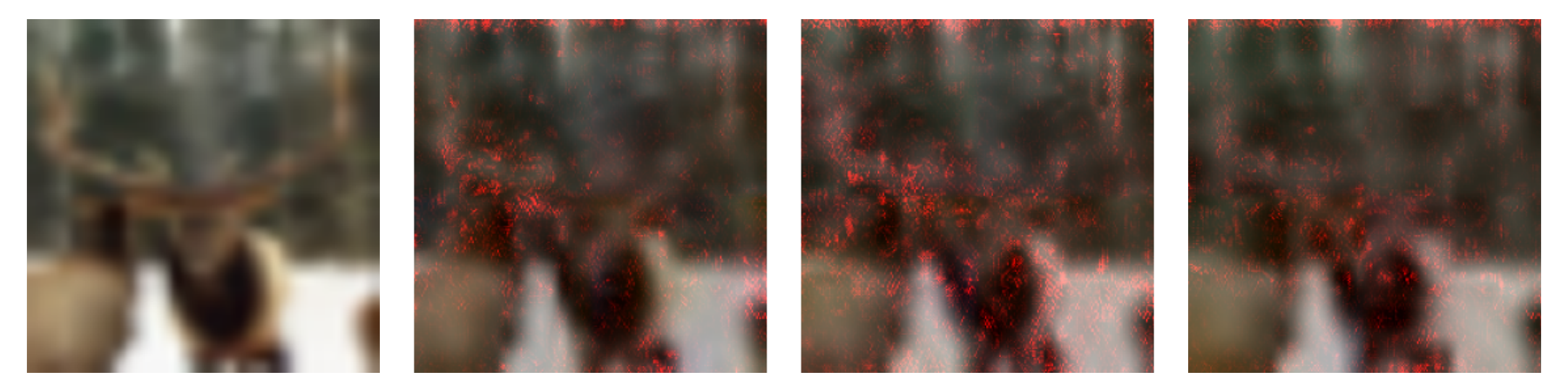}\\
			\includegraphics[width=\linewidth ]{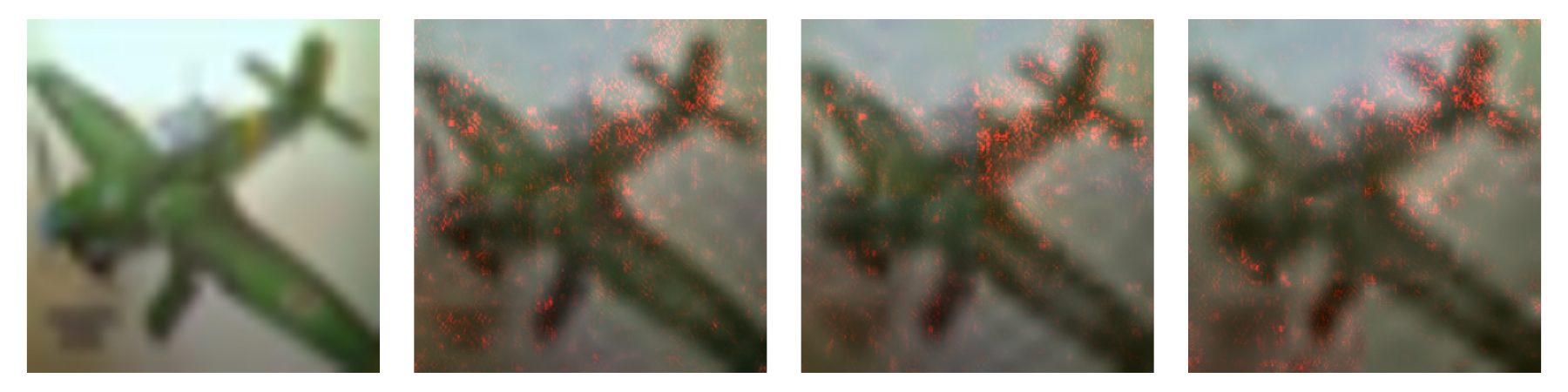}\\
			\includegraphics[width=\linewidth ]{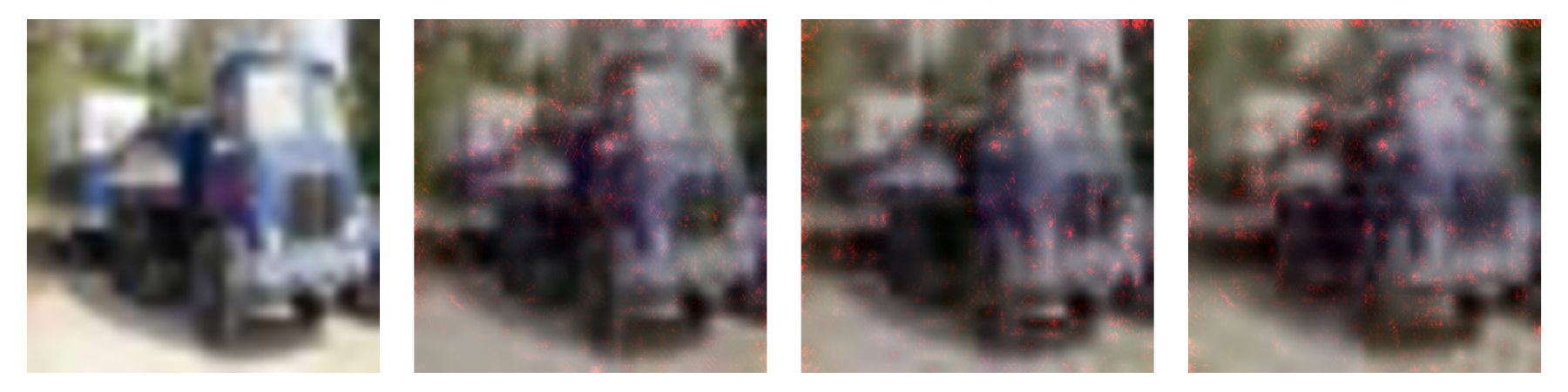}\\
			\includegraphics[width=\linewidth ]{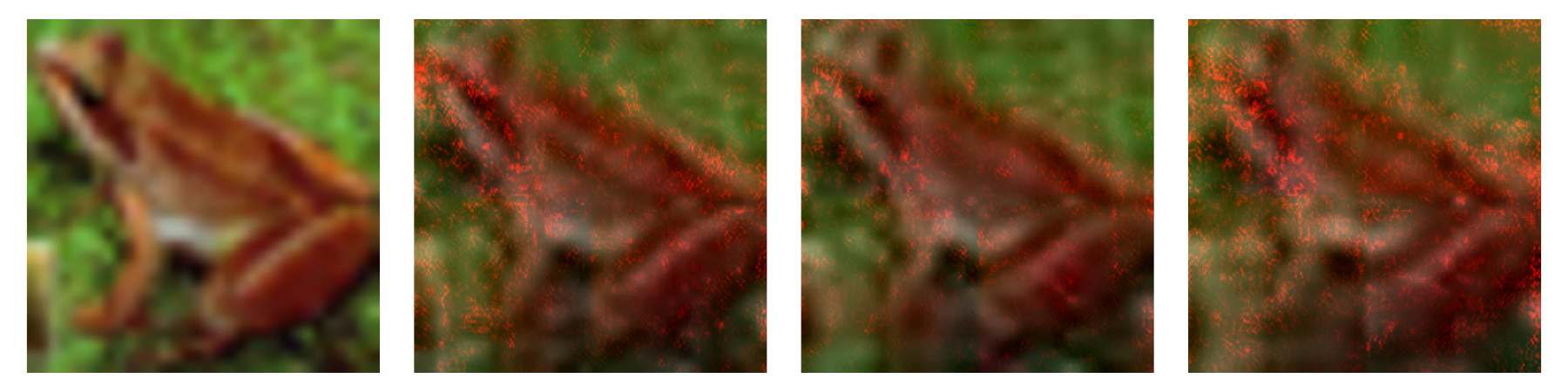}\\
			\caption{ResNet-50 negative}
		\end{subfigure}\hspace{1cm}
		\begin{subfigure}[h]{0.45\textwidth}
			\includegraphics[width=\linewidth ]{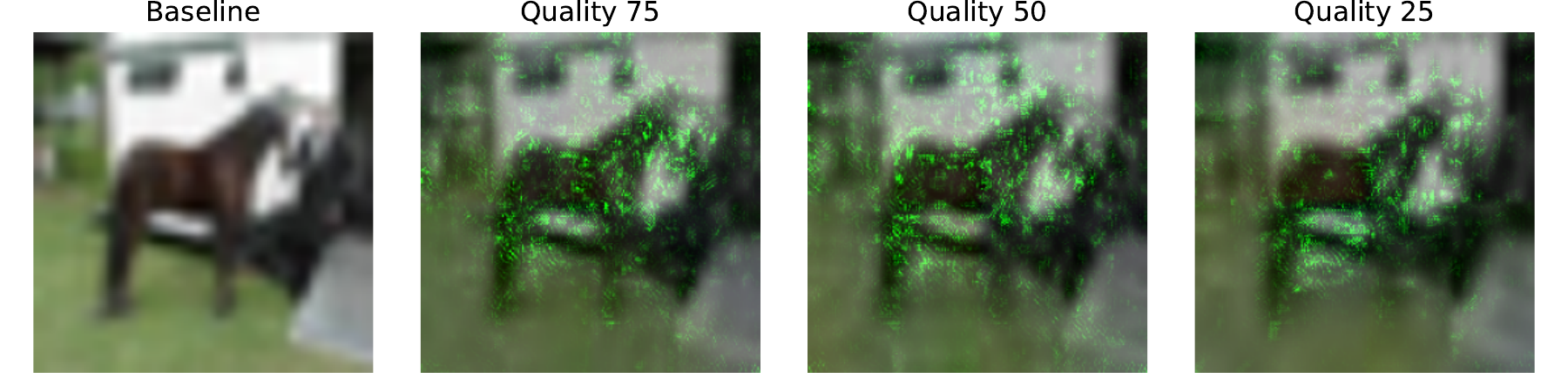}\\
			\includegraphics[width=\linewidth ]{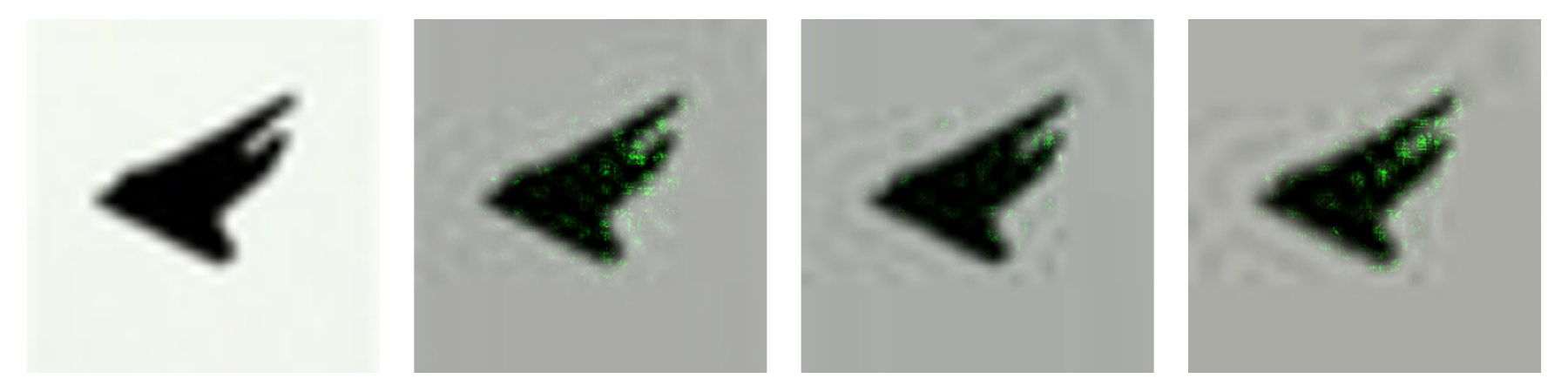}\\
			\includegraphics[width=\linewidth ]{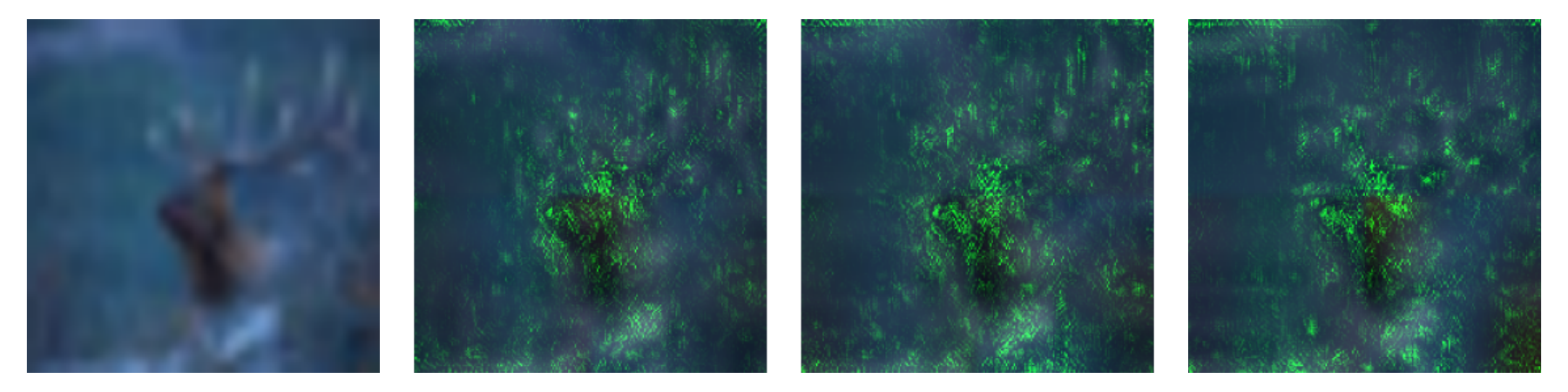}\\
			\includegraphics[width=\linewidth ]{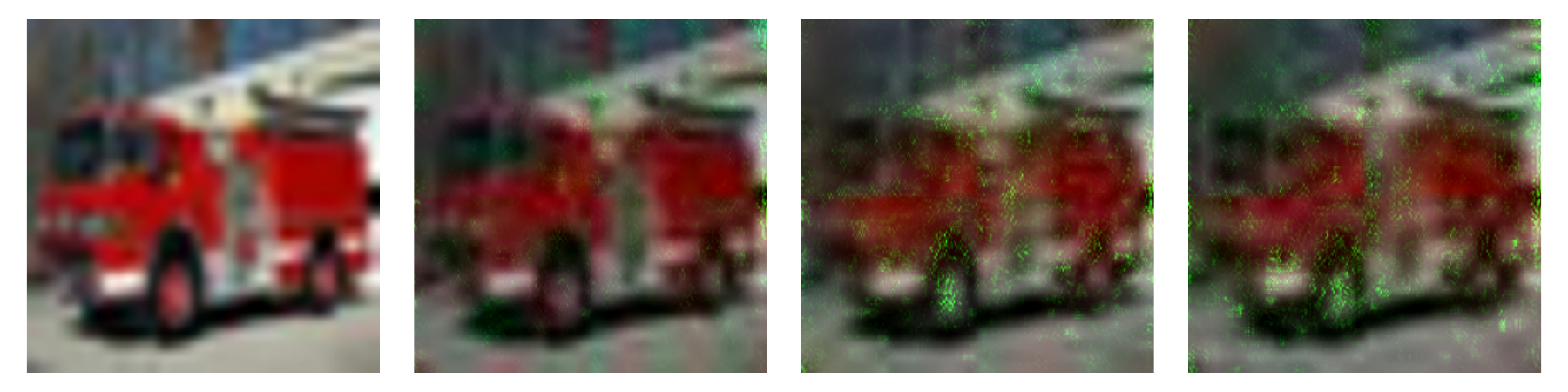}\\
			\includegraphics[width=\linewidth ]{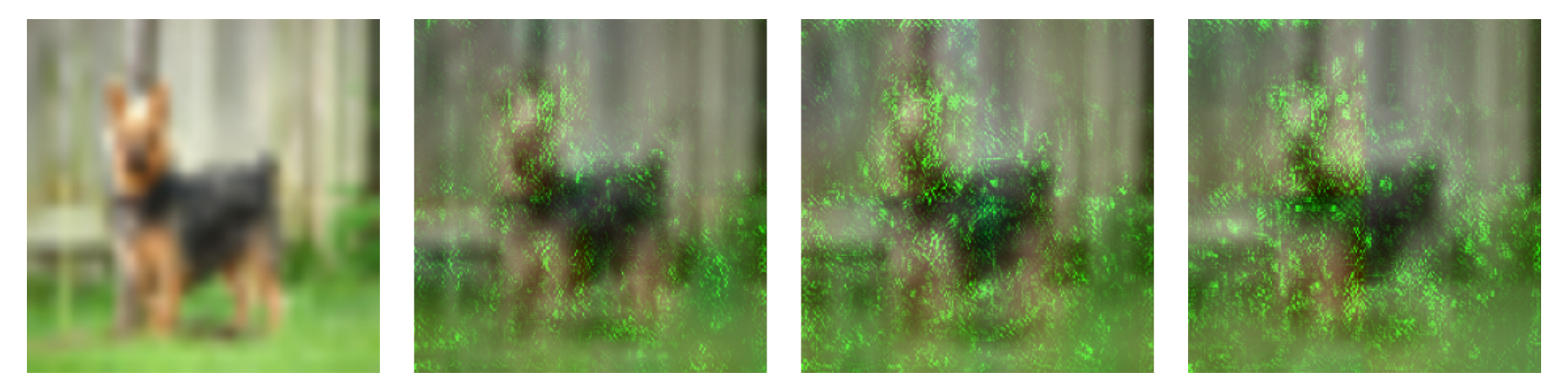}\\
			\includegraphics[width=\linewidth ]{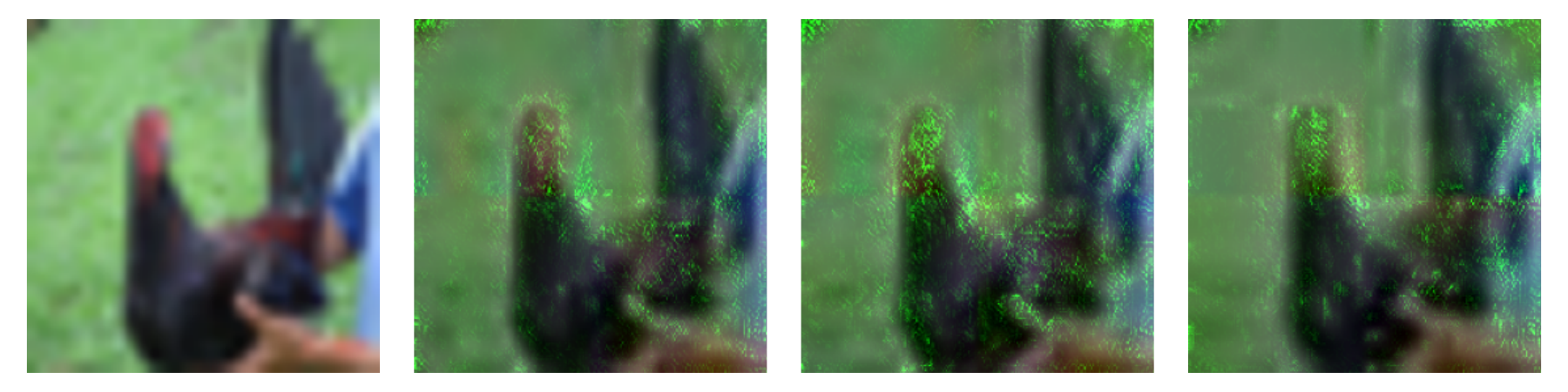}\\
			\includegraphics[width=\linewidth ]{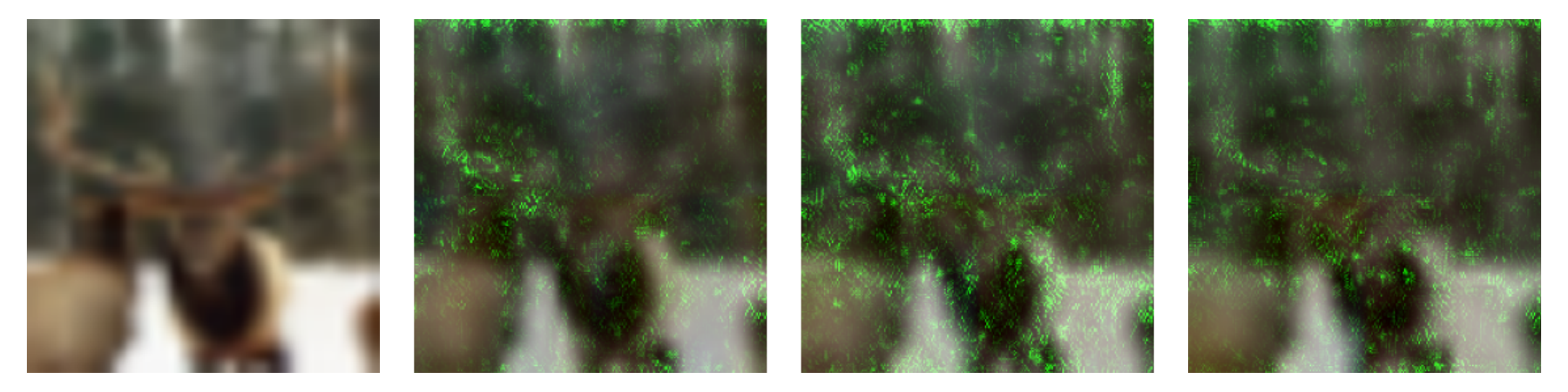}\\
			\includegraphics[width=\linewidth ]{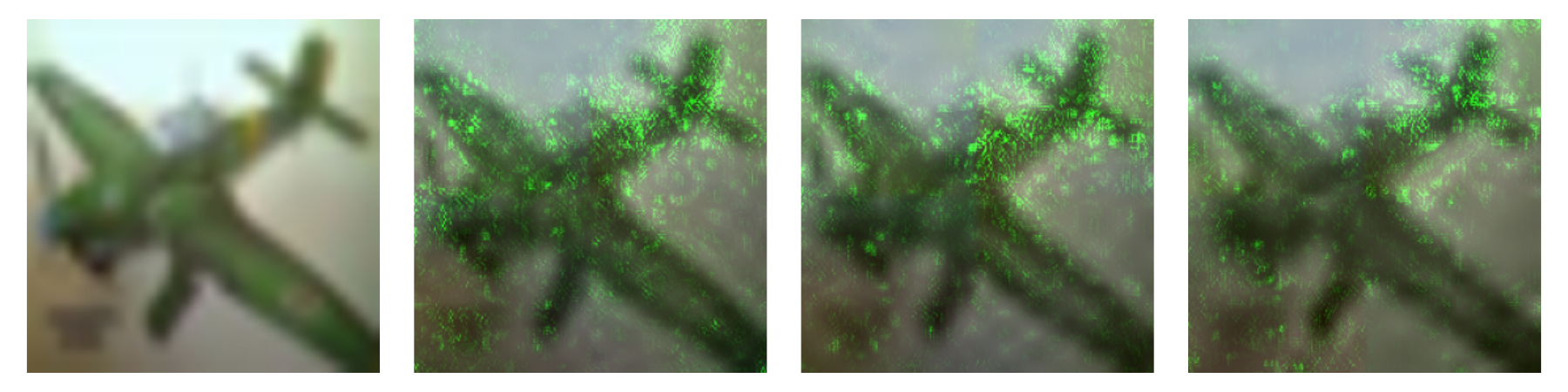}\\
			\includegraphics[width=\linewidth ]{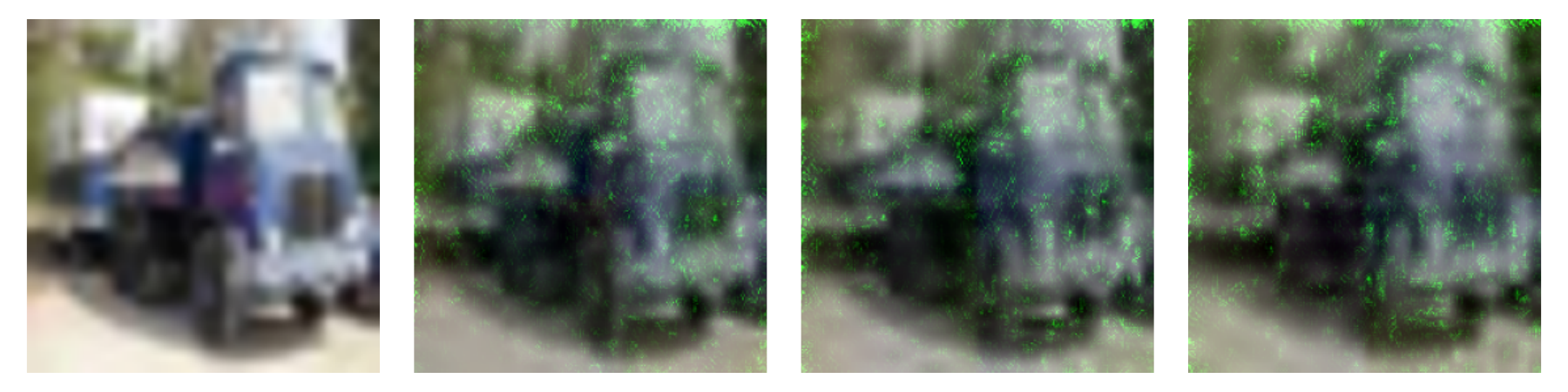}\\
			\includegraphics[width=\linewidth ]{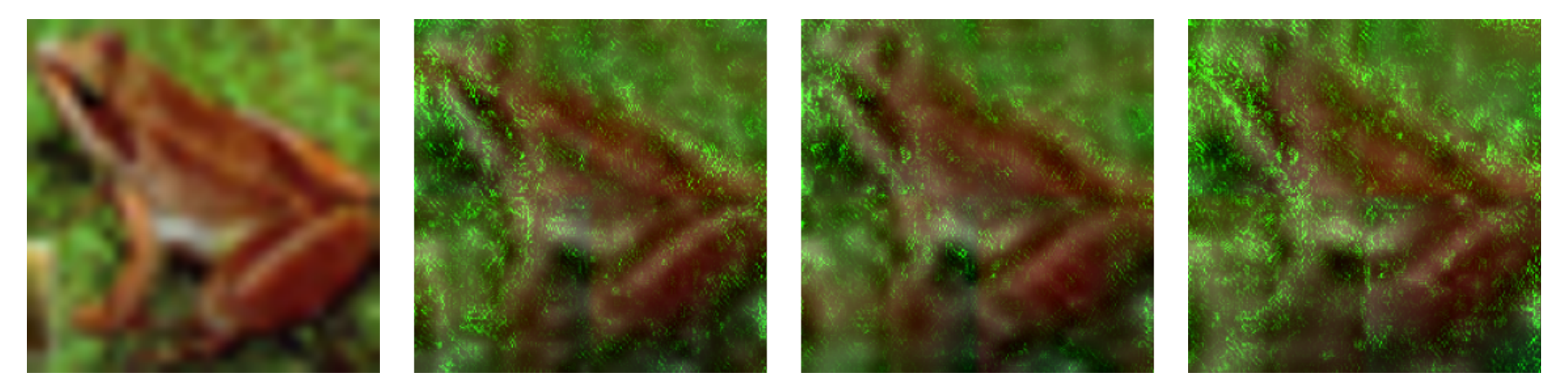}\\
			\caption{ResNet-50 positive}
		\end{subfigure} \\
    \caption{Examples on visualisation of integrated gradients for ResNet-50.}\label{fig:RN50 supp cifar-10}
\end{figure}
\begin{figure}[ht]
		\centering
		%
		%
		\begin{subfigure}[h]{0.45\textwidth}
			\includegraphics[width=\linewidth]{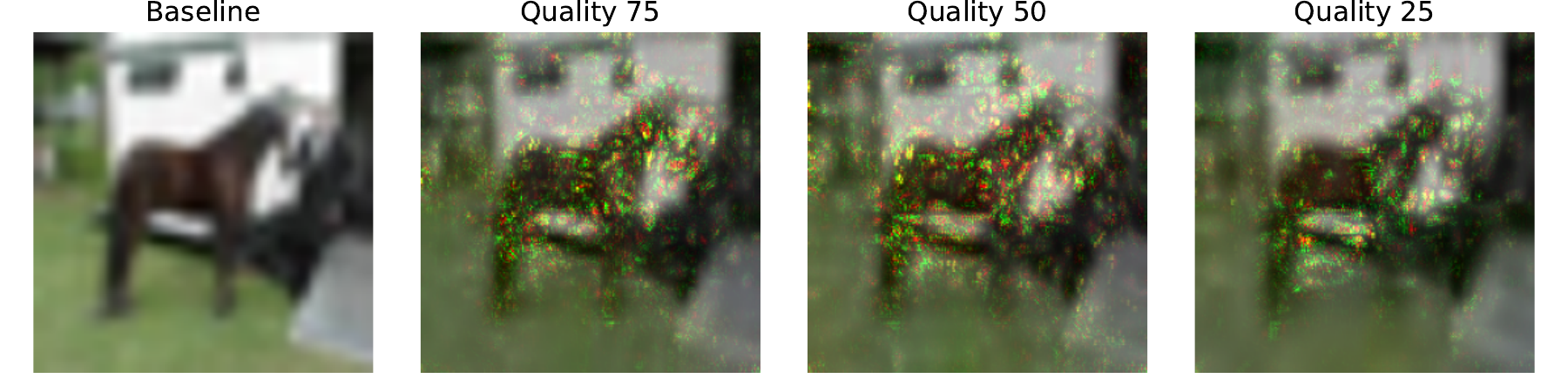}\\
			\includegraphics[width=\linewidth]{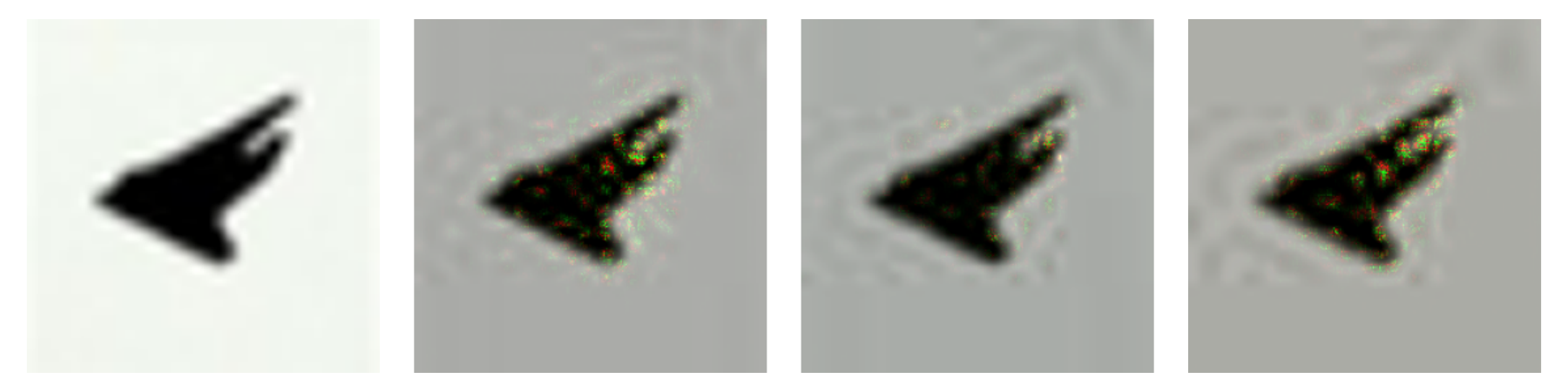}\\
			\includegraphics[width=\linewidth]{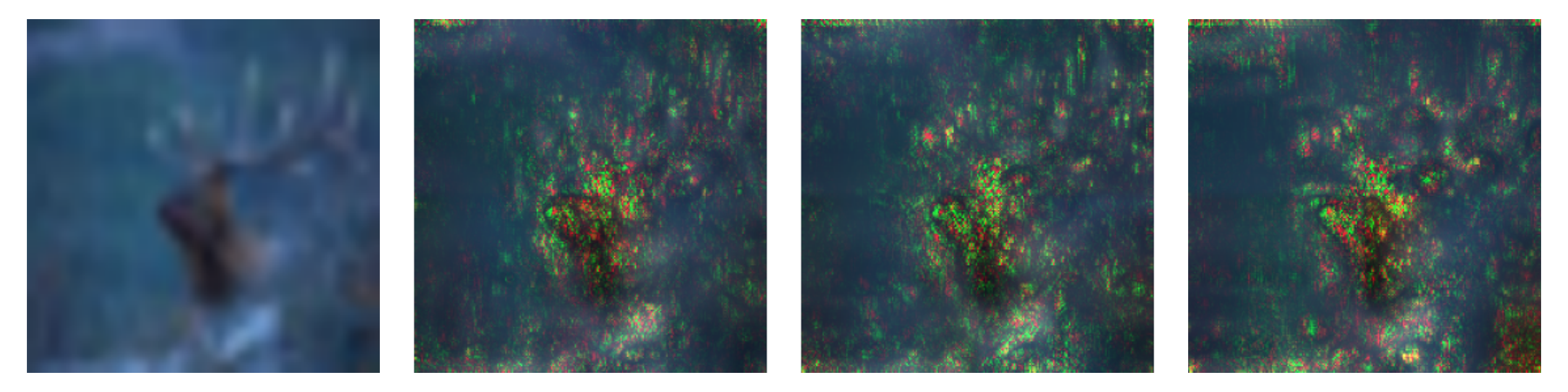}\\
			\includegraphics[width=\linewidth]{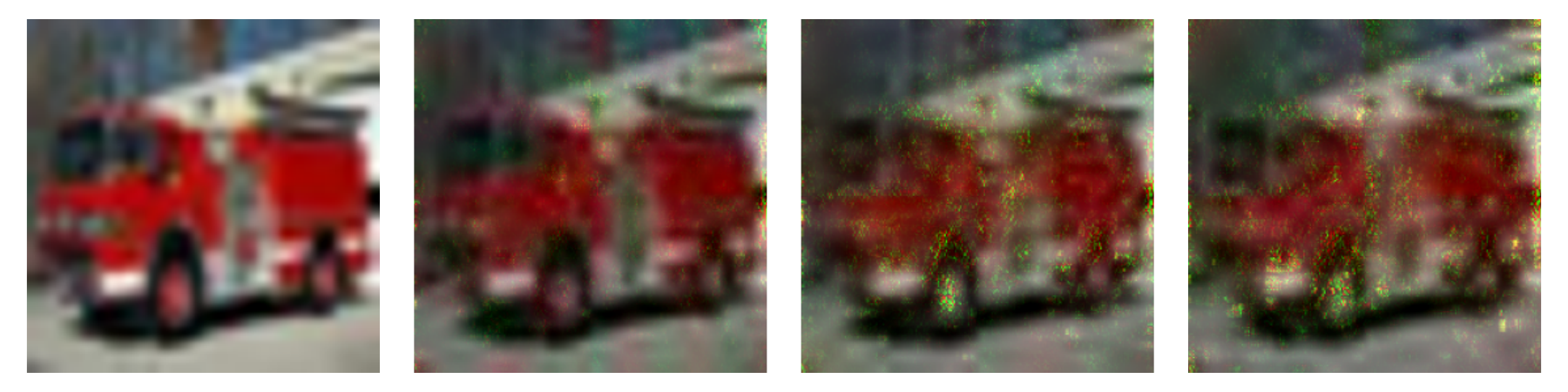}\\
			\includegraphics[width=\linewidth]{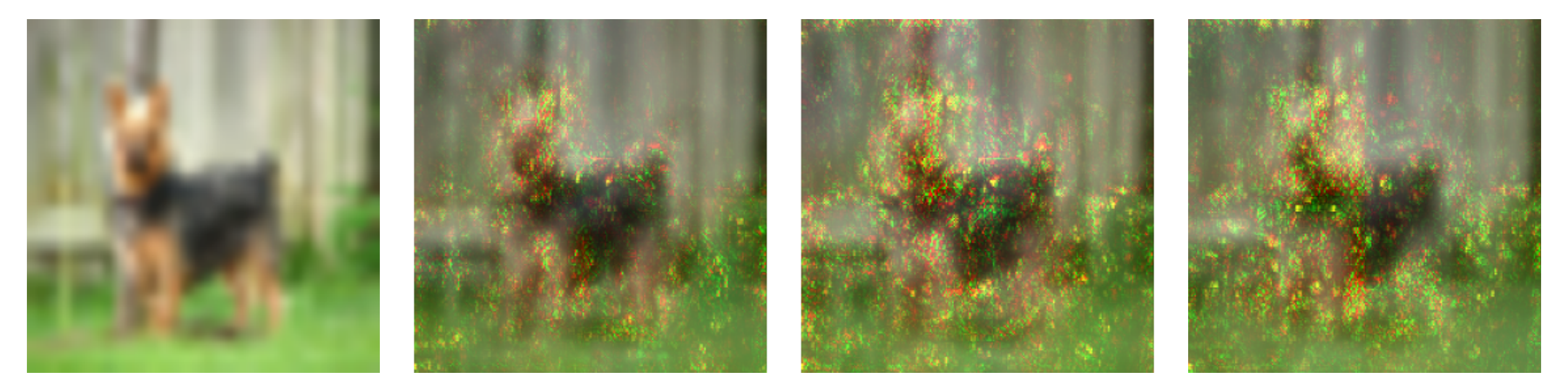}\\
			\includegraphics[width=\linewidth]{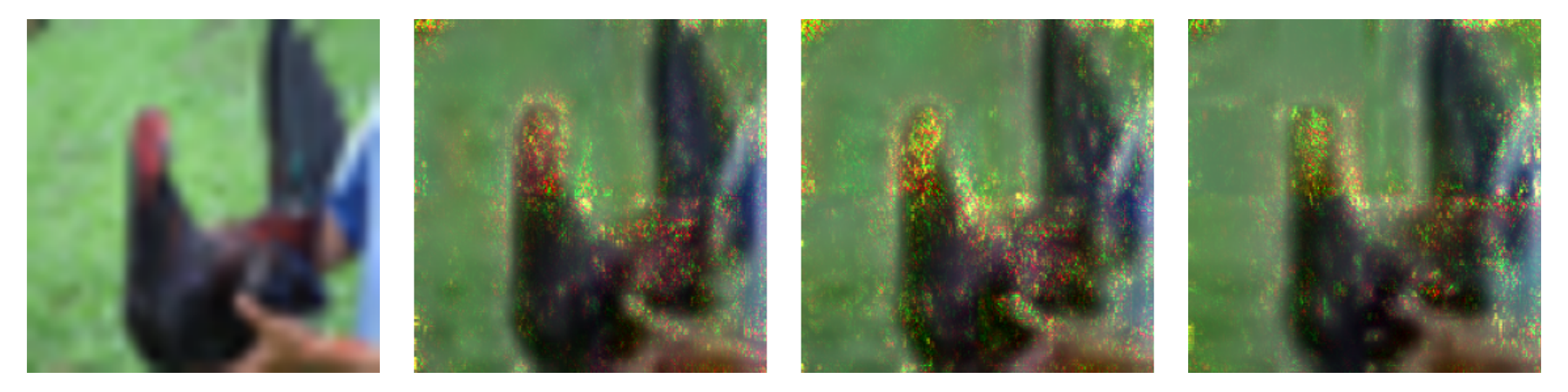}\\
			\includegraphics[width=\linewidth]{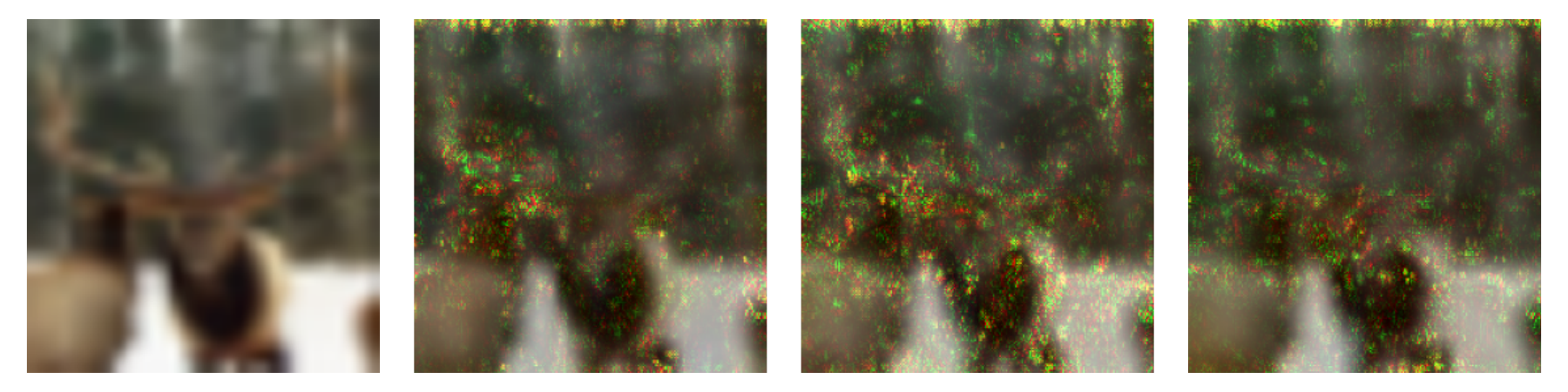}\\
			\includegraphics[width=\linewidth]{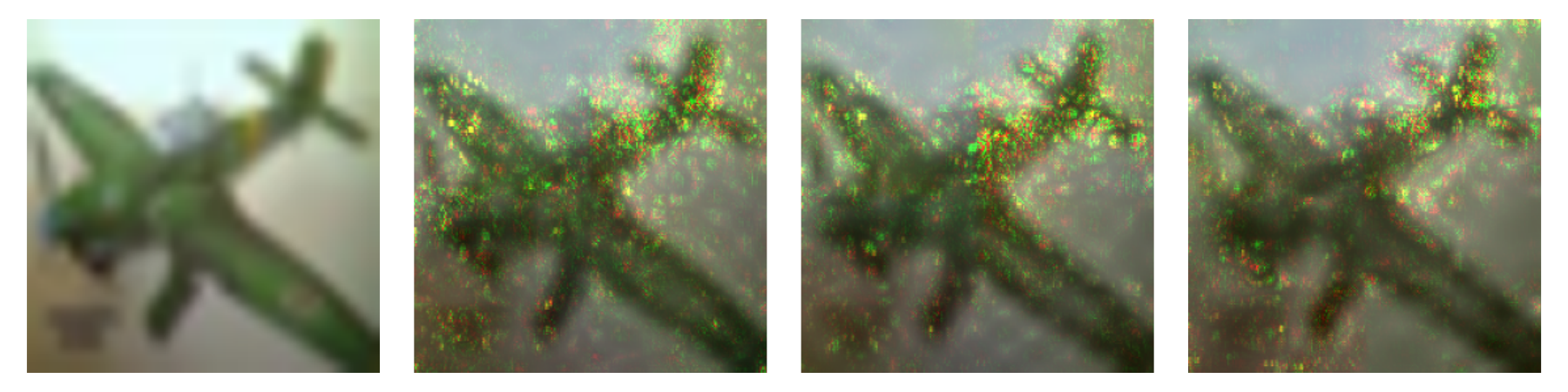}\\
			\includegraphics[width=\linewidth]{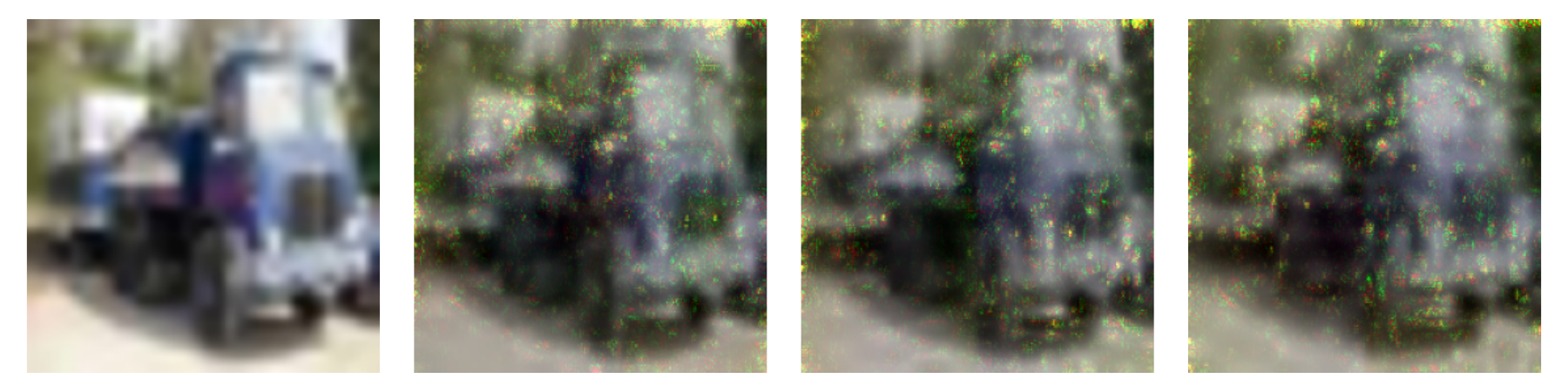}\\
			\includegraphics[width=\linewidth]{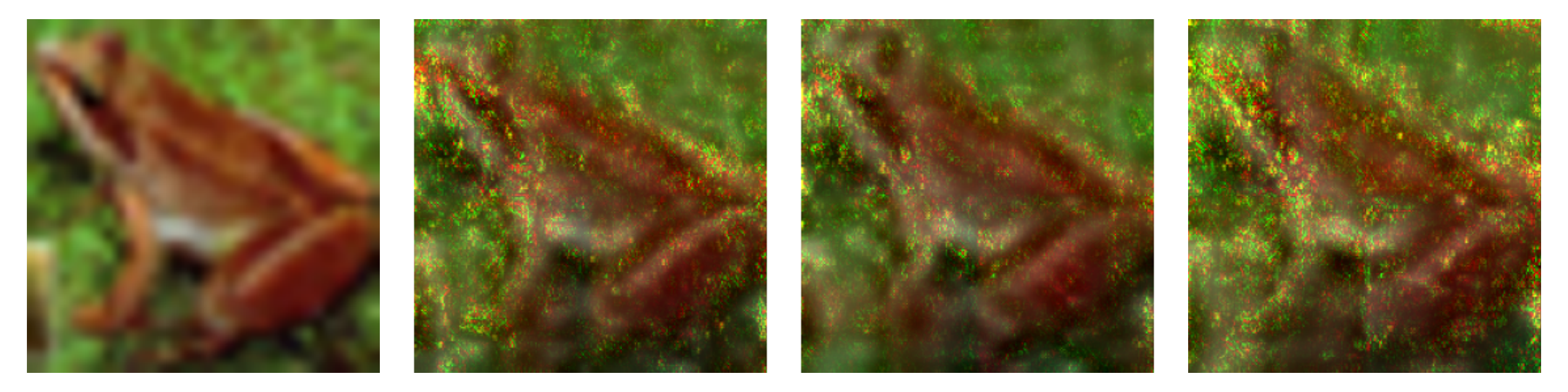}\\
			\caption{ResNet-50 both}\label{fig:Supp RN50 both CIFAR}
		\end{subfigure}\hspace{1cm}
		\begin{subfigure}[h]{0.45\textwidth}
			\includegraphics[width=\linewidth]{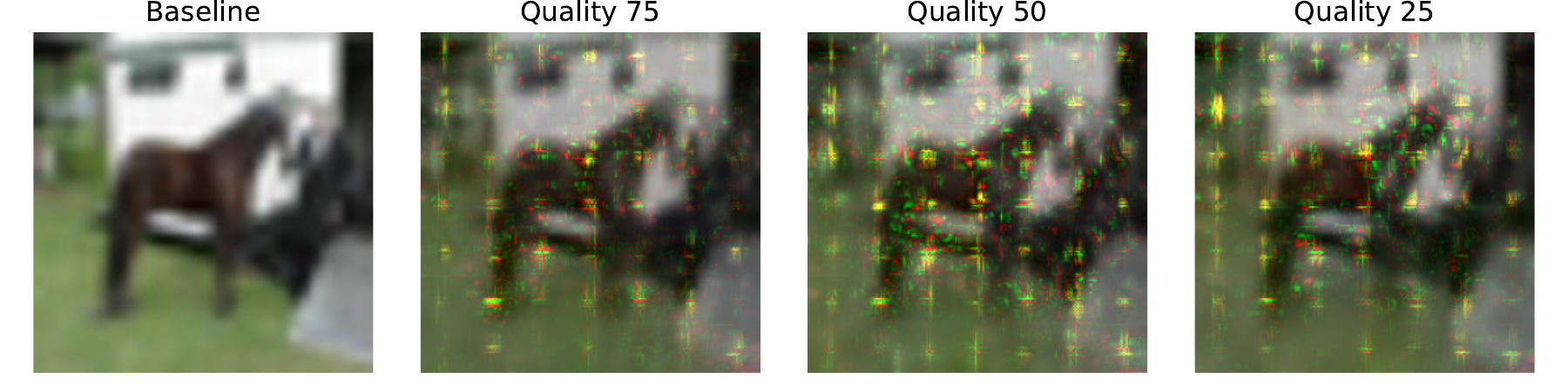}\\
			\includegraphics[width=\linewidth]{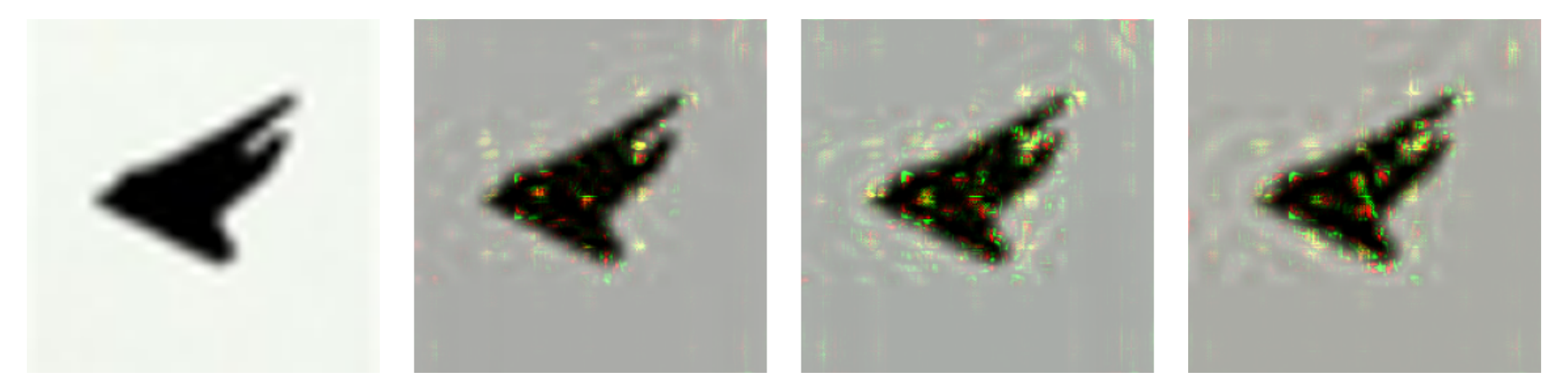}\\
			\includegraphics[width=\linewidth]{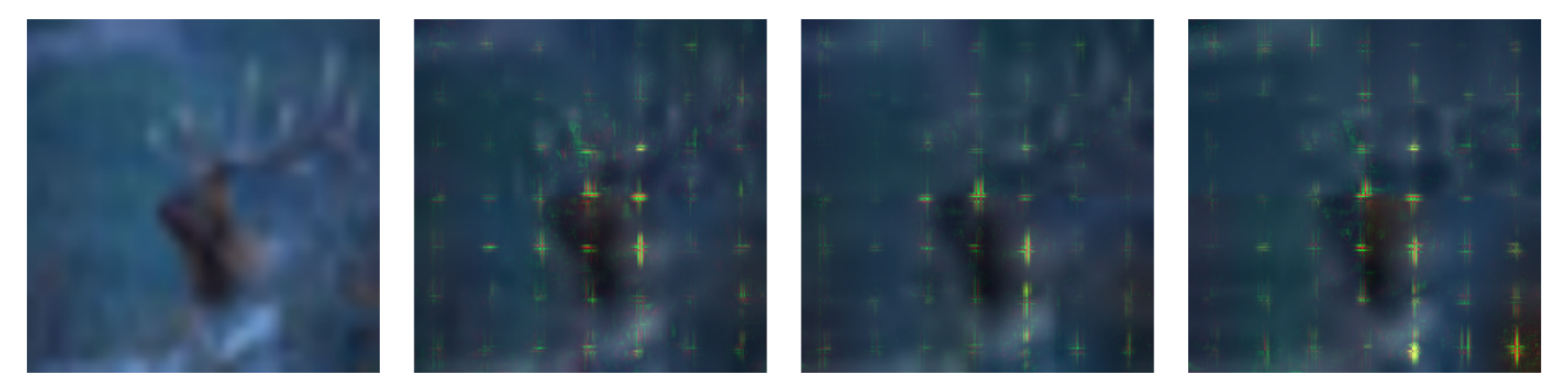}\\
			\includegraphics[width=\linewidth]{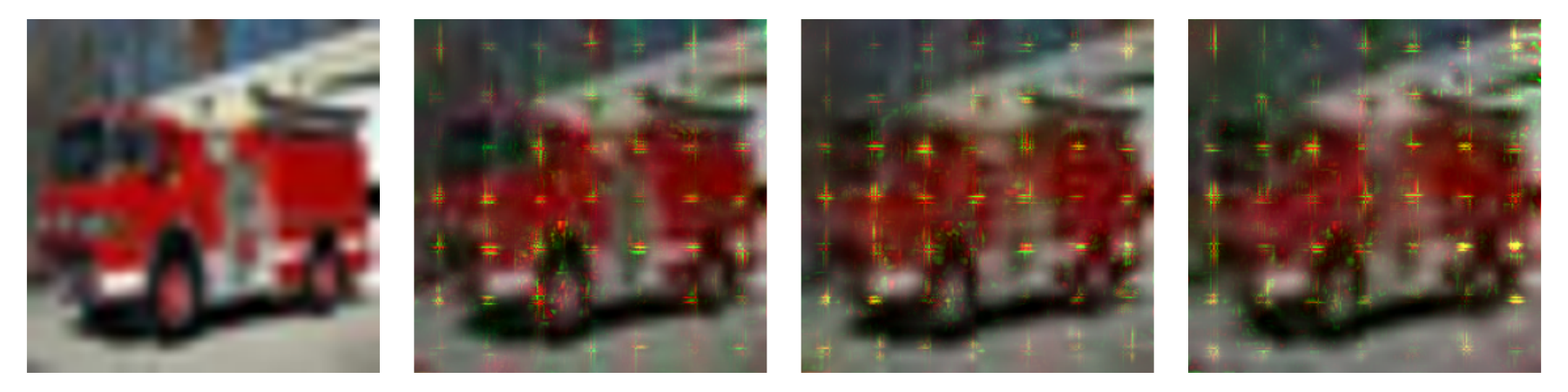}\\
			\includegraphics[width=\linewidth]{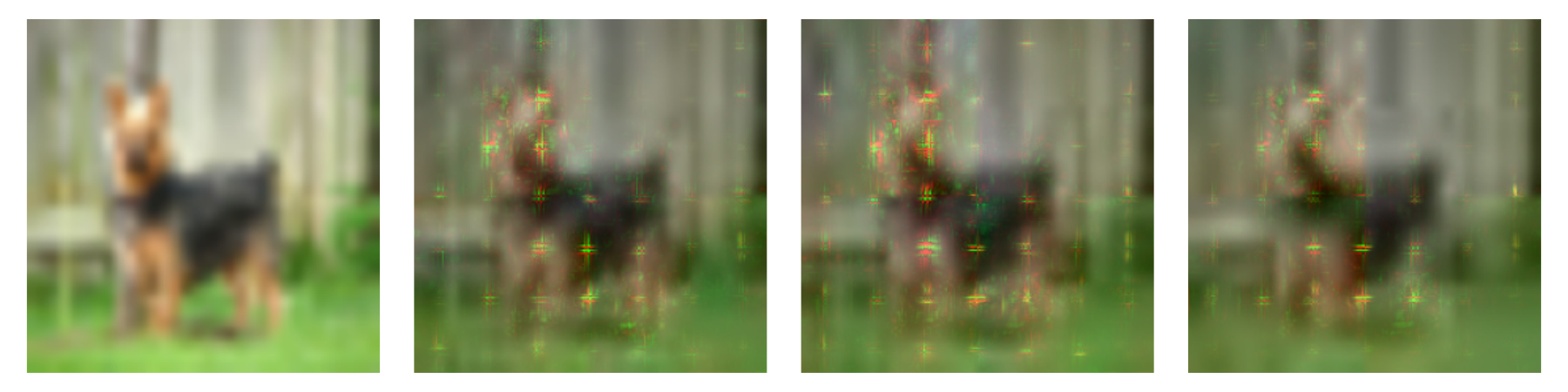}\\
			\includegraphics[width=\linewidth]{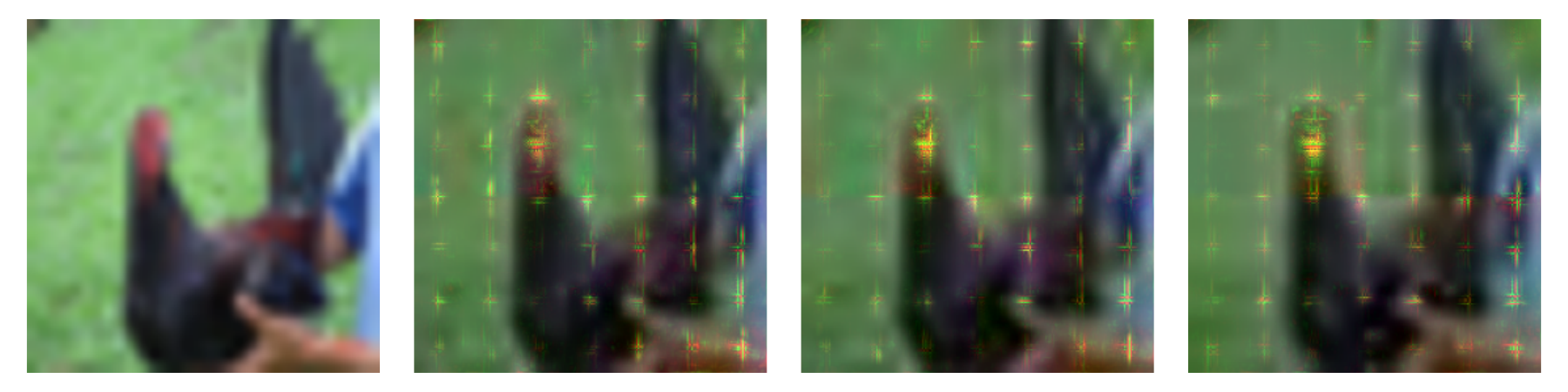}\\
			\includegraphics[width=\linewidth]{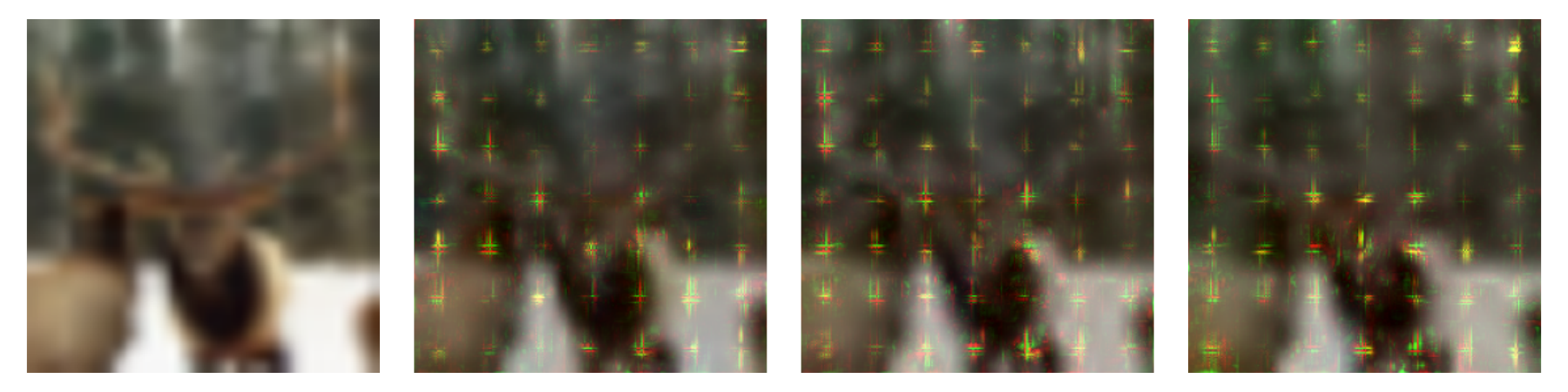}\\
			\includegraphics[width=\linewidth]{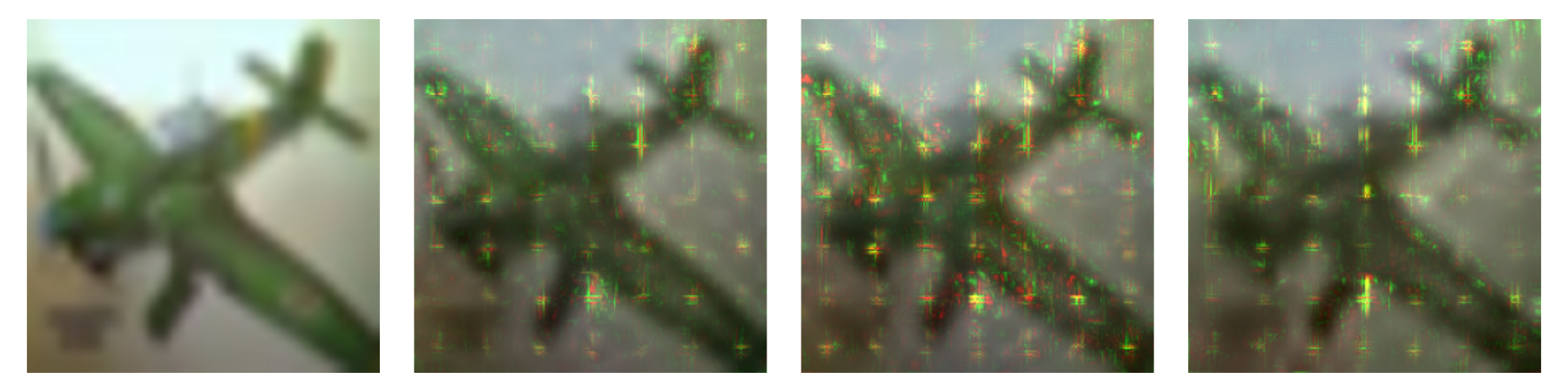}\\
			\includegraphics[width=\linewidth]{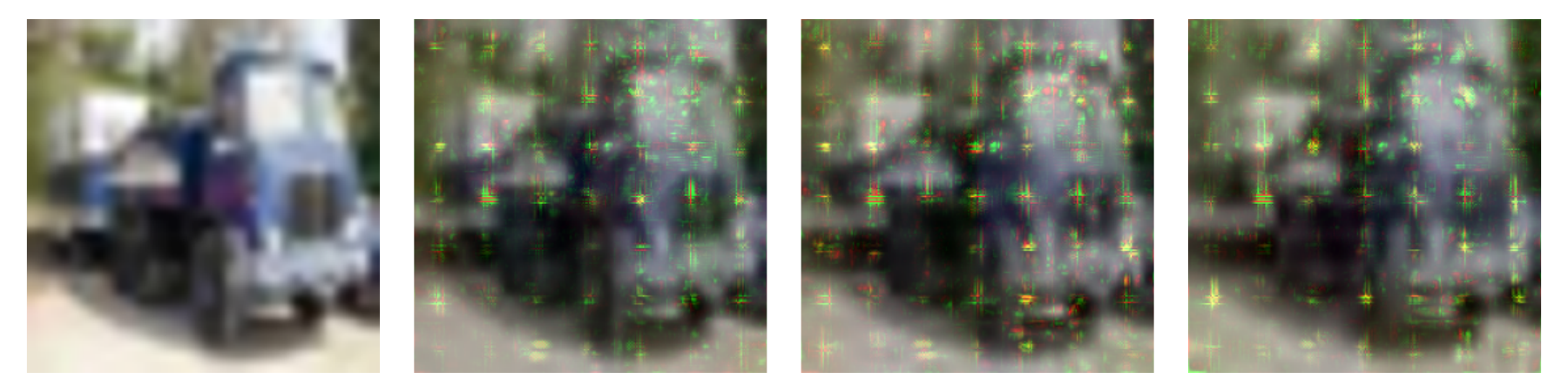}\\
			\includegraphics[width=\linewidth]{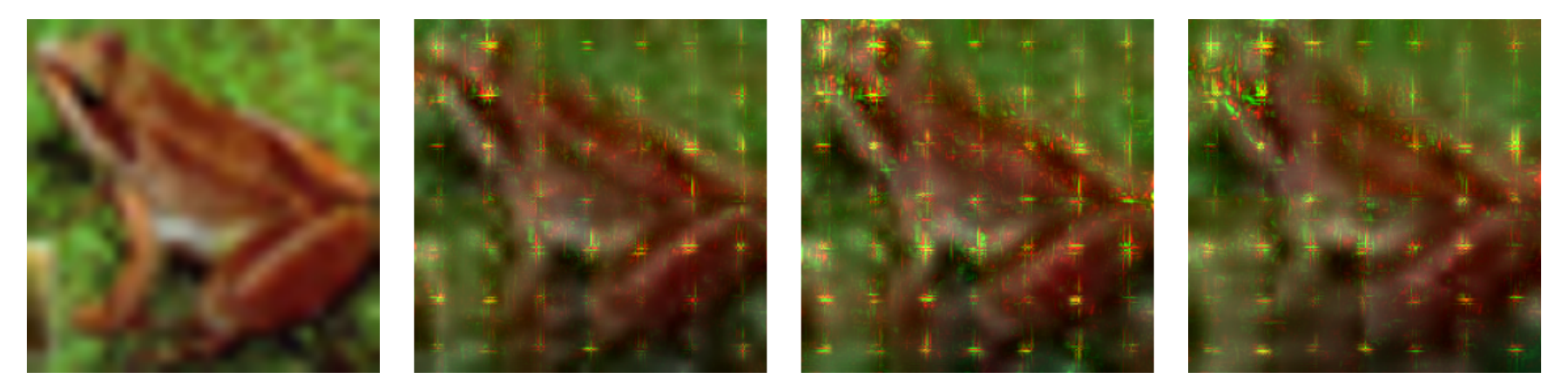}\\
			\caption{ViT both}\label{fig:Supp ViT both CIFAR}
		\end{subfigure} \\
		\caption{Examples on visualisation of integrated gradients for ResNet-50 and ViT-B/32.}\label{fig:RN and ViT both supp cifar-10}
\end{figure}
\begin{figure}[ht]
		\centering
		%
		%
		\begin{subfigure}[h]{0.45\textwidth}
			\includegraphics[width=\linewidth ]{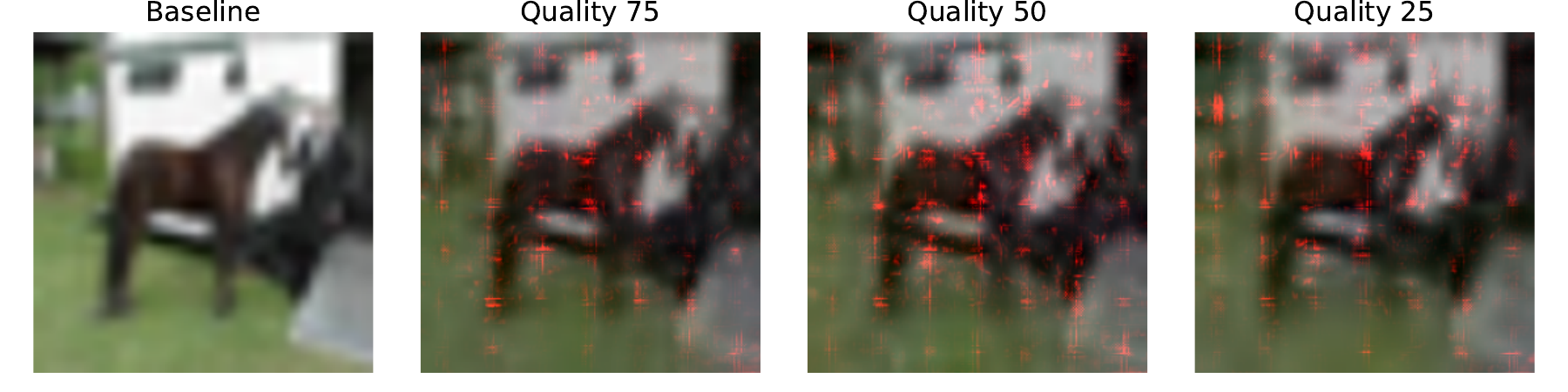}\\
			\includegraphics[width=\linewidth ]{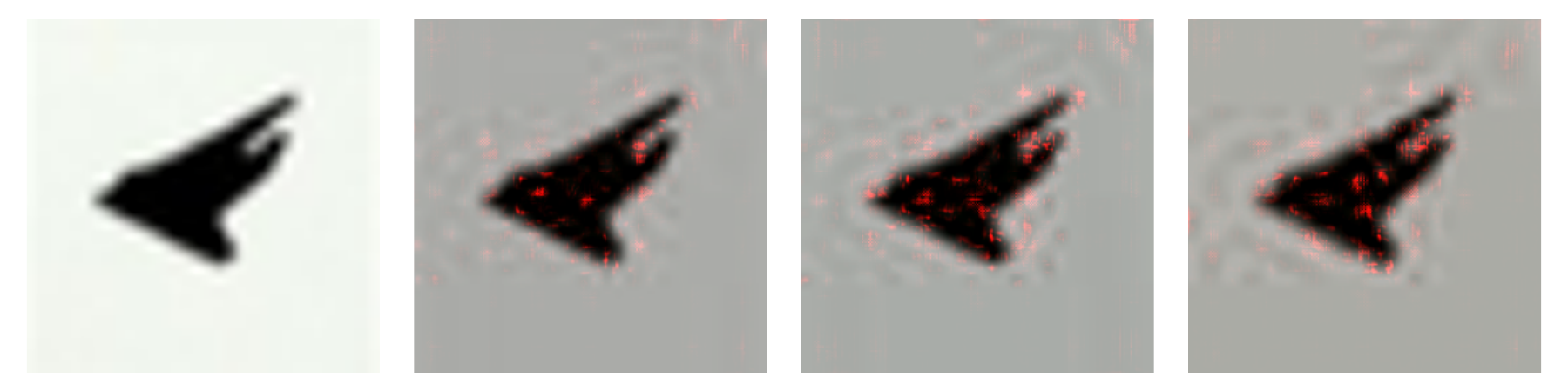}\\
			\includegraphics[width=\linewidth ]{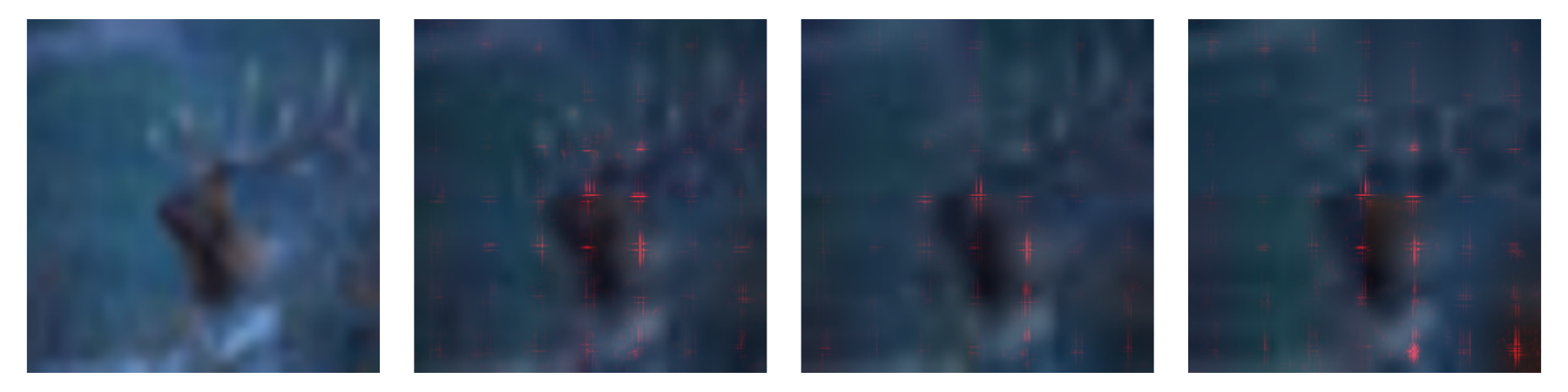}\\
			\includegraphics[width=\linewidth ]{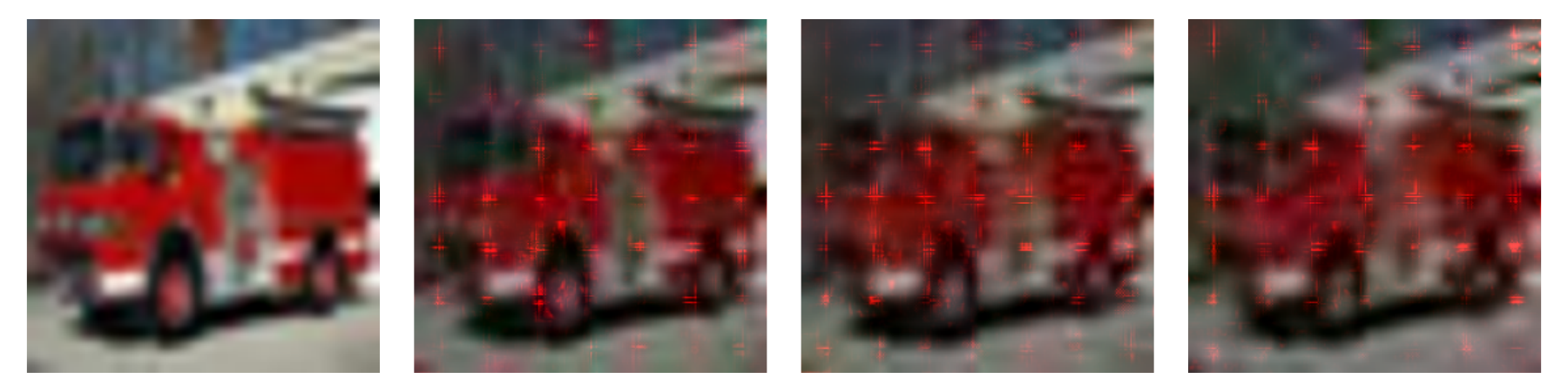}\\
			\includegraphics[width=\linewidth ]{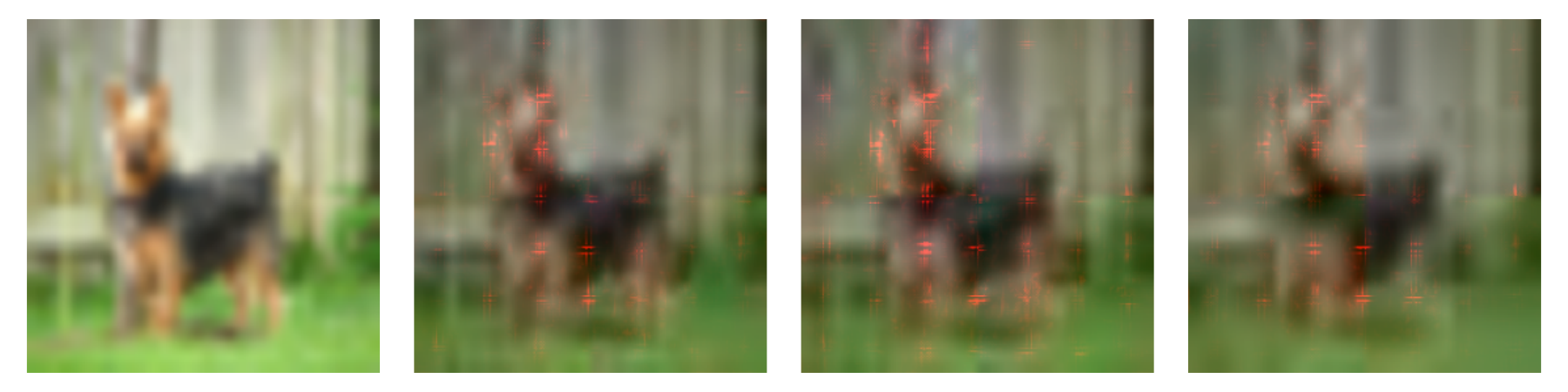}\\
			\includegraphics[width=\linewidth ]{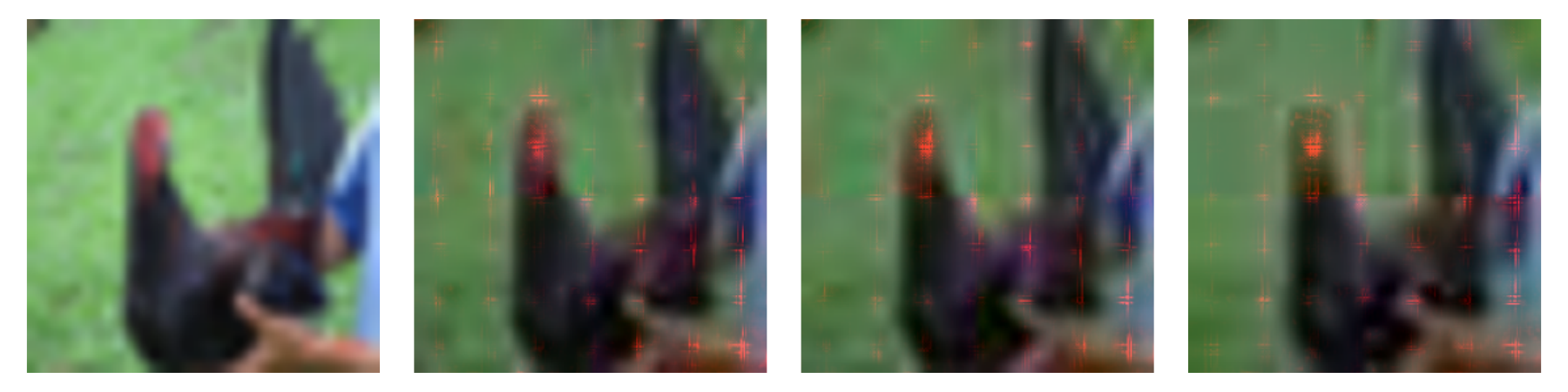}\\
			\includegraphics[width=\linewidth ]{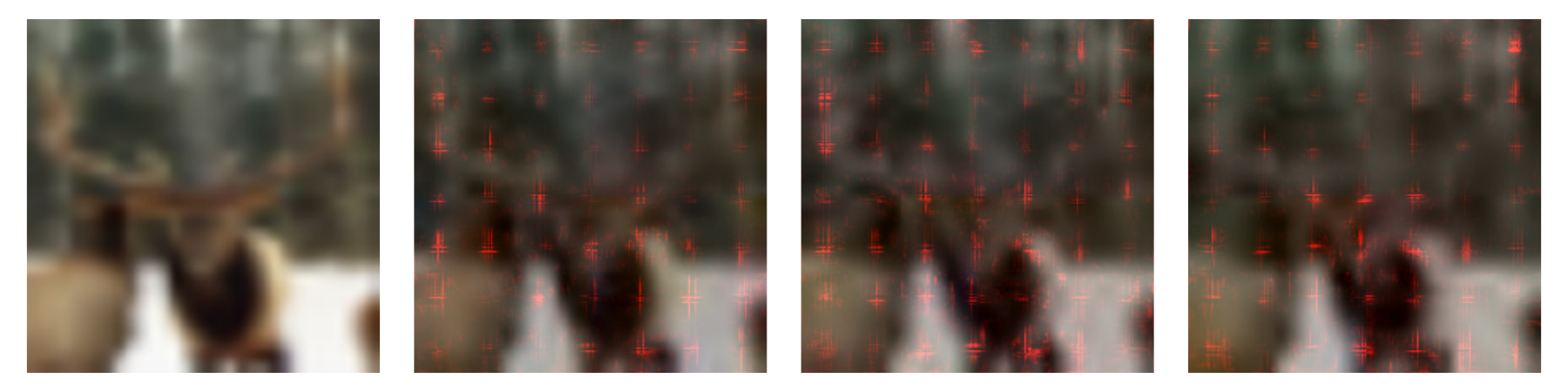}\\
			\includegraphics[width=\linewidth ]{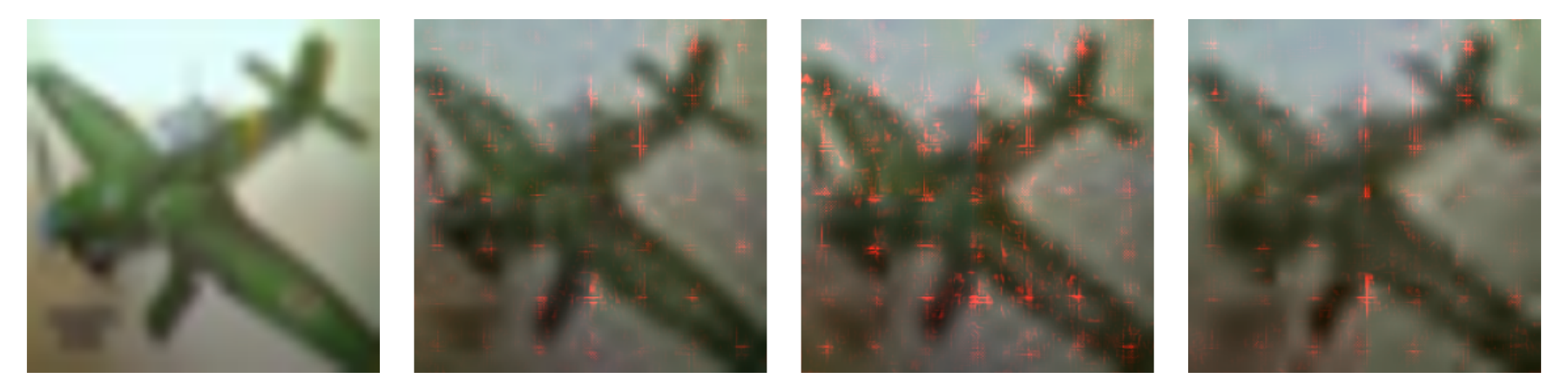}\\
			\includegraphics[width=\linewidth ]{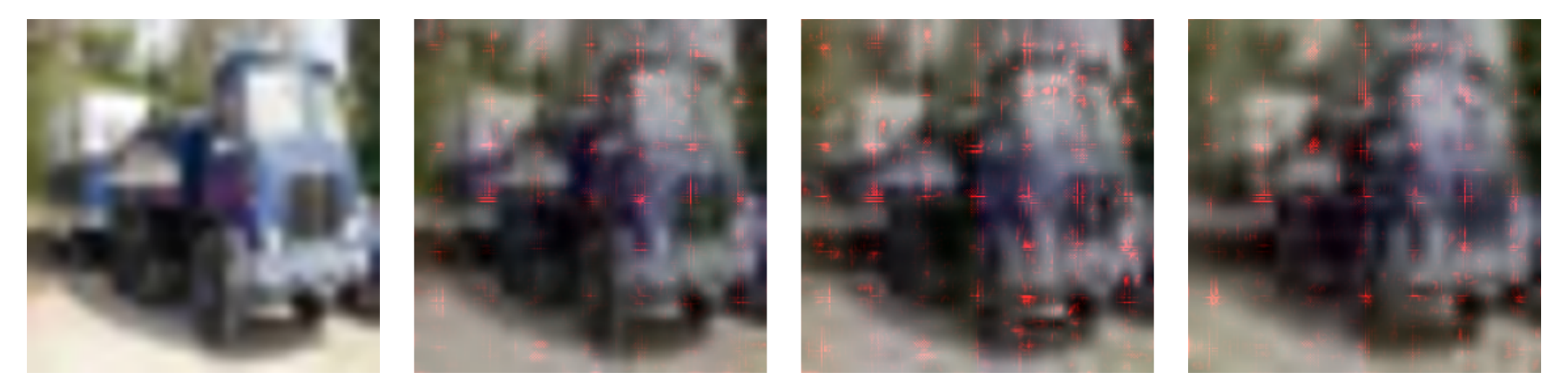}\\
			\includegraphics[width=\linewidth ]{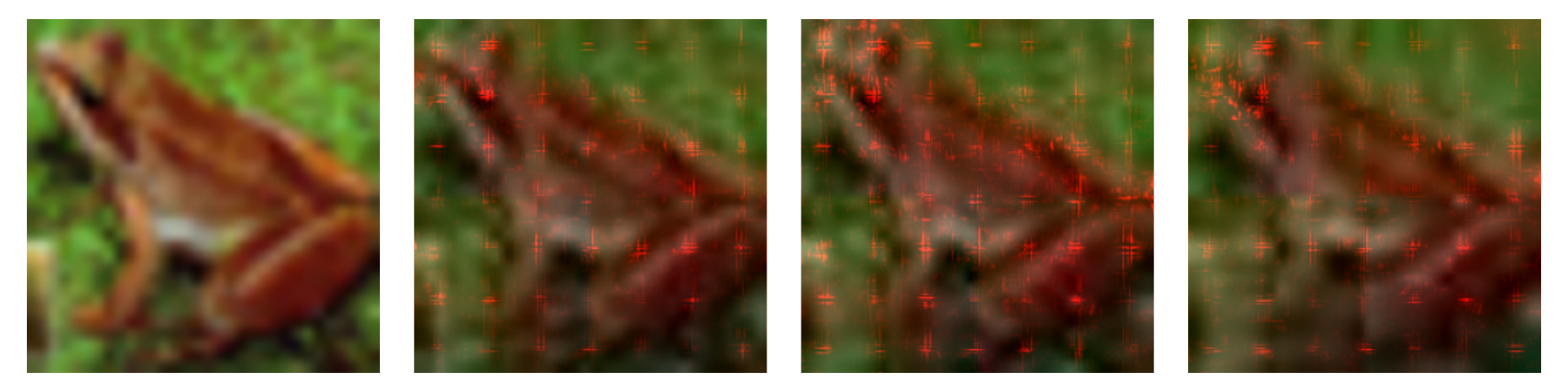}\\
			\caption{ViT negative}
		\end{subfigure}\hspace{1cm}
		\begin{subfigure}[h]{0.45\textwidth}
			\includegraphics[width=\linewidth ]{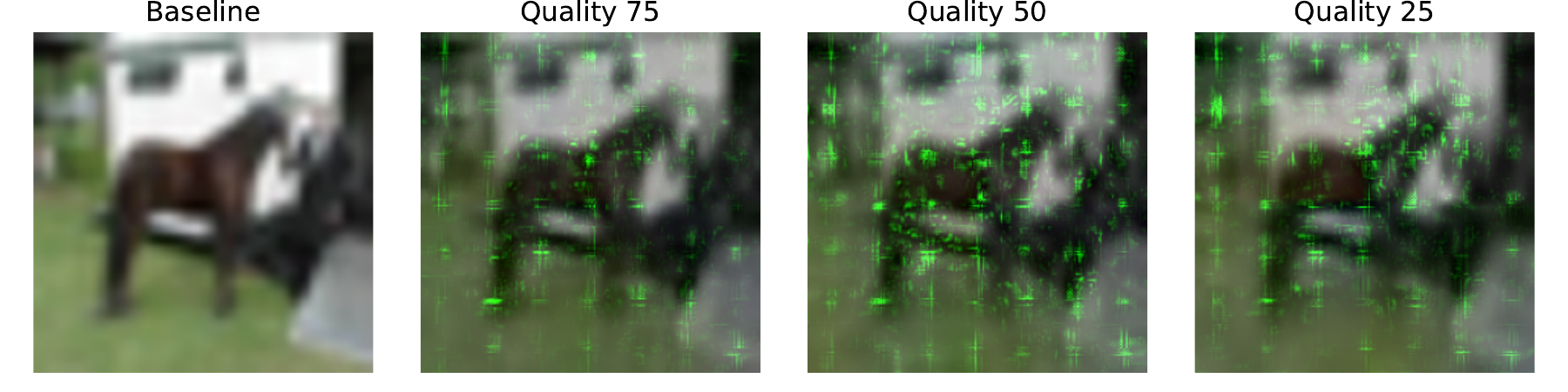}\\
			\includegraphics[width=\linewidth ]{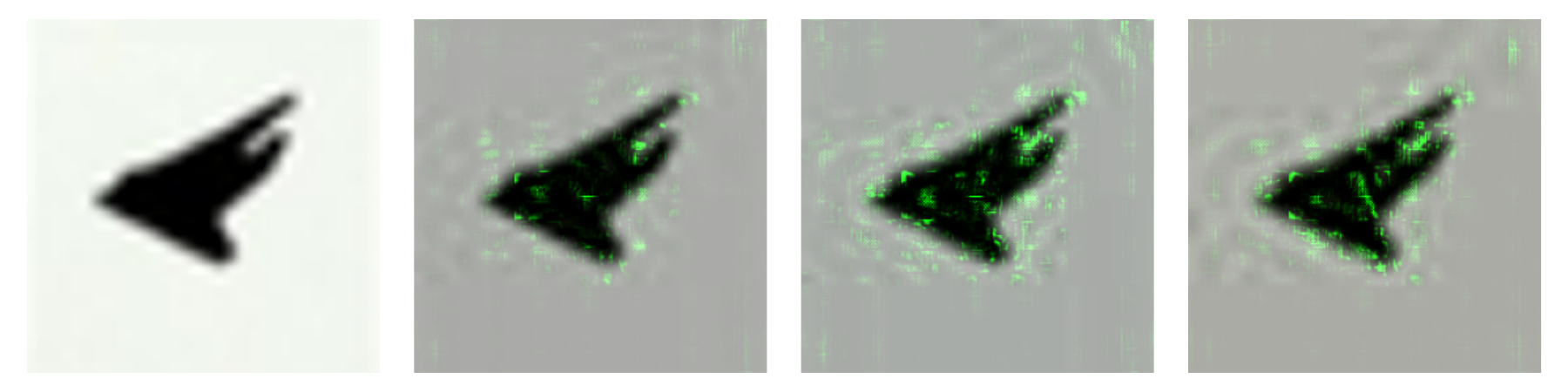}\\
			\includegraphics[width=\linewidth ]{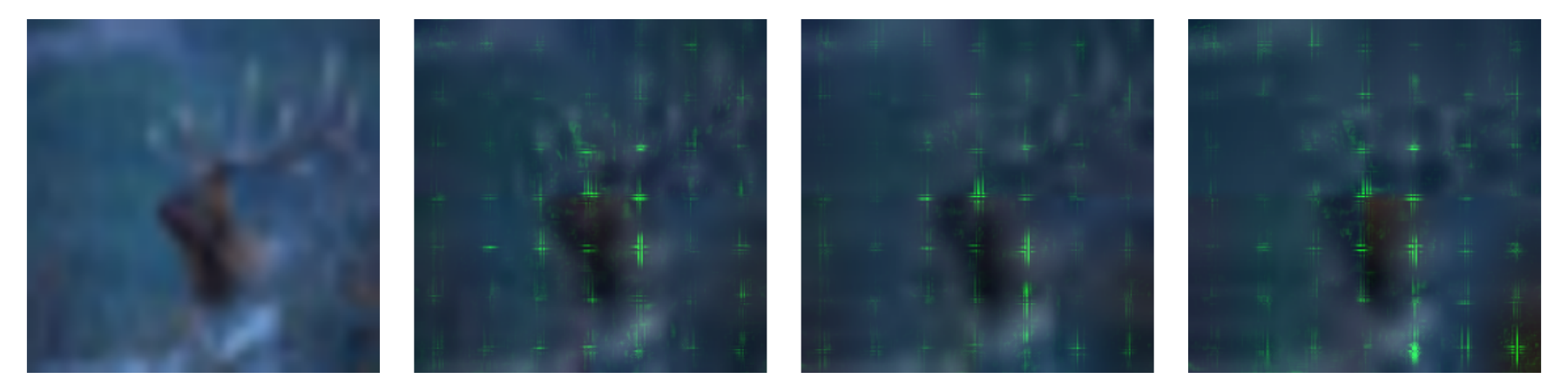}\\
			\includegraphics[width=\linewidth ]{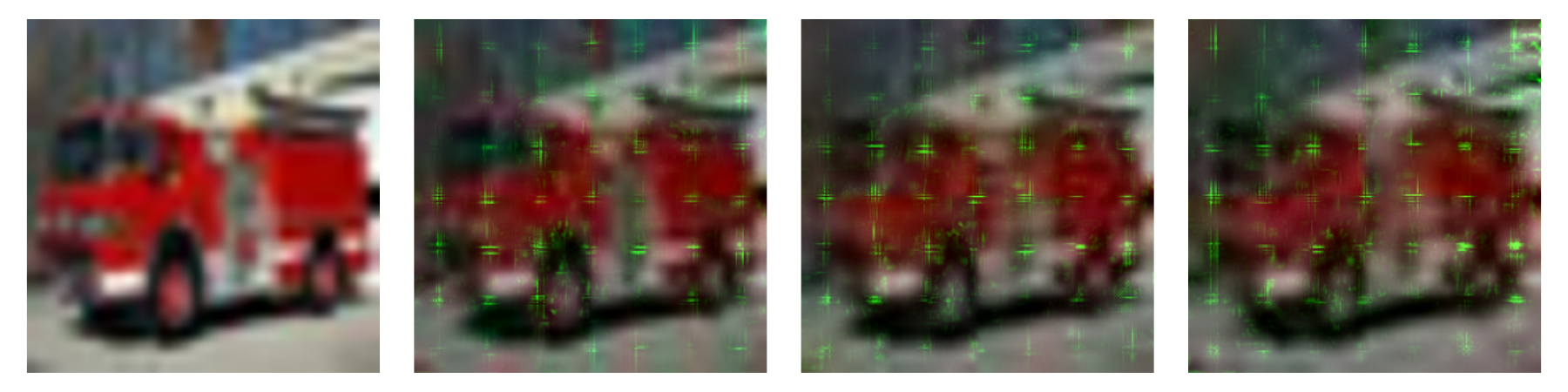}\\
			\includegraphics[width=\linewidth ]{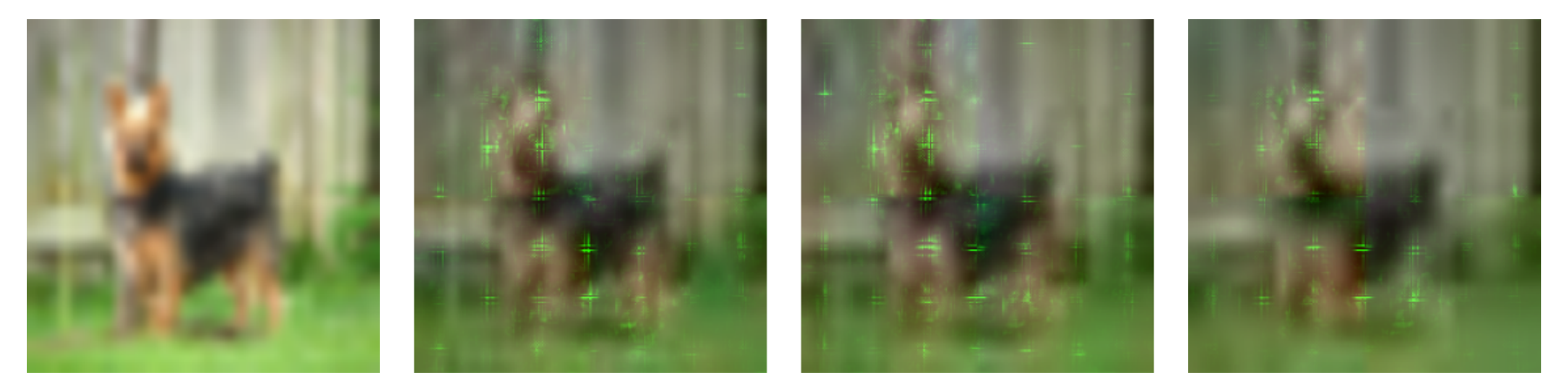}\\
			\includegraphics[width=\linewidth ]{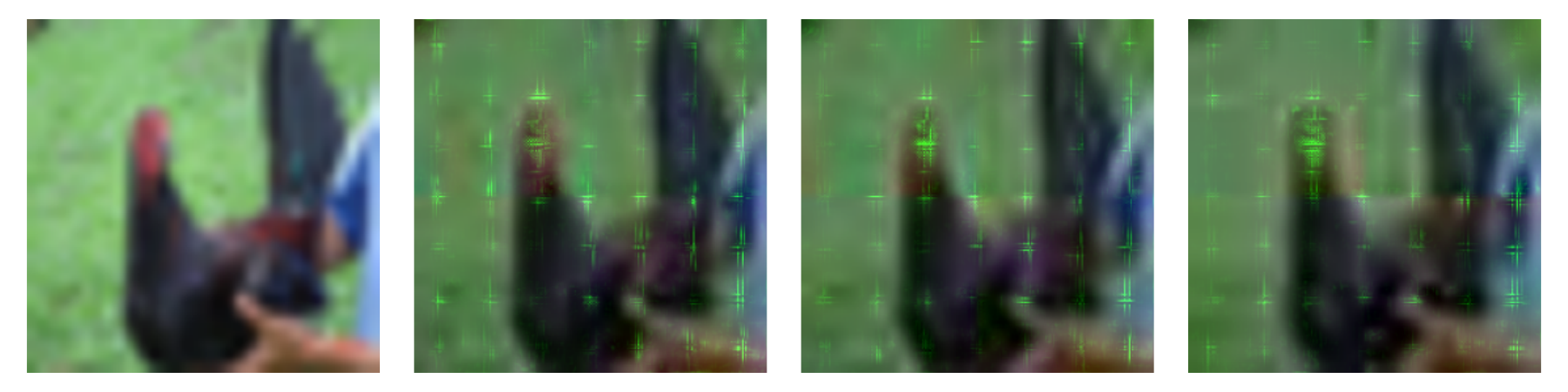}\\
			\includegraphics[width=\linewidth ]{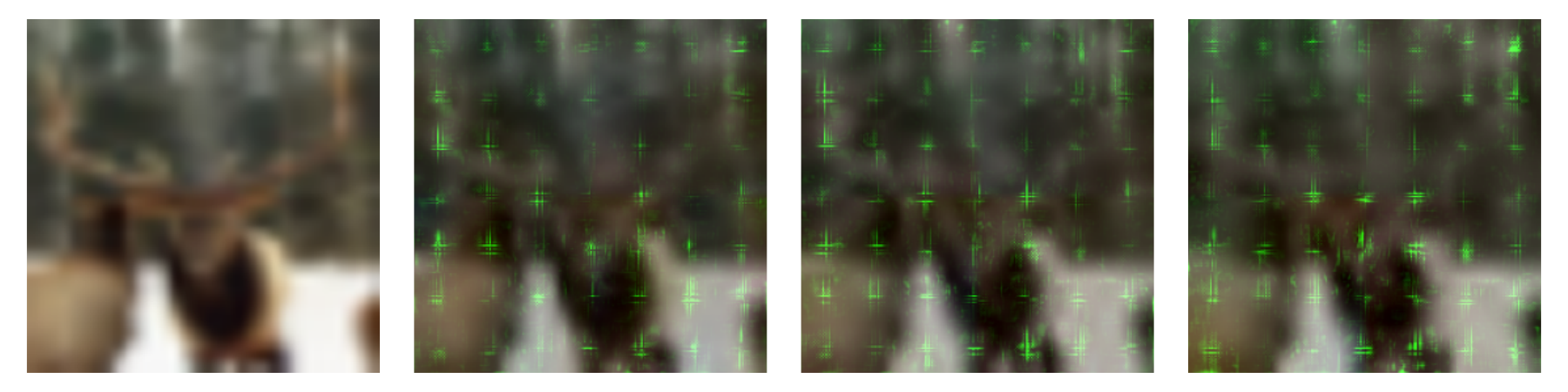}\\
			\includegraphics[width=\linewidth ]{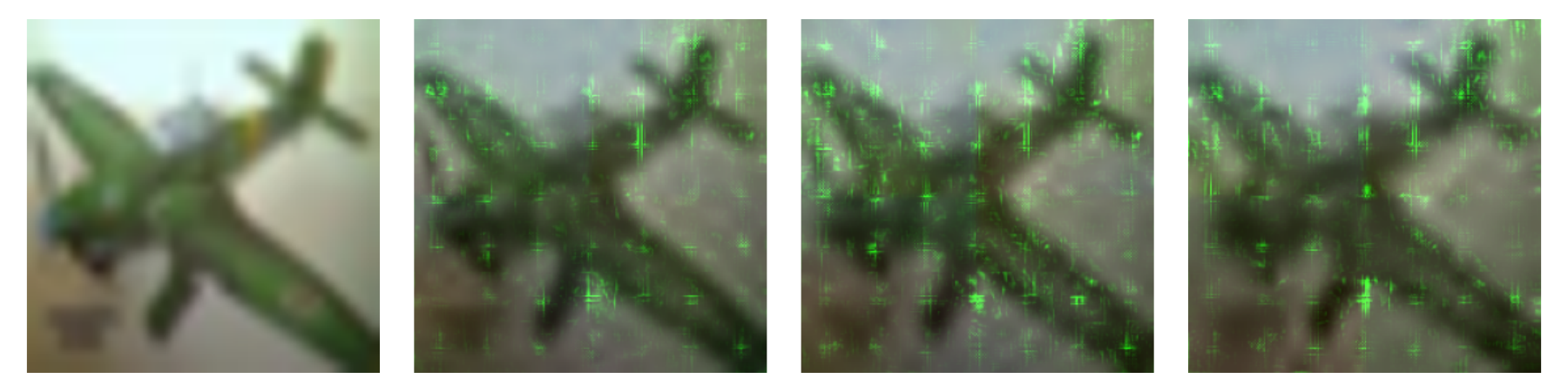}\\
			\includegraphics[width=\linewidth ]{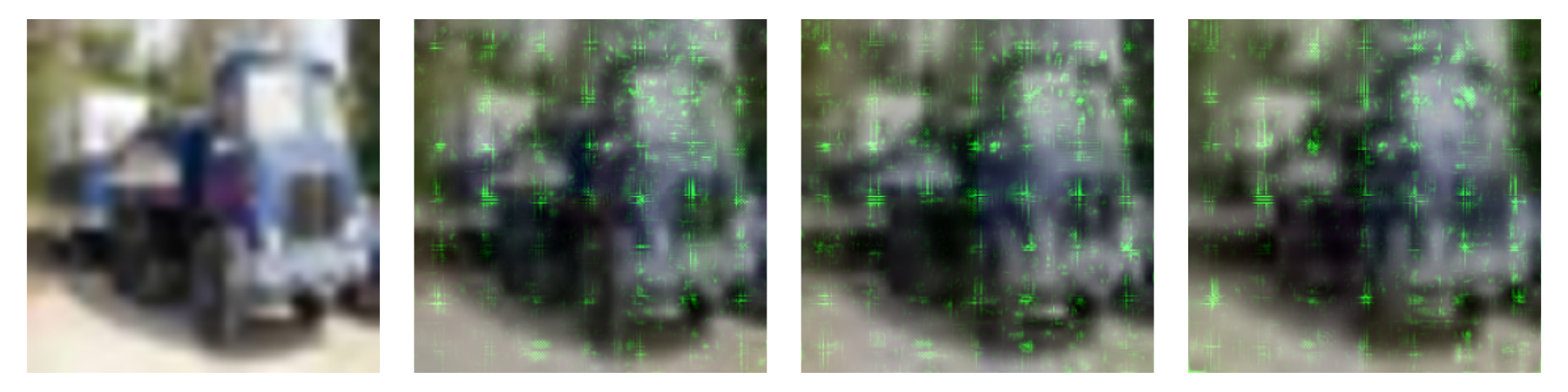}\\
			\includegraphics[width=\linewidth ]{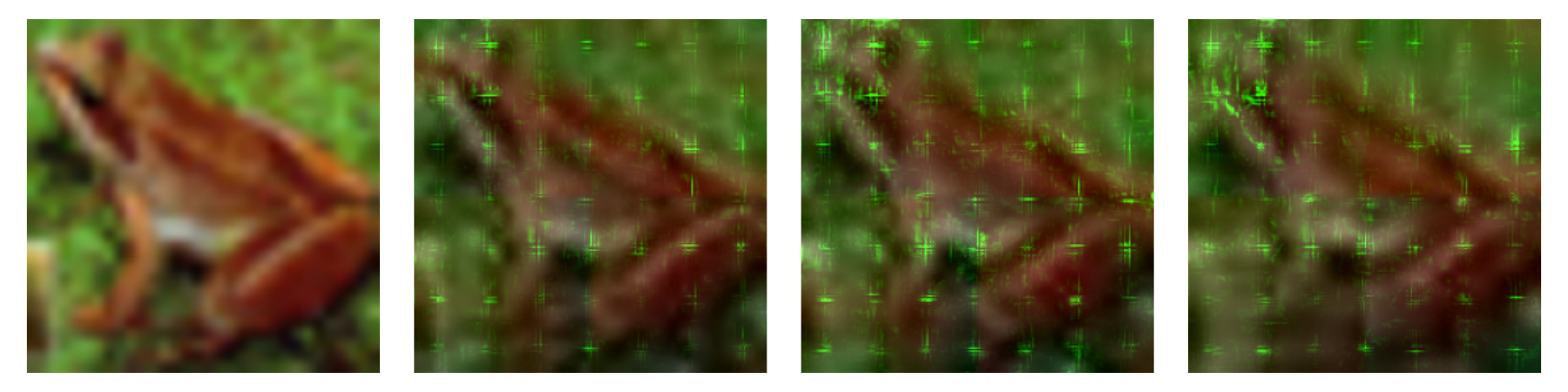}\\
			\caption{ViT positive}
		\end{subfigure} \\
		\caption{Examples on visualisation of integrated gradients for ViT-B/32.} \label{fig:ViT-B-32 supp cifar-10}
\end{figure}
\begin{table*}[!ht]
		\centering
		\scalebox{0.85}{%
		\begin{tabular}{lccc}
		\toprule
		      True label & Predicted label & Predicted score & IG \\
		     \midrule
		     horse & bird, airplane, airplane, airplane & 0.1187, 0.1197, 0.1149, 0.0742 & -0.0202, 0.0328, 0.4472 \\
             airplane & airplane, airplane, airplane, airplane & 0.7611, 0.6988, 0.5371, 0.8033 & 0.0906, 0.3691, -0.0326 \\
             deer  & airplane, airplane, airplane, airplane & 0.0397, 0.0361, 0.0321, 0.0351 & 0.0987, 0.2167, 0.1294 \\
             truck & airplane, airplane, cat, bird & 0.0690, 0.0294, 0.0407, 0.0273 & 0.8535, 0.5069, 0.9046 \\
             dog & airplane, airplane, airplane, bird & 0.0877, 0.1194, 0.1259, 0.1060 & -0.3258, -0.3718, -0.1809 \\
             bird & bird, airplane, bird, airplane & 0.2332, 0.1935, 0.2175, 0.1925 & 0.1676, 0.0807, 0.2077 \\
             deer & bird, cat, bird, bird & 0.0673, 0.0523, 0.0590, 0.0634 & 0.2476, 0.1415, 0.0646 \\
             airplane & airplane, bird, airplane, bird & 0.6400, 0.2479, 0.3366, 0.2096 & 0.9459, 0.6292, 1.0857 \\
             truck & airplane, airplane, airplane, airplane & 0.0531, 0.0349, 0.0417, 0.0376 & 0.4420, 0.2453, 0.3690 \\
             frog & bird, bird, bird, bird & 0.0538, 0.0835, 0.1056, 0.0686 & -0.4301, -0.6973, -0.2598 \\

		     \bottomrule
		\end{tabular}
		}
        \caption{Detailed information on the visual outputs in Figure \ref{fig:RN50 supp cifar-10} and \ref{fig:Supp RN50 both CIFAR}. }\label{tbl:Supp CIFAR-10 RN50 comparisons}
\end{table*}
\begin{table*}[!ht]
		\centering
		\scalebox{0.85}{%
		\begin{tabular}{lccc}
		\toprule
		      True label & Predicted label & Predicted score & IG \\
		     \midrule
		     horse & horse, horse, horse,  airplane & 0.3562, 0.3369, 0.2686, 0.1544 & 0.0592, 0.2884, 0.8333 \\
             airplane & bird, bird, airplane, airplane & 0.2443, 0.3639, 0.3562, 0.3412 & -0.3993, -0.3754, -0.3324 \\
             deer  & bird, airplane, cat, airplane & 0.0558, 0.0552, 0.0419, 0.0467 & 0.0156, 0.2918, 0.1769 \\
             truck & truck, truck, automobile, automobile & 0.7724, 0.4120, 0.3104, 0.1449 & 0.6323, 0.9206, 1.6725 \\
             dog & dog, dog, cat, cat & 0.4909, 0.3269, 0.1232, 0.1679 & 0.4164, 1.3838, 1.0830 \\
             bird & horse, bird, bird, airplane & 0.1973, 0.2273, 0.2110, 0.1442 & -0.1410, -0.0749, 0.3141 \\
             deer & bird, airplane, airplane, cat & 0.0759, 0.0273, 0.0303, 0.0316 & 1.0250, 0.9265, 0.8778 \\
             airplane & airplane, airplane, airplane, airplane & 0.8823, 0.7552, 0.5130, 0.4556 & 0.1556, 0.5430, 0.6607 \\
             truck & automobile, airplane, ship, airplane & 0.3672, 0.0703, 0.0441, 0.0512 & 1.6619, 2.1206, 1.9866 \\
             frog & bird, airplane, airplane, bird & 0.2131, 0.0864, 0.1178, 0.1230 & 0.9112, 0.5917, 0.5477 \\
		     \bottomrule
		\end{tabular}
		}
        \caption{Detailed information on the visual outputs in Figure \ref{fig:ViT-B-32 supp cifar-10} and \ref{fig:Supp ViT both CIFAR}. }\label{tbl:Supp CIFAR-10 ViT comparisons}
\end{table*}
\section{More examples of Integrated Gradients on STL-10}
We provide more examples of visualised integrated gradients on STL-10.
\begin{figure}[ht]
		\centering
		%
		%
		\begin{subfigure}[h]{0.45\textwidth}
			\includegraphics[width=\linewidth ]{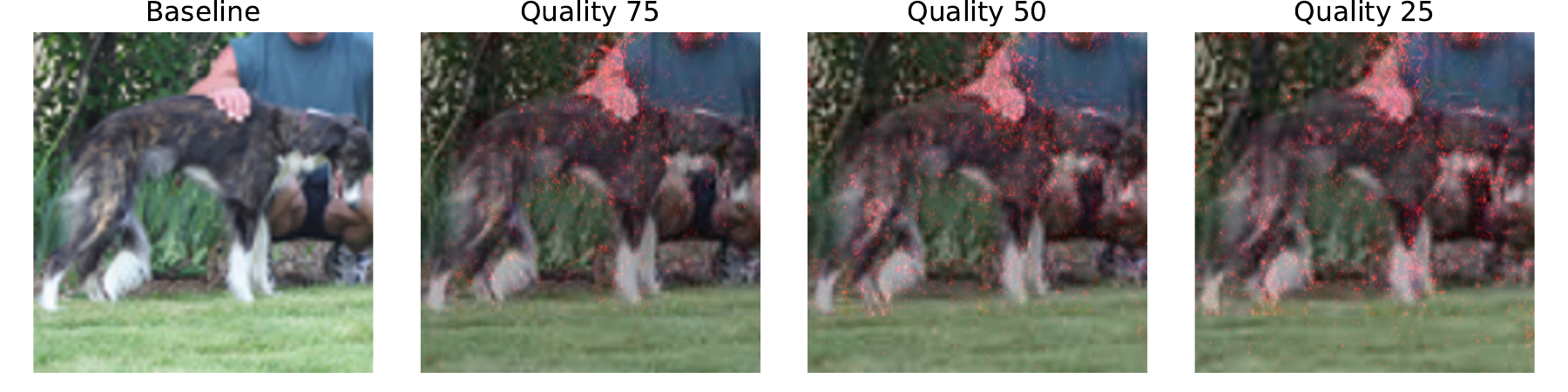}\\
			\includegraphics[width=\linewidth ]{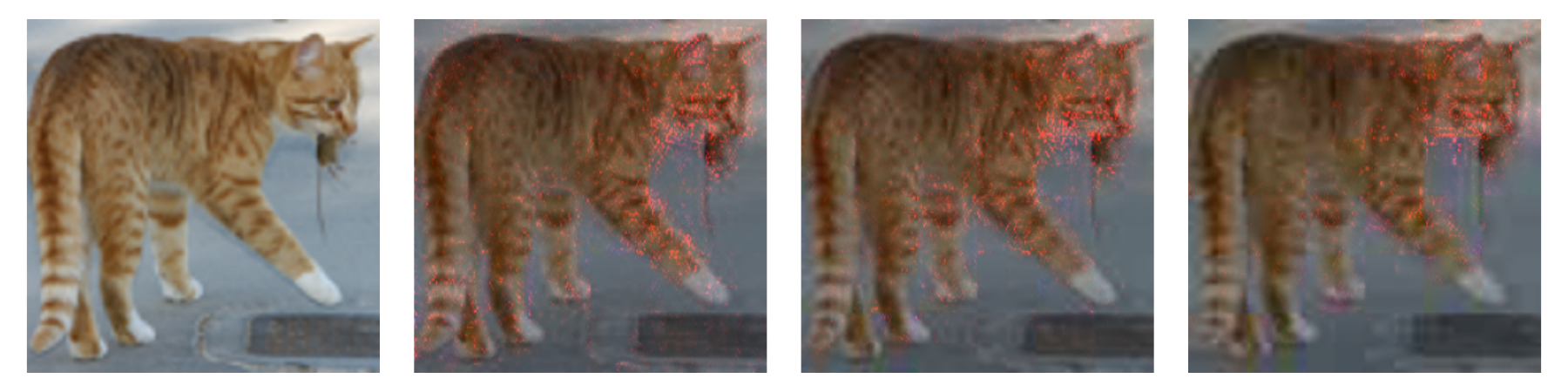}\\
			\includegraphics[width=\linewidth ]{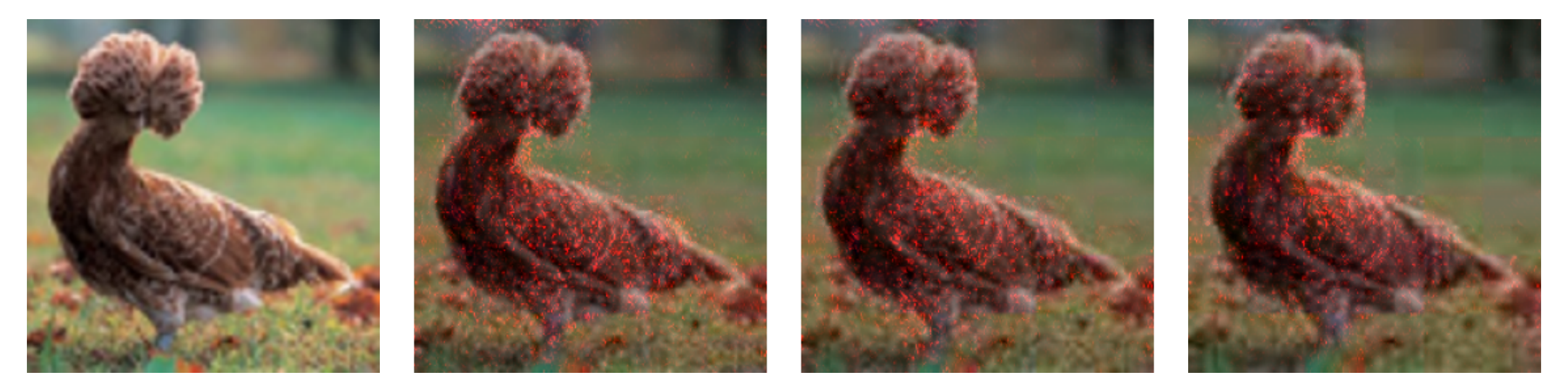}\\
			\includegraphics[width=\linewidth ]{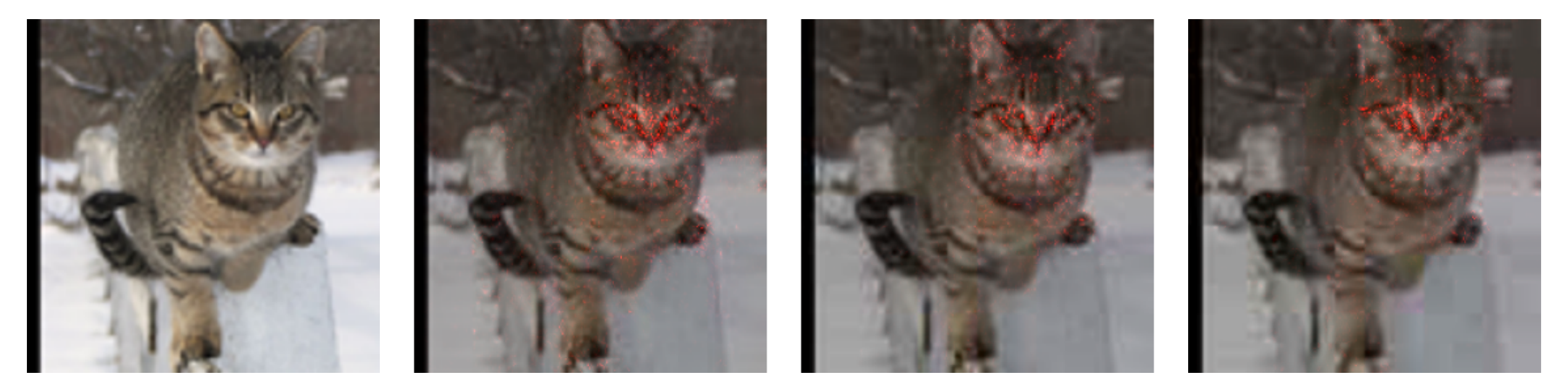}\\
			\includegraphics[width=\linewidth ]{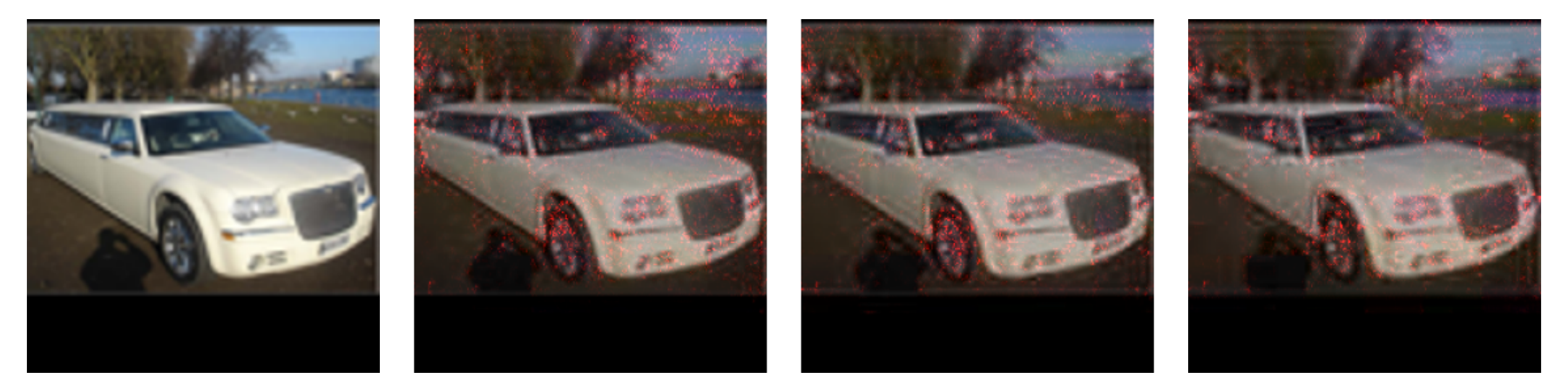}\\
			\includegraphics[width=\linewidth ]{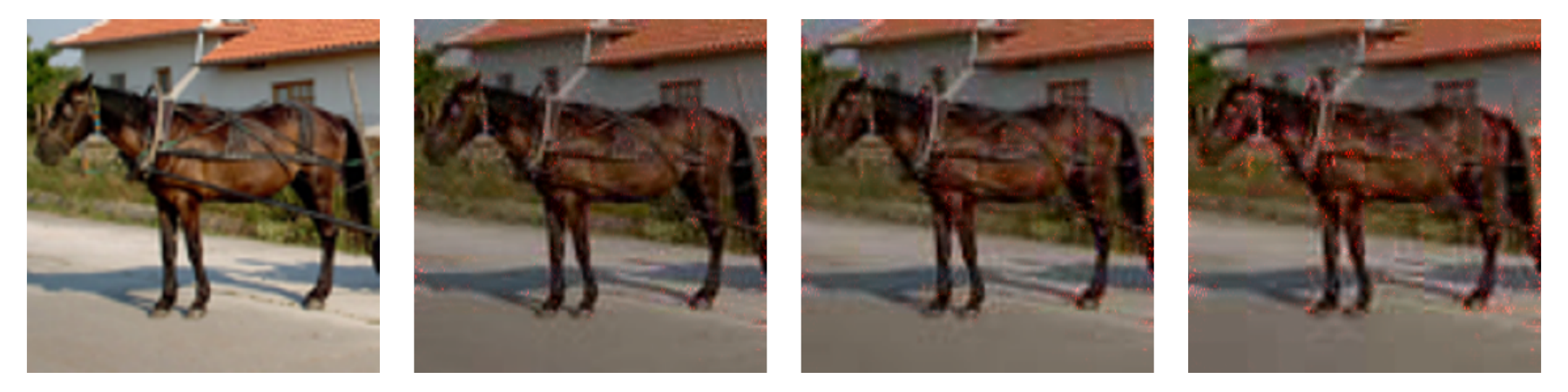}\\
			\includegraphics[width=\linewidth ]{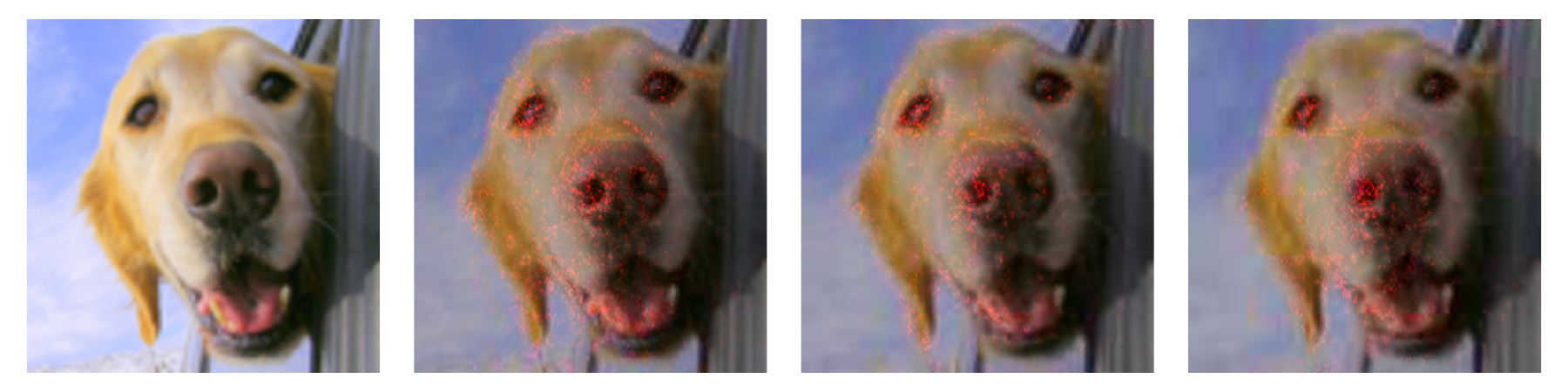}\\
			\includegraphics[width=\linewidth ]{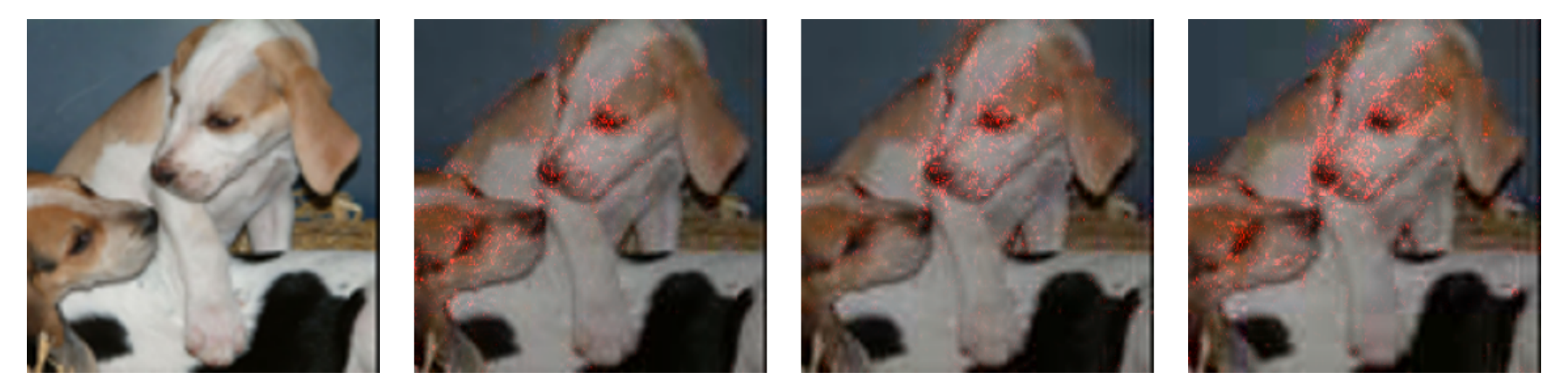}\\
			\includegraphics[width=\linewidth ]{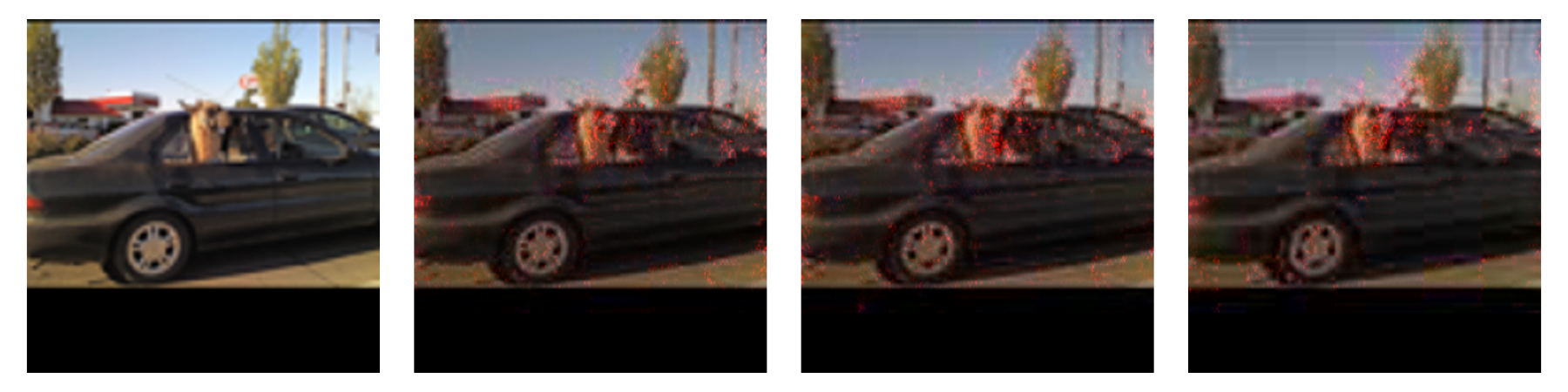}\\
			\includegraphics[width=\linewidth ]{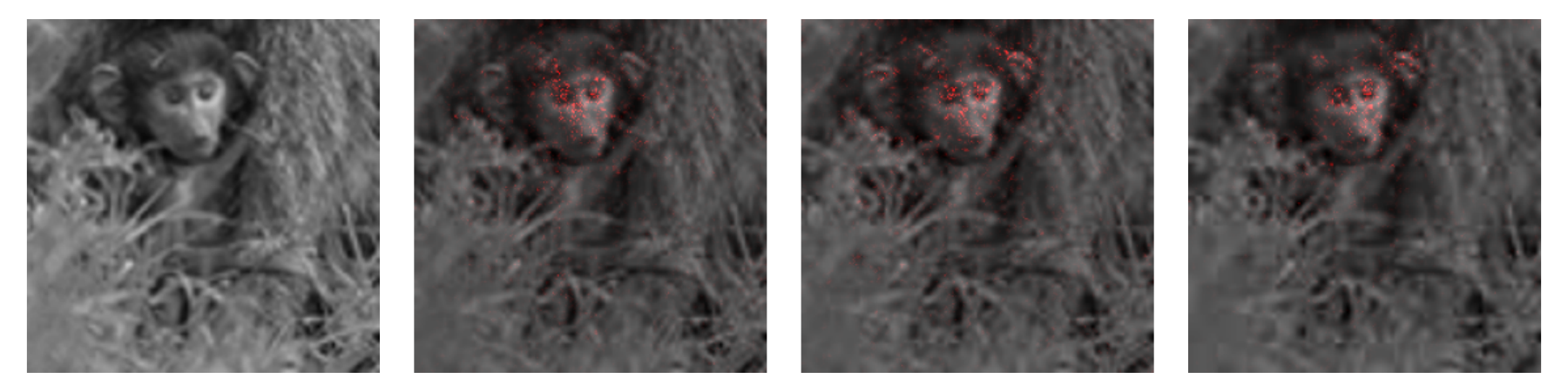}\\
			\caption{ResNet-50 negative}
		\end{subfigure}\hspace{1cm}
		\begin{subfigure}[h]{0.45\textwidth}
			\includegraphics[width=\linewidth ]{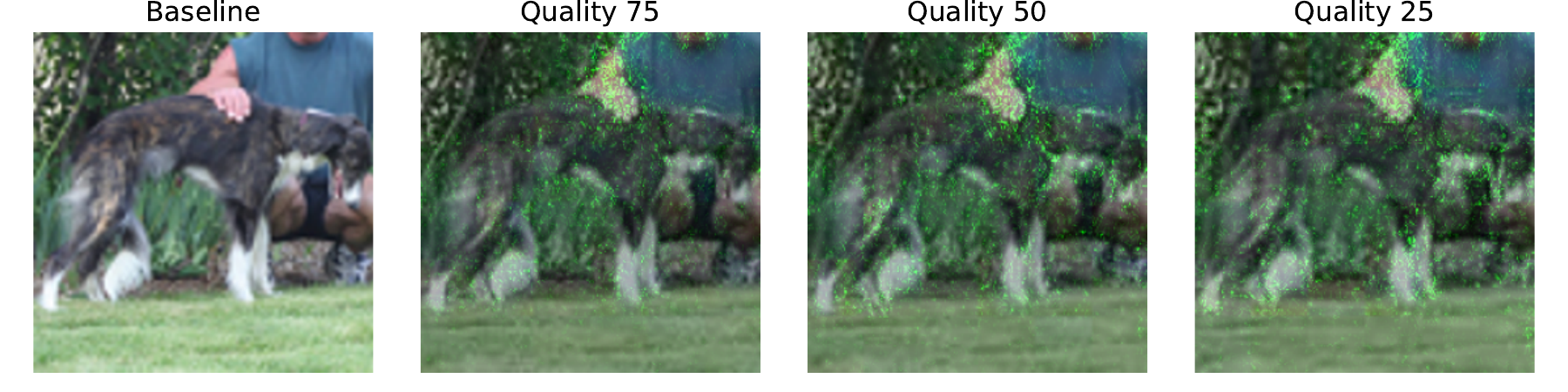}\\
			\includegraphics[width=\linewidth ]{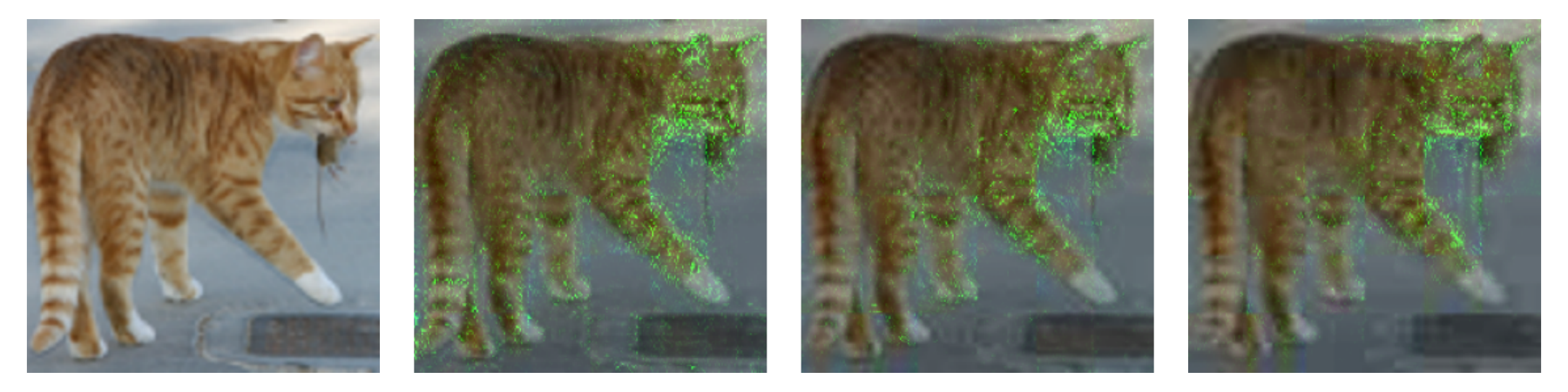}\\
			\includegraphics[width=\linewidth ]{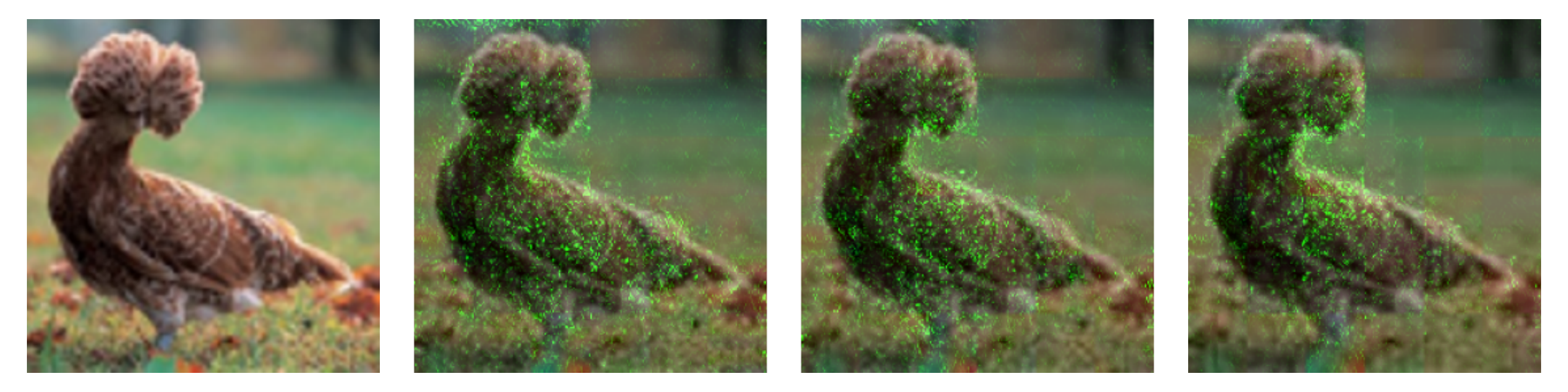}\\
			\includegraphics[width=\linewidth ]{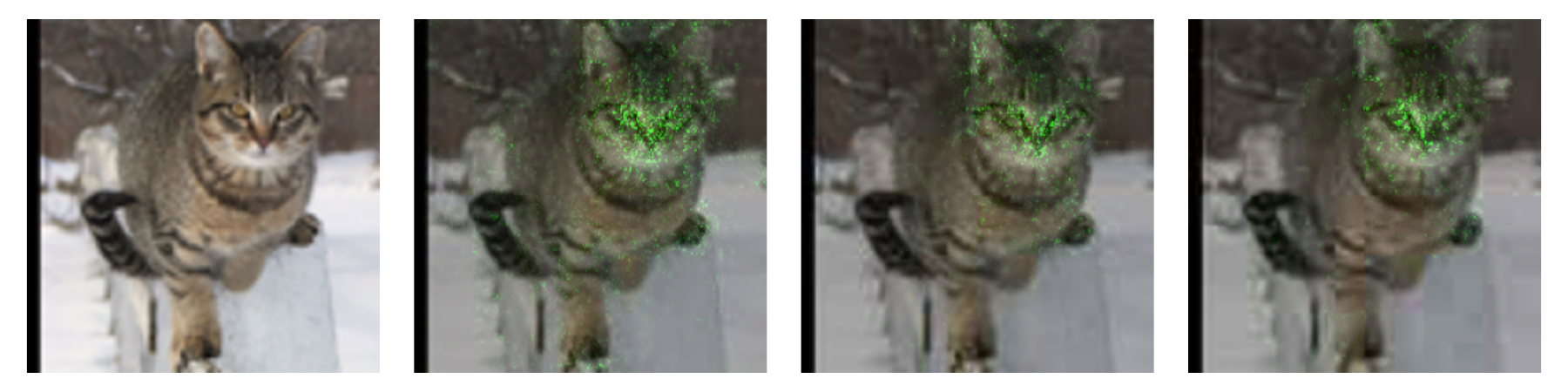}\\
			\includegraphics[width=\linewidth ]{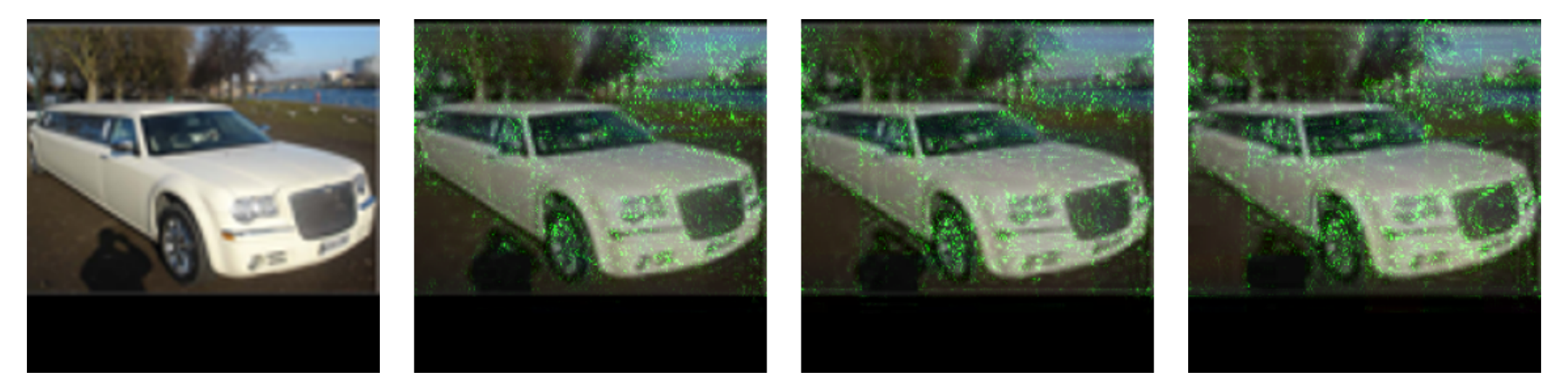}\\
			\includegraphics[width=\linewidth ]{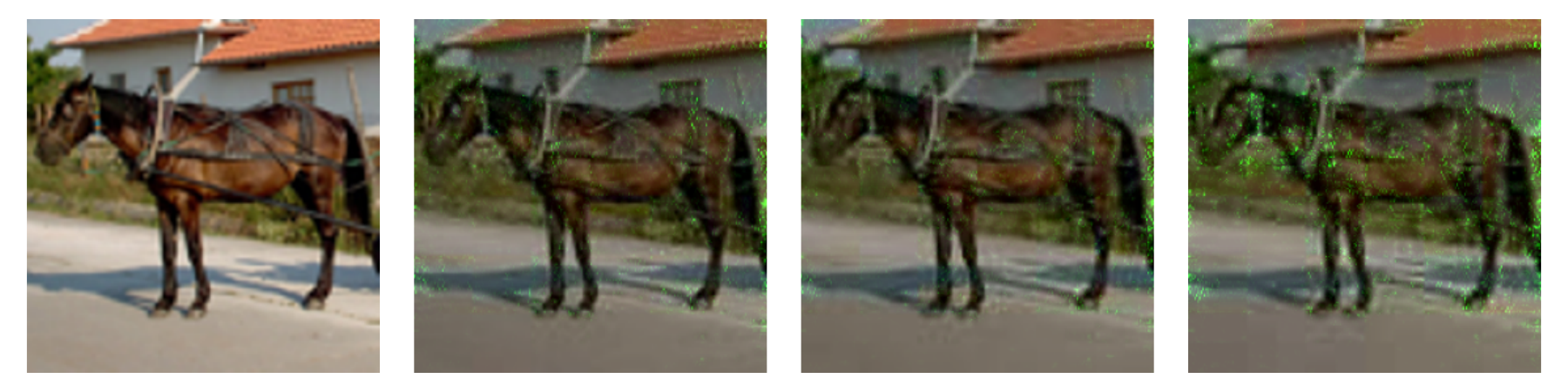}\\
			\includegraphics[width=\linewidth ]{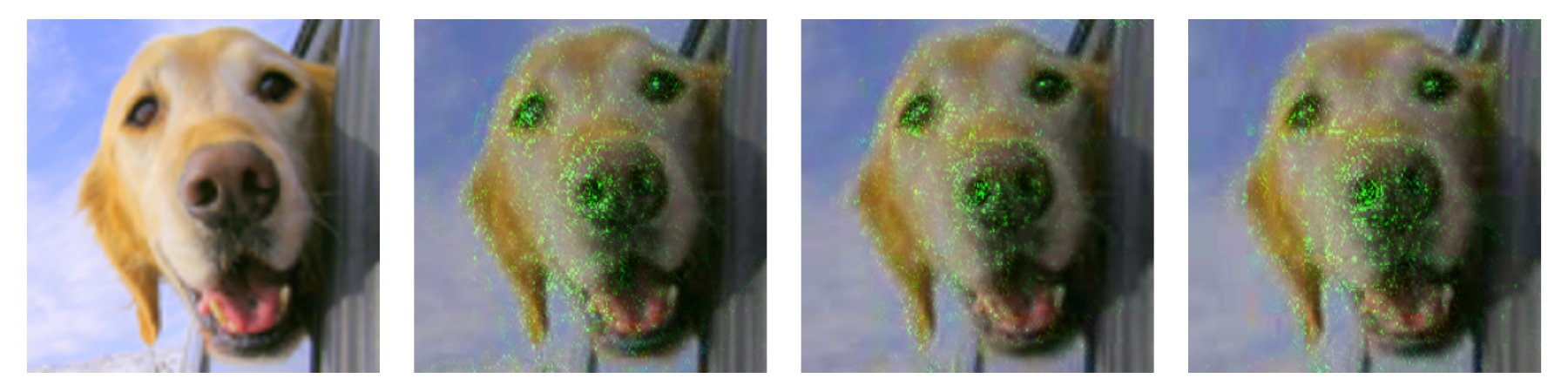}\\
			\includegraphics[width=\linewidth ]{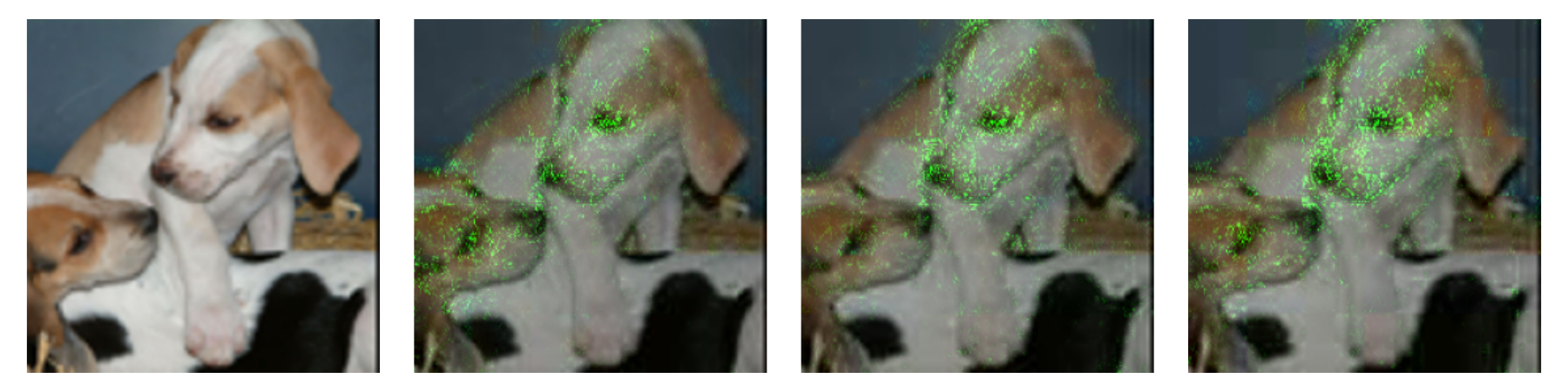}\\
			\includegraphics[width=\linewidth ]{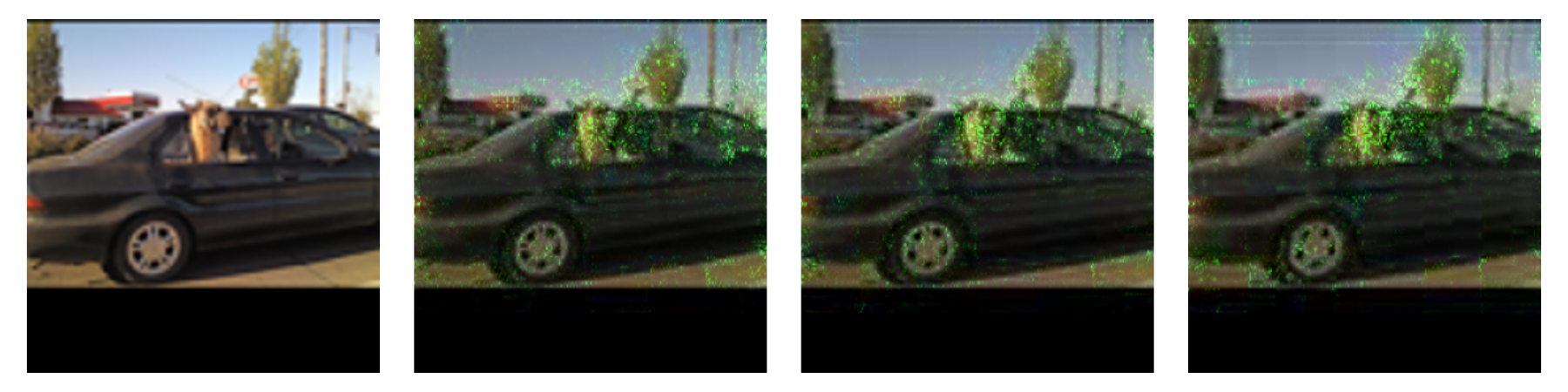}\\
			\includegraphics[width=\linewidth ]{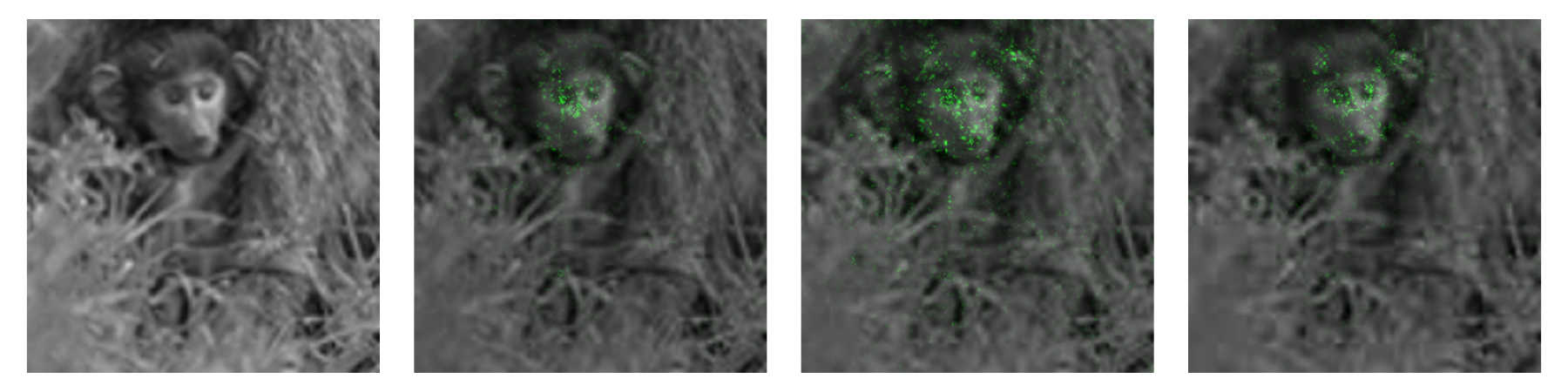}\\
			\caption{ResNet-50 positive}
		\end{subfigure} \\
    \caption{Examples on visualisation of integrated gradients for ResNet-50.}\label{fig:RN50 supp stl-10}
\end{figure}
\begin{figure}[ht]
		\centering
		%
		%
		\begin{subfigure}[h]{0.45\textwidth}
			\includegraphics[width=\linewidth]{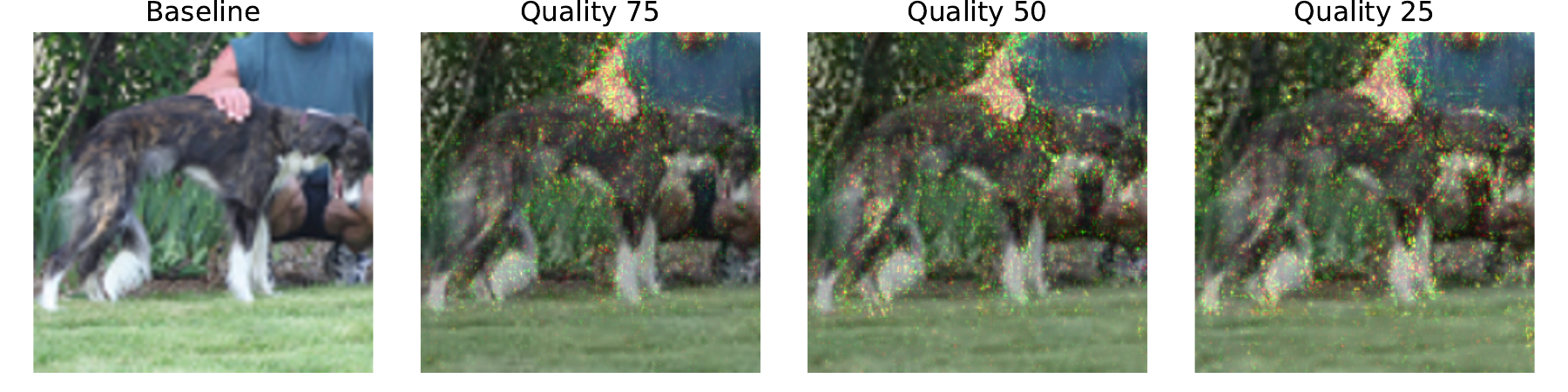}\\
			\includegraphics[width=\linewidth]{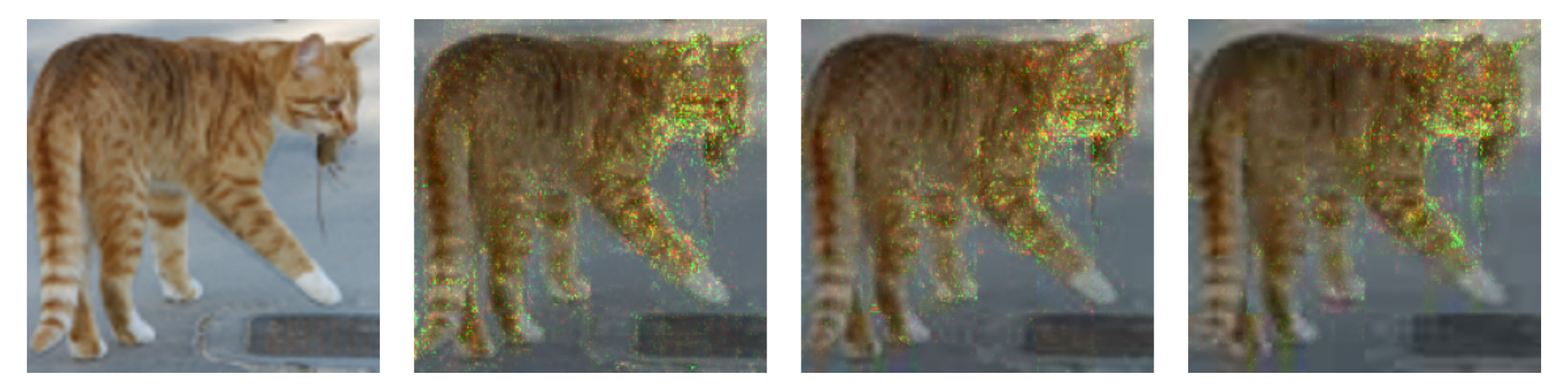}\\
			\includegraphics[width=\linewidth]{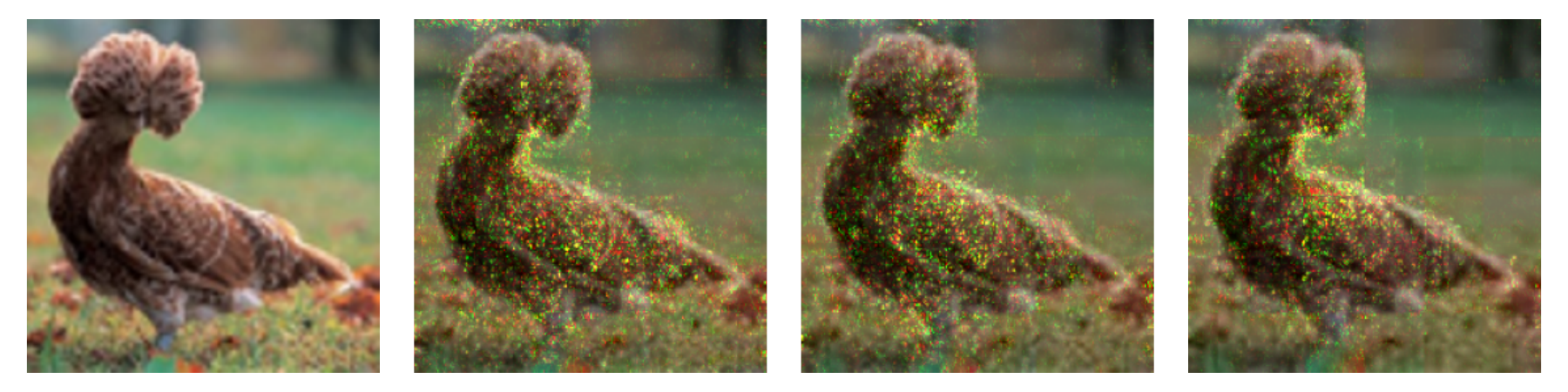}\\
			\includegraphics[width=\linewidth]{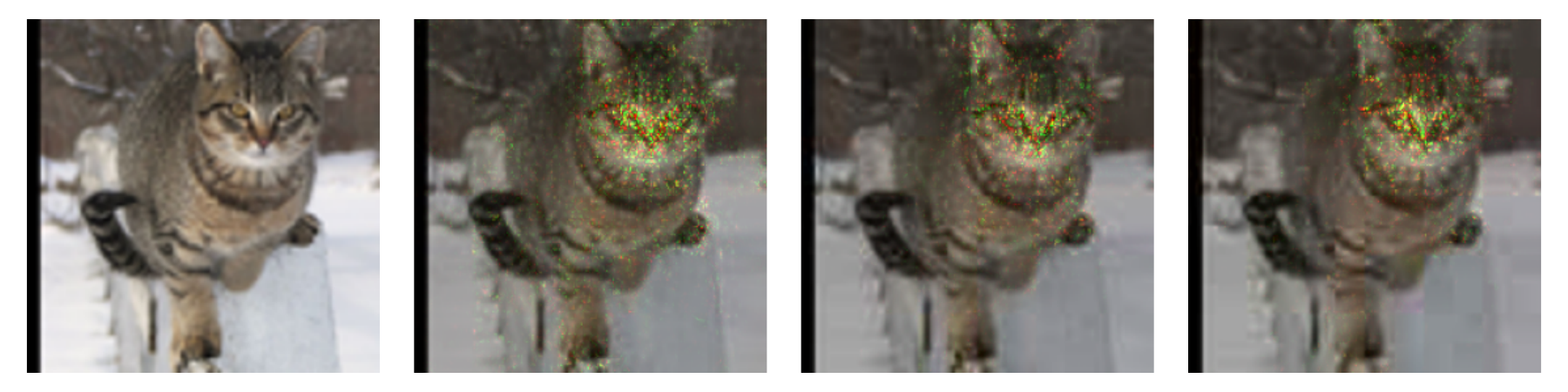}\\
			\includegraphics[width=\linewidth]{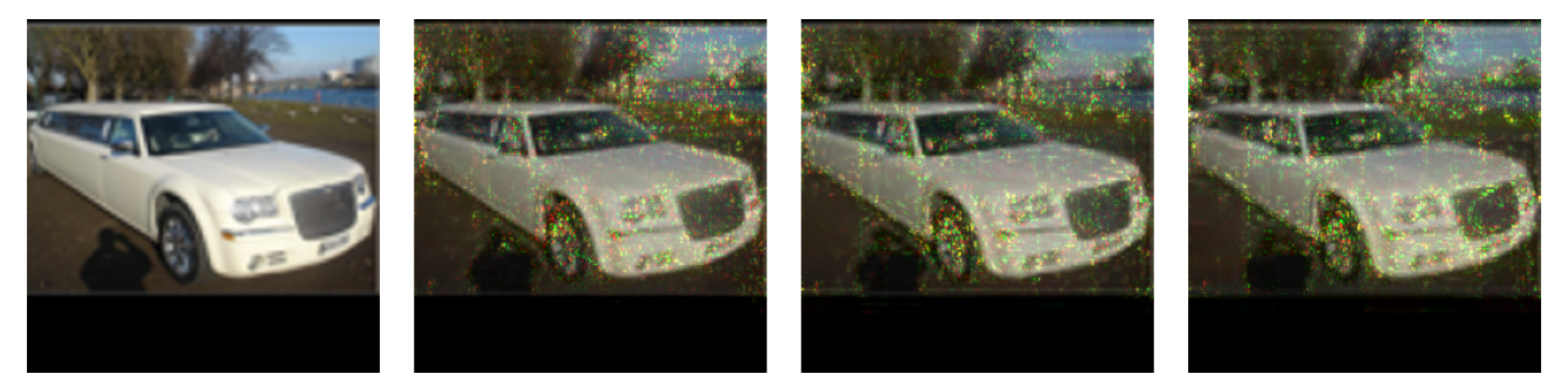}\\
			\includegraphics[width=\linewidth]{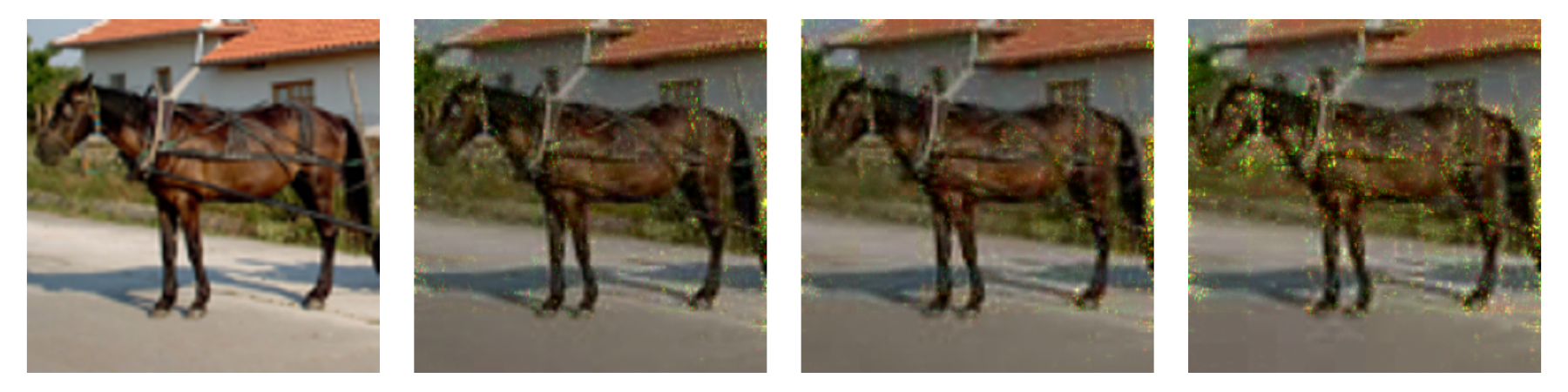}\\
			\includegraphics[width=\linewidth]{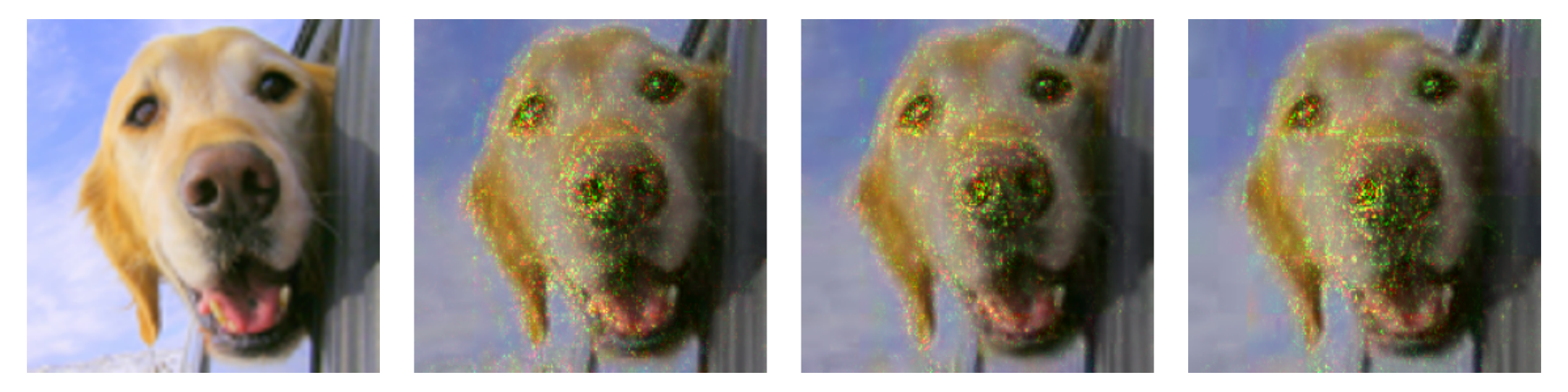}\\
			\includegraphics[width=\linewidth]{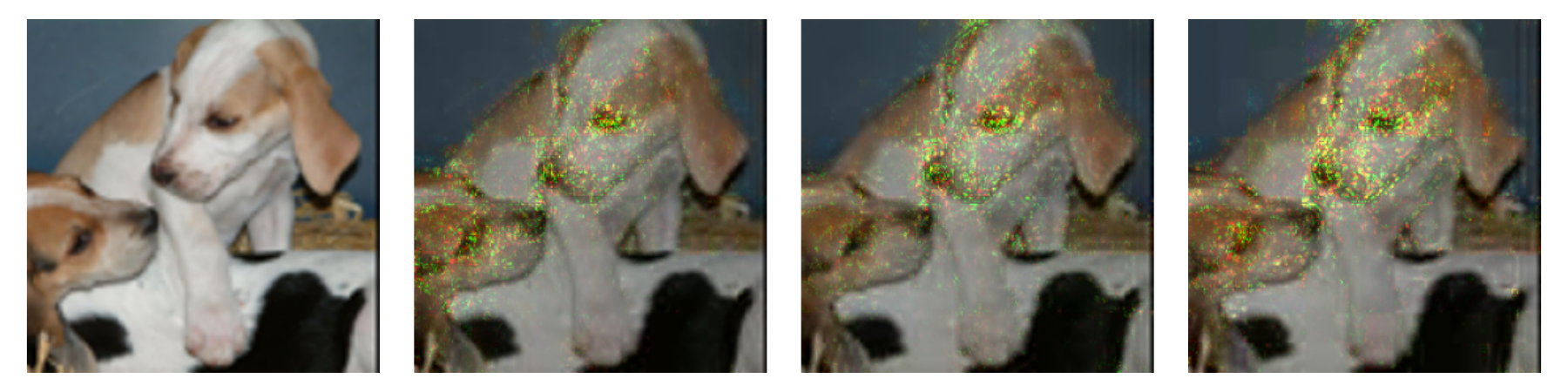}\\
			\includegraphics[width=\linewidth]{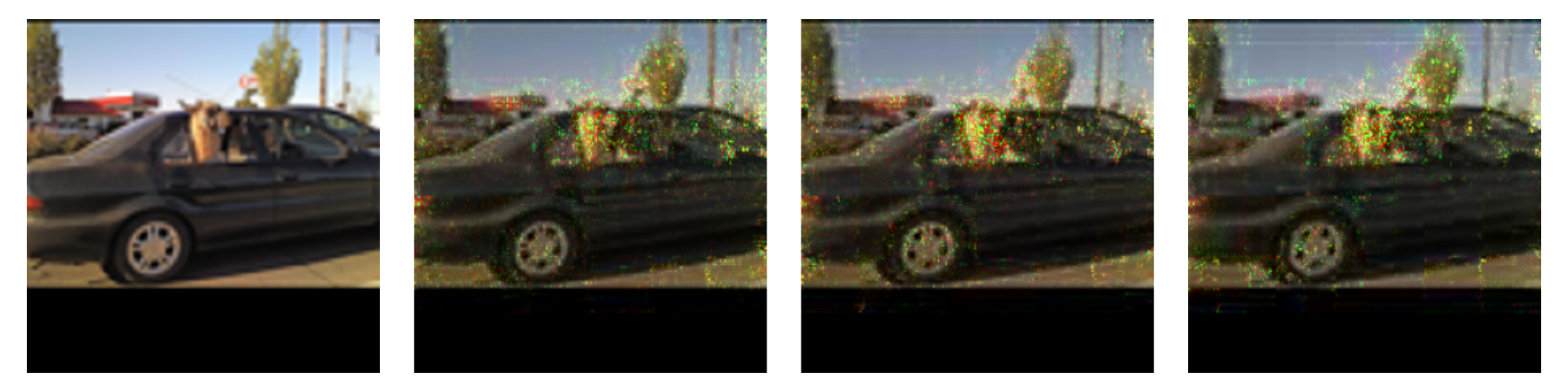}\\
			\includegraphics[width=\linewidth]{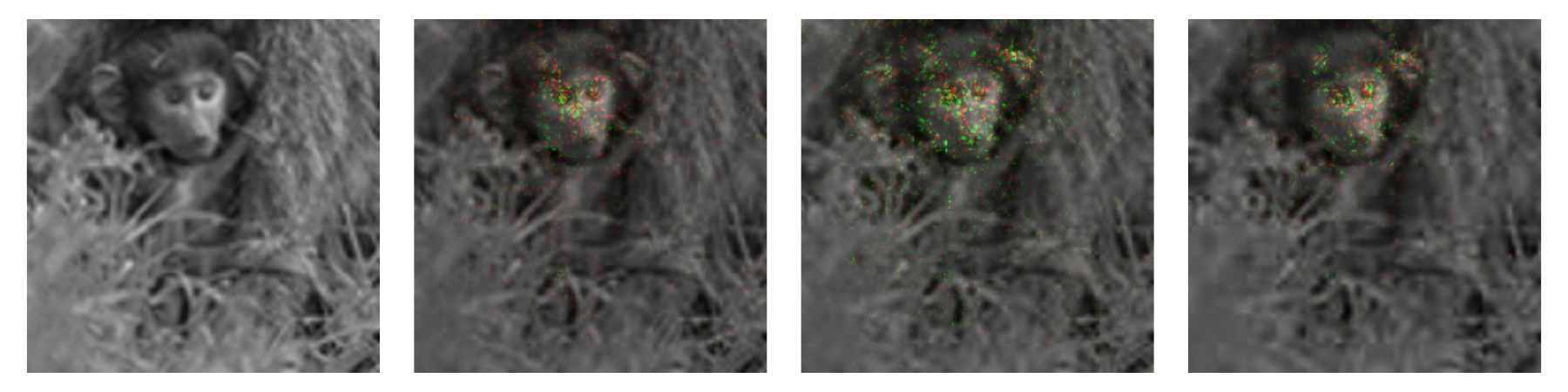}\\
			\caption{ResNet-50 both}\label{fig:Supp RN50 both stl-10}
		\end{subfigure}\hspace{1cm}
		\begin{subfigure}[h]{0.45\textwidth}
			\includegraphics[width=\linewidth]{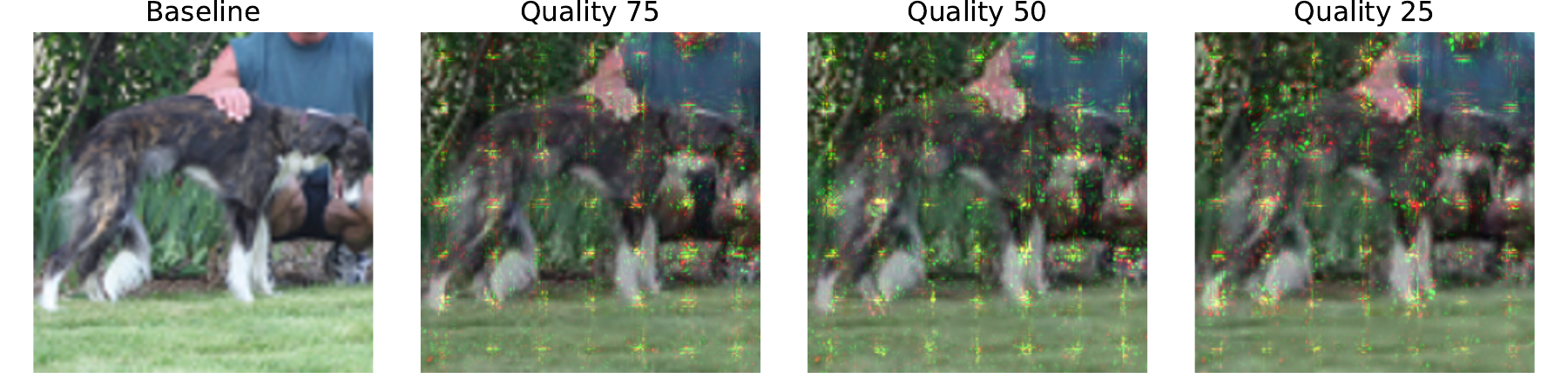}\\
			\includegraphics[width=\linewidth]{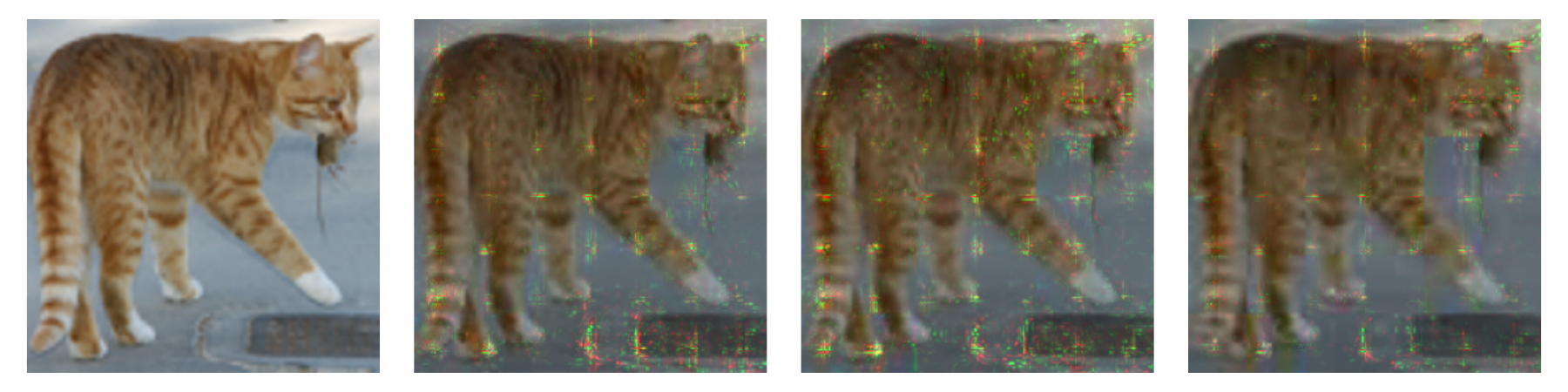}\\
			\includegraphics[width=\linewidth]{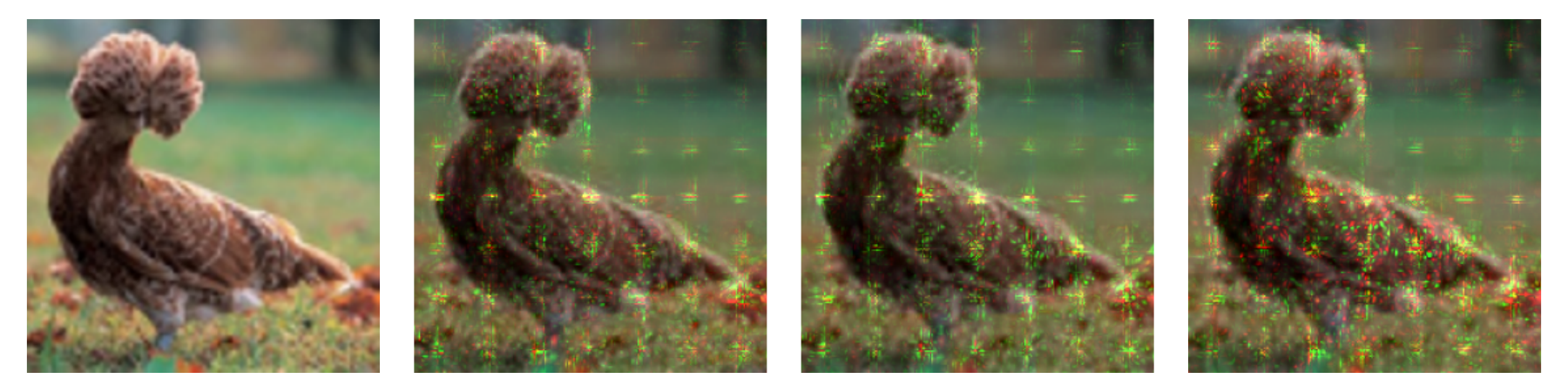}\\
			\includegraphics[width=\linewidth]{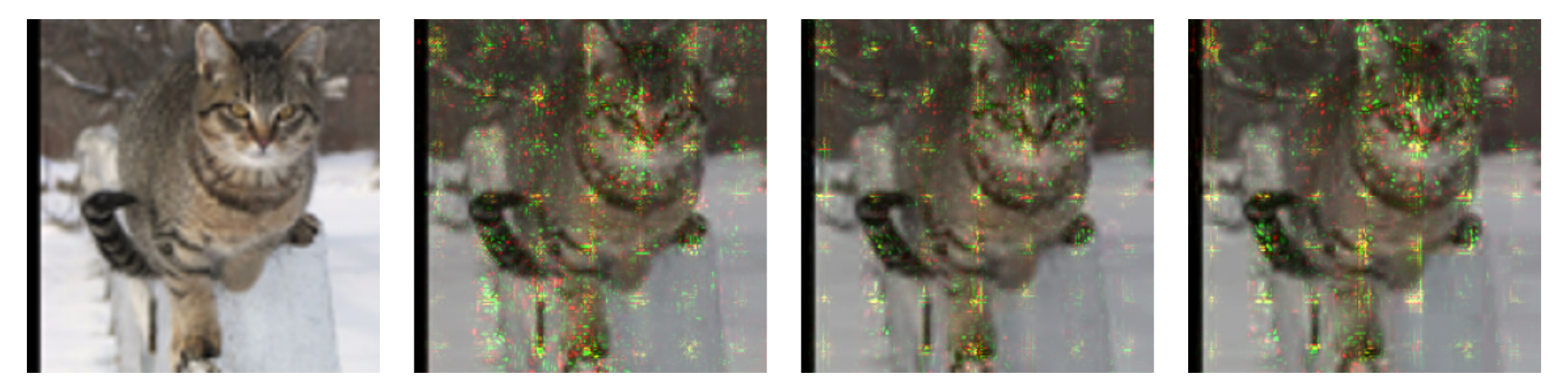}\\
			\includegraphics[width=\linewidth]{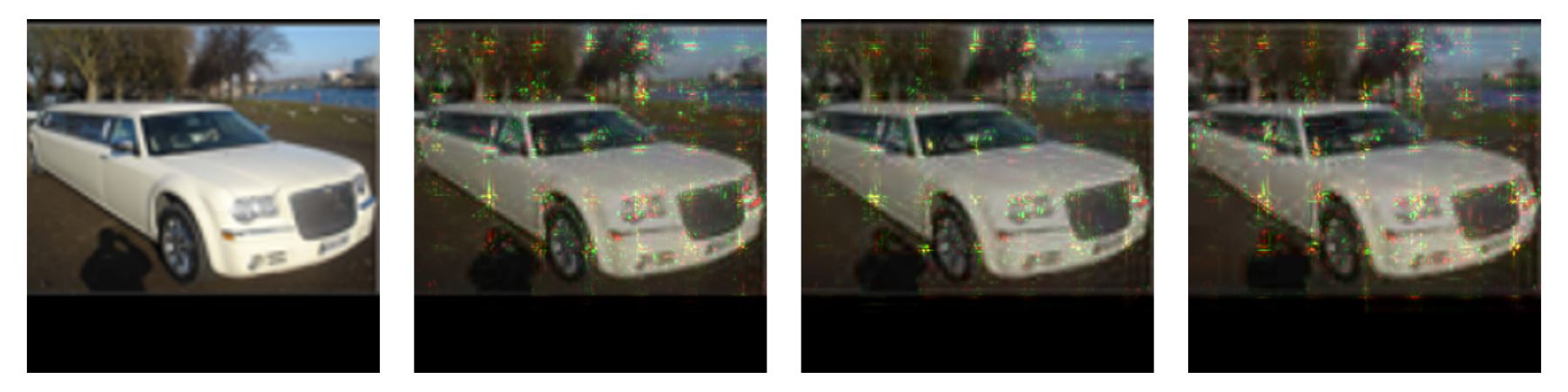}\\
			\includegraphics[width=\linewidth]{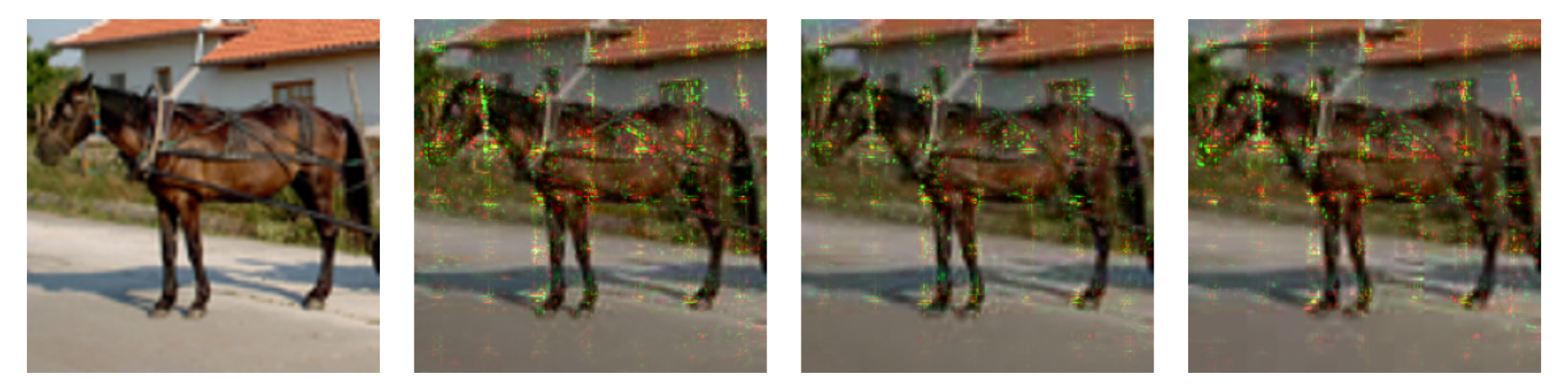}\\
			\includegraphics[width=\linewidth]{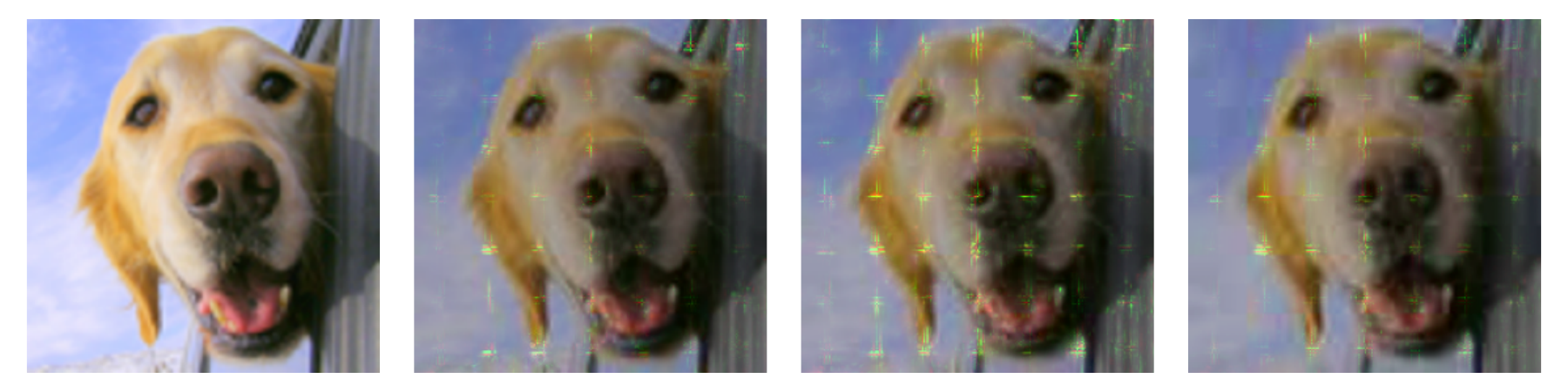}\\
			\includegraphics[width=\linewidth]{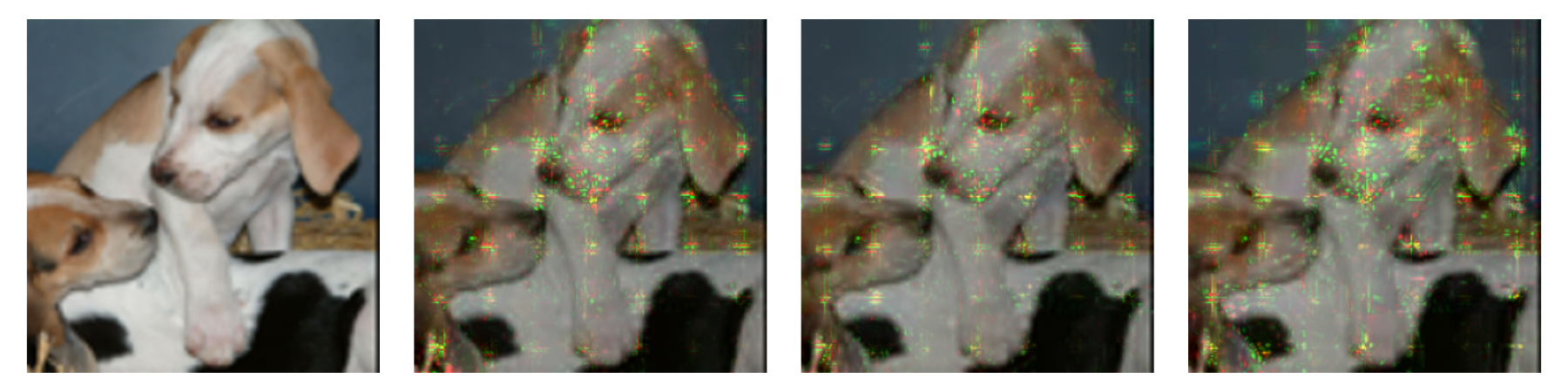}\\
			\includegraphics[width=\linewidth]{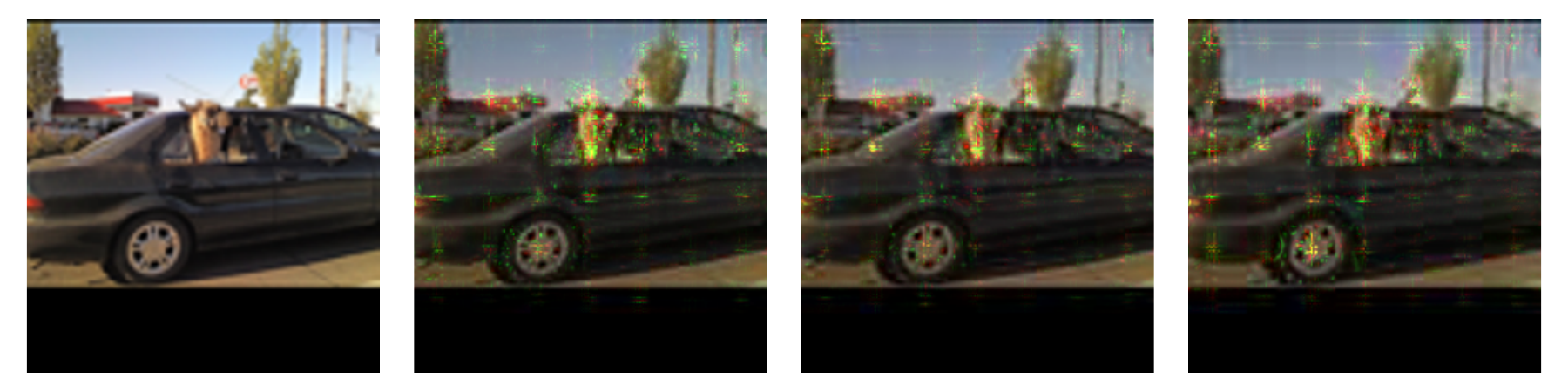}\\
			\includegraphics[width=\linewidth]{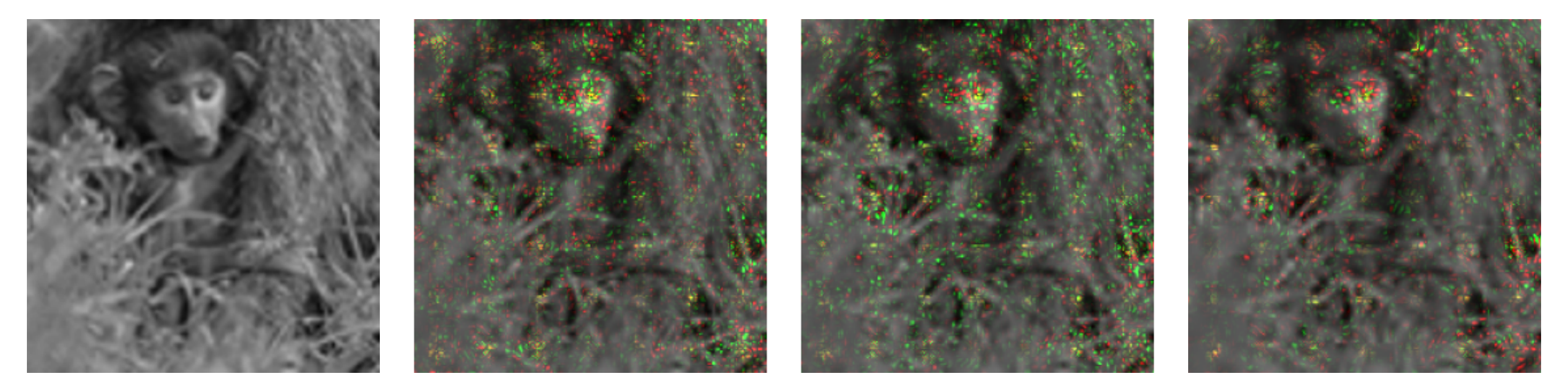}\\
			\caption{ViT both}\label{fig:Supp ViT both stl-10}
		\end{subfigure} \\
		\caption{Examples on visualisation of integrated gradients for ResNet-50 and ViT-B/32.}\label{fig:RN and ViT both supp stl-10}
\end{figure}
\begin{figure}[ht]
		\centering
		%
		%
		\begin{subfigure}[h]{0.45\textwidth}
			\includegraphics[width=\linewidth ]{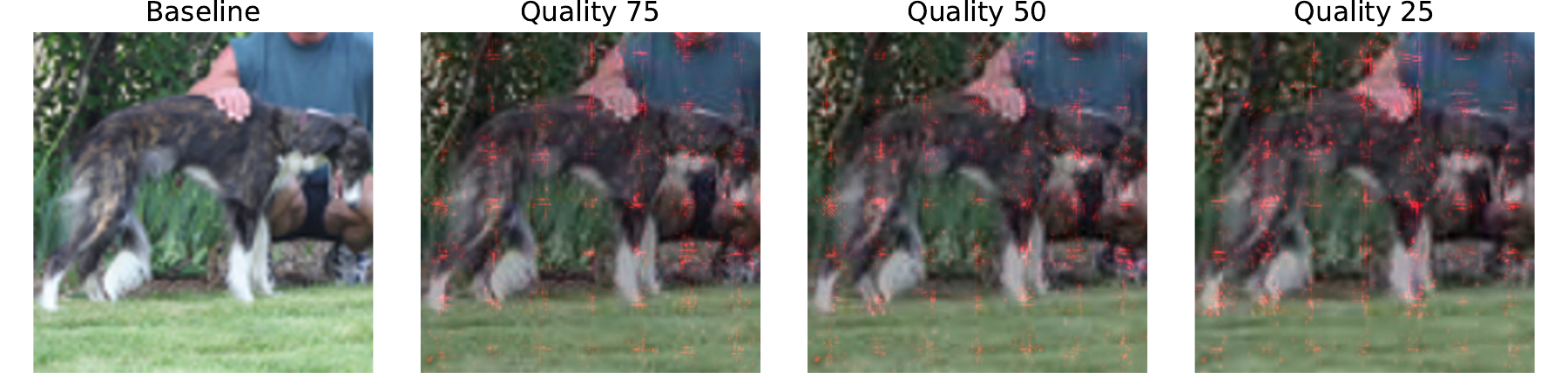}\\
			\includegraphics[width=\linewidth ]{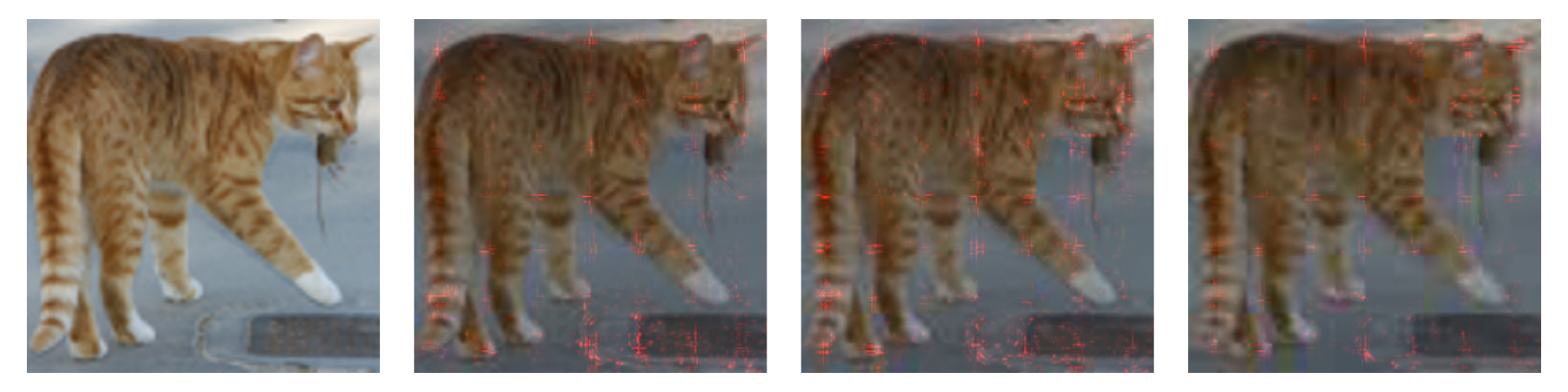}\\
			\includegraphics[width=\linewidth ]{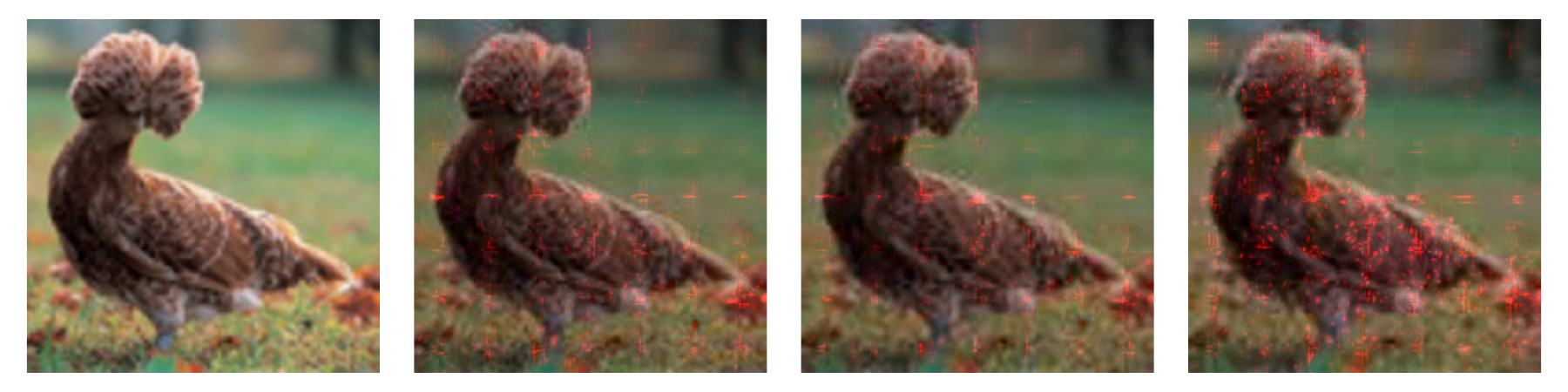}\\
			\includegraphics[width=\linewidth ]{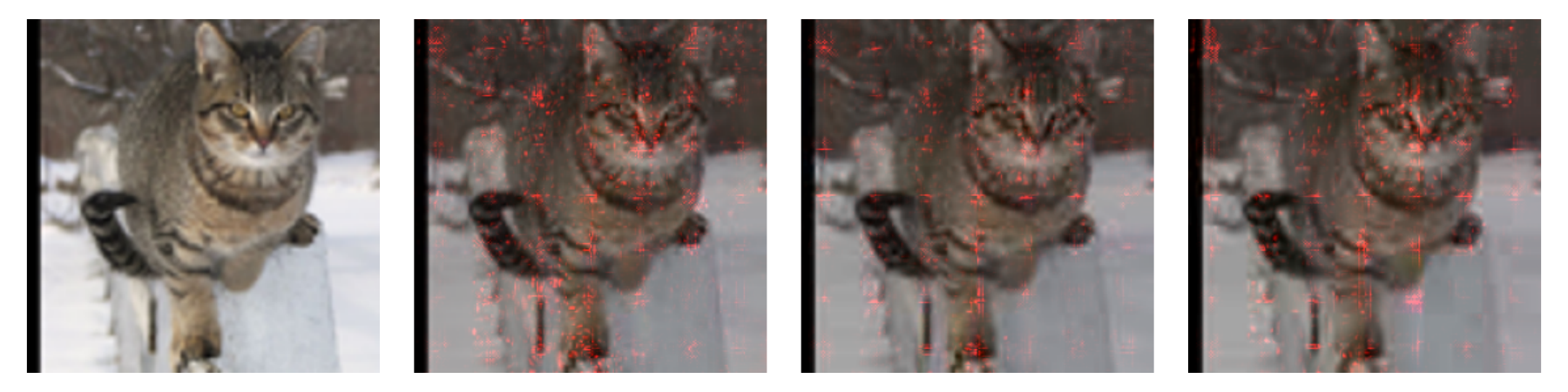}\\
			\includegraphics[width=\linewidth ]{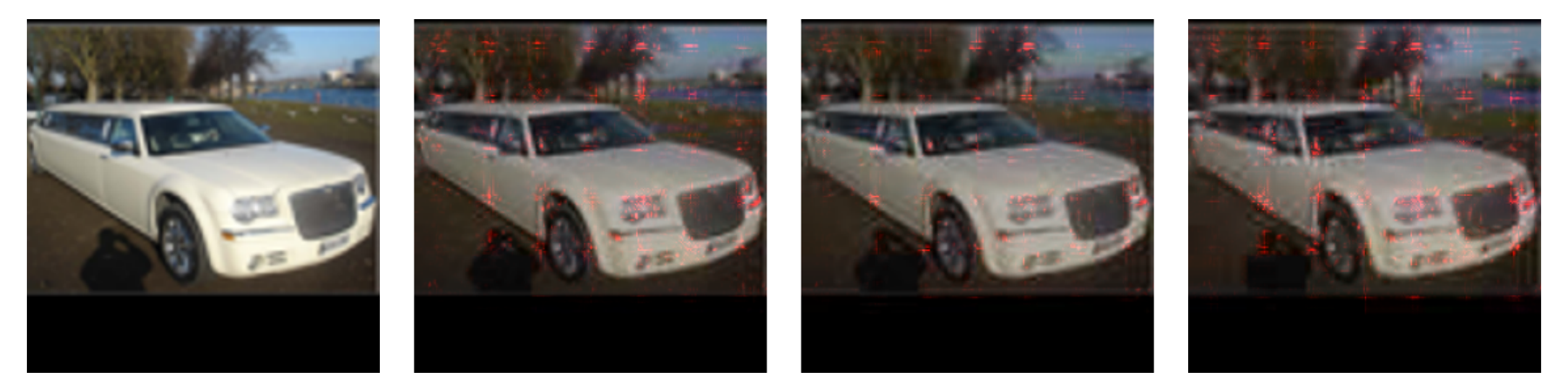}\\
			\includegraphics[width=\linewidth ]{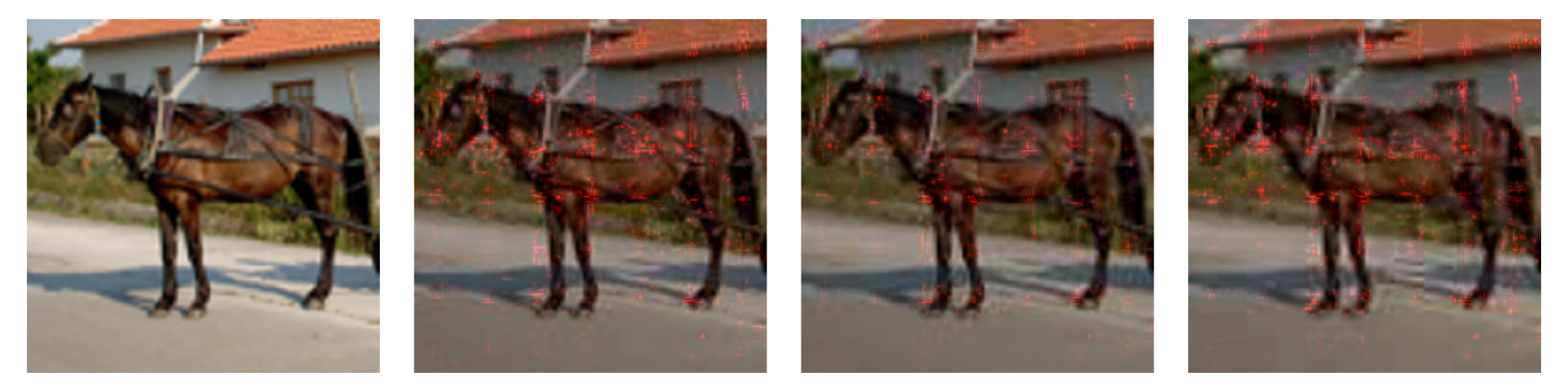}\\
			\includegraphics[width=\linewidth ]{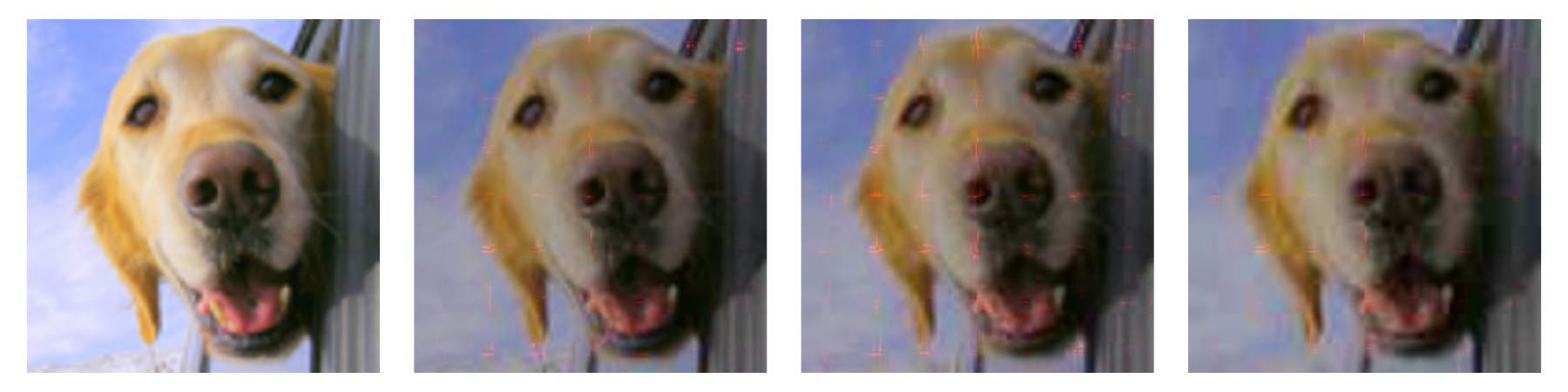}\\
			\includegraphics[width=\linewidth ]{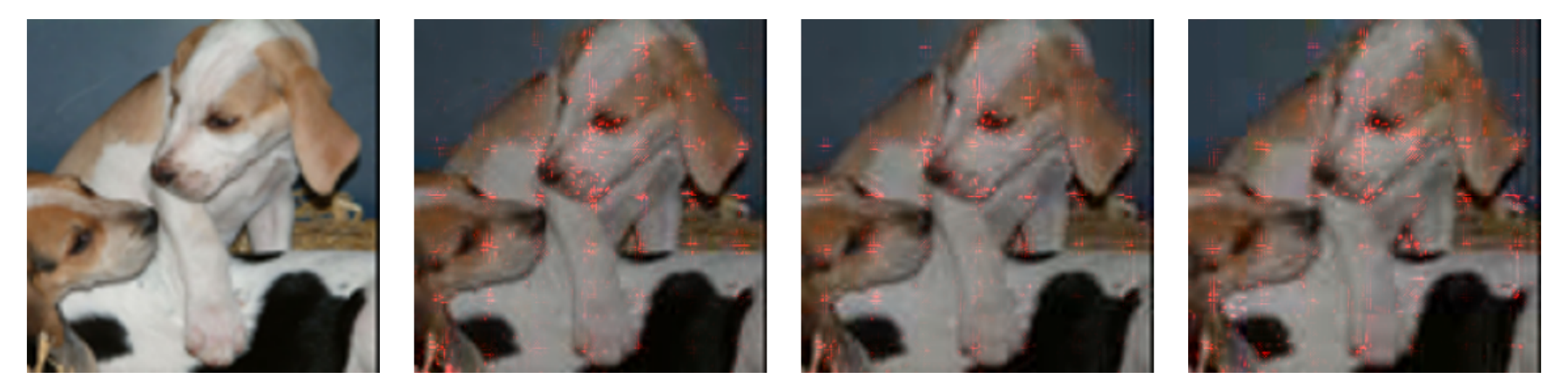}\\
			\includegraphics[width=\linewidth ]{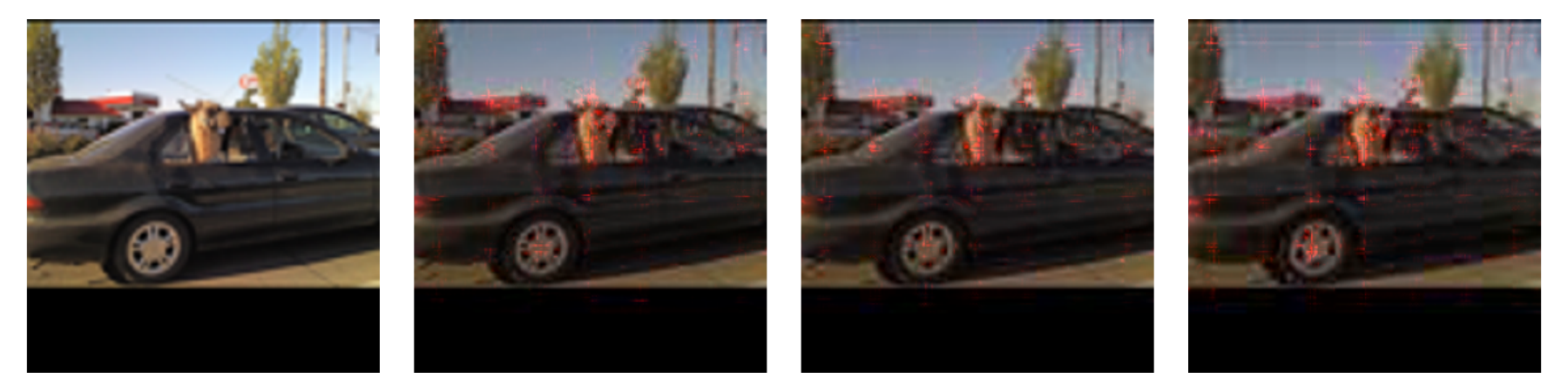}\\
			\includegraphics[width=\linewidth ]{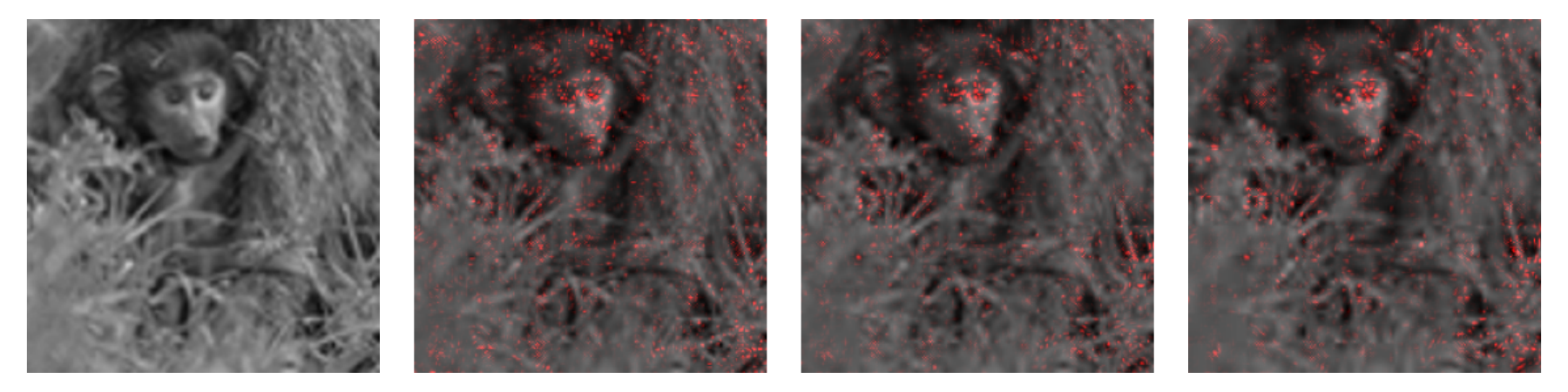}\\
			\caption{ViT negative}
		\end{subfigure}\hspace{1cm}
		\begin{subfigure}[h]{0.45\textwidth}
			\includegraphics[width=\linewidth ]{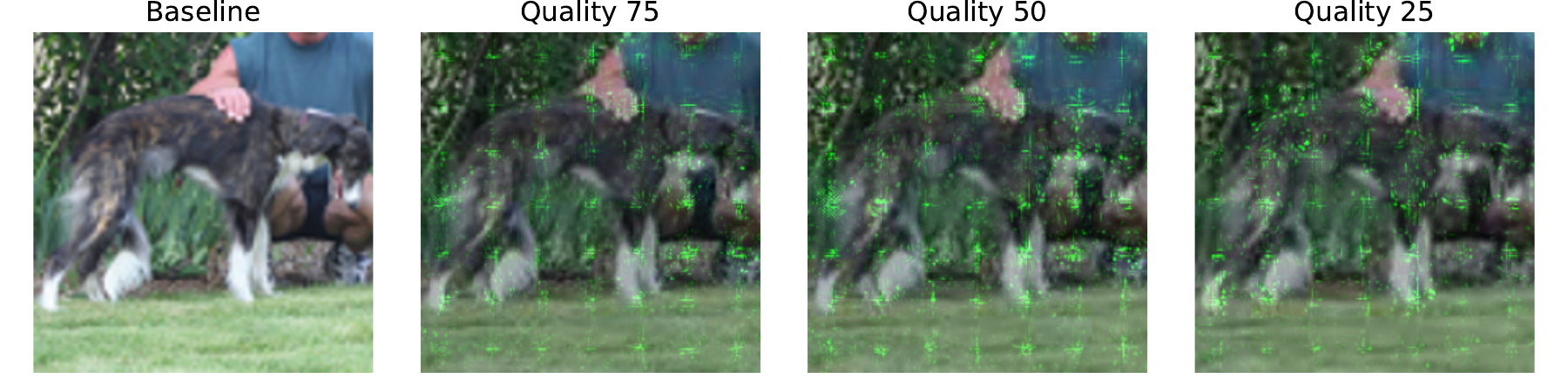}\\
			\includegraphics[width=\linewidth ]{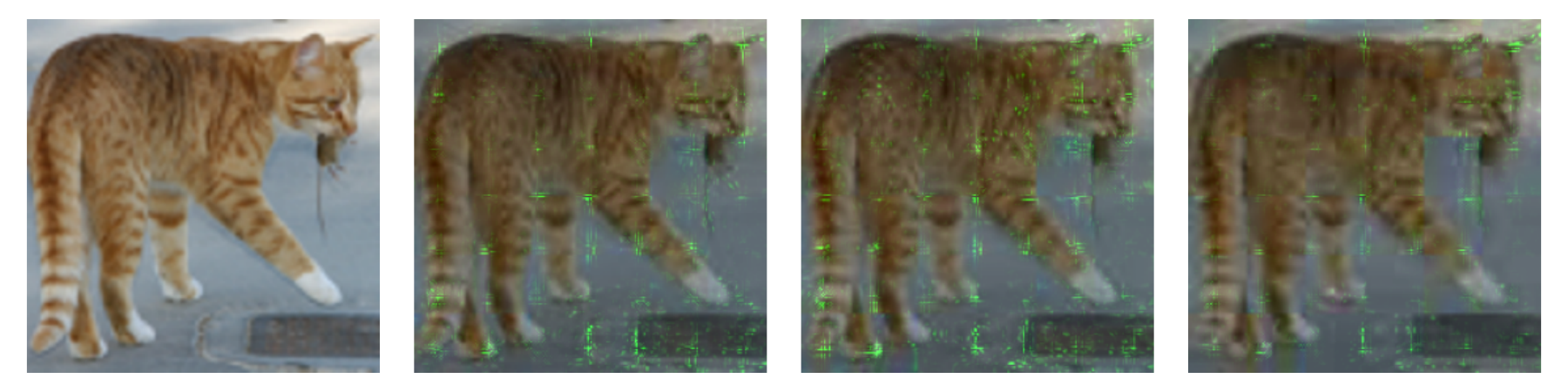}\\
			\includegraphics[width=\linewidth ]{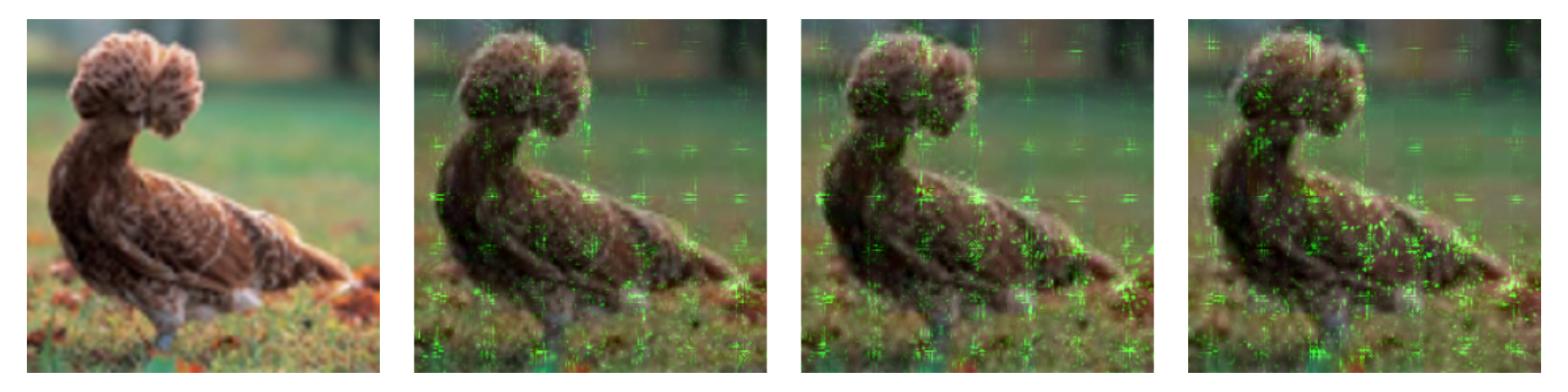}\\
			\includegraphics[width=\linewidth ]{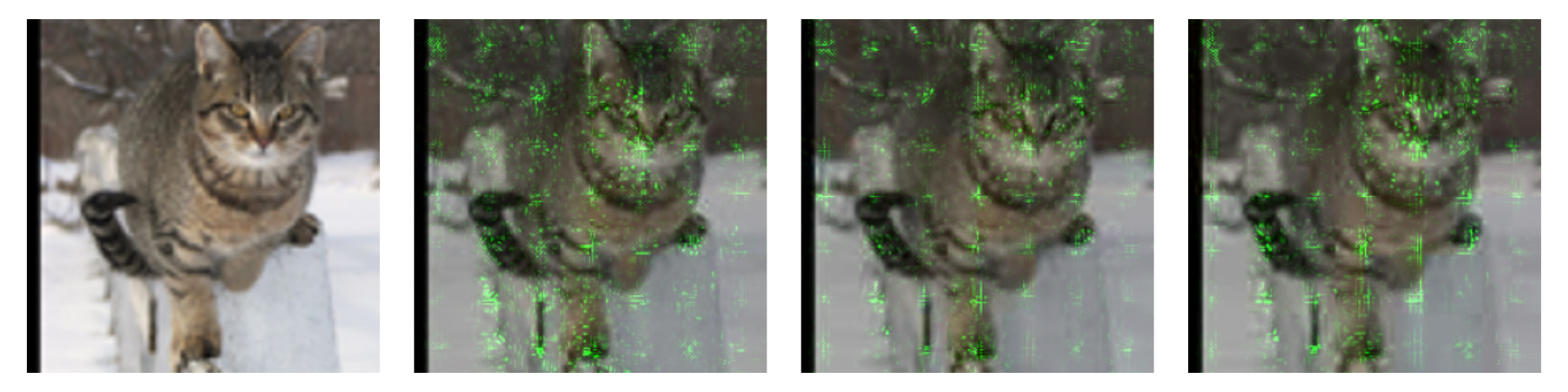}\\
			\includegraphics[width=\linewidth ]{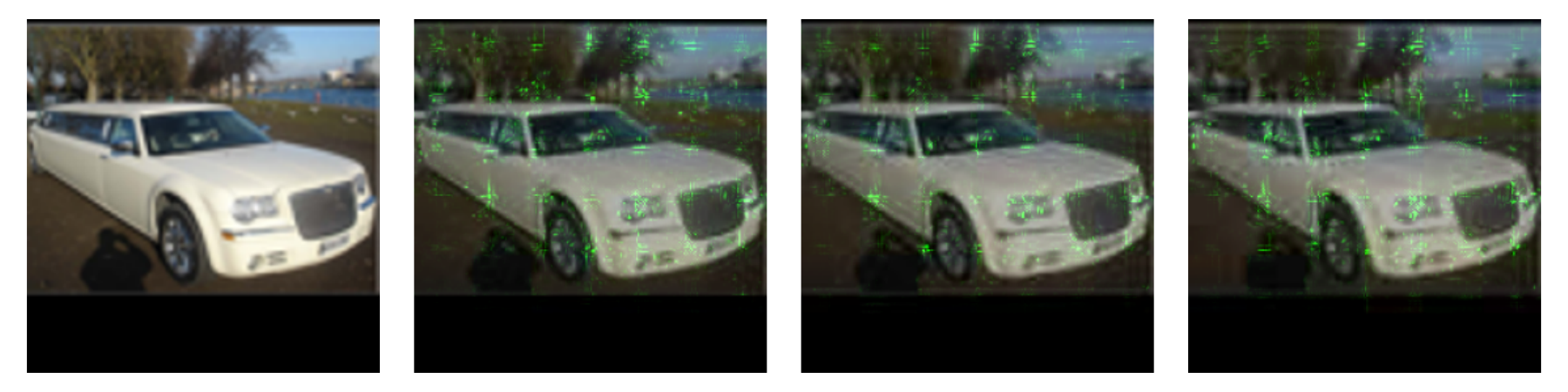}\\
			\includegraphics[width=\linewidth ]{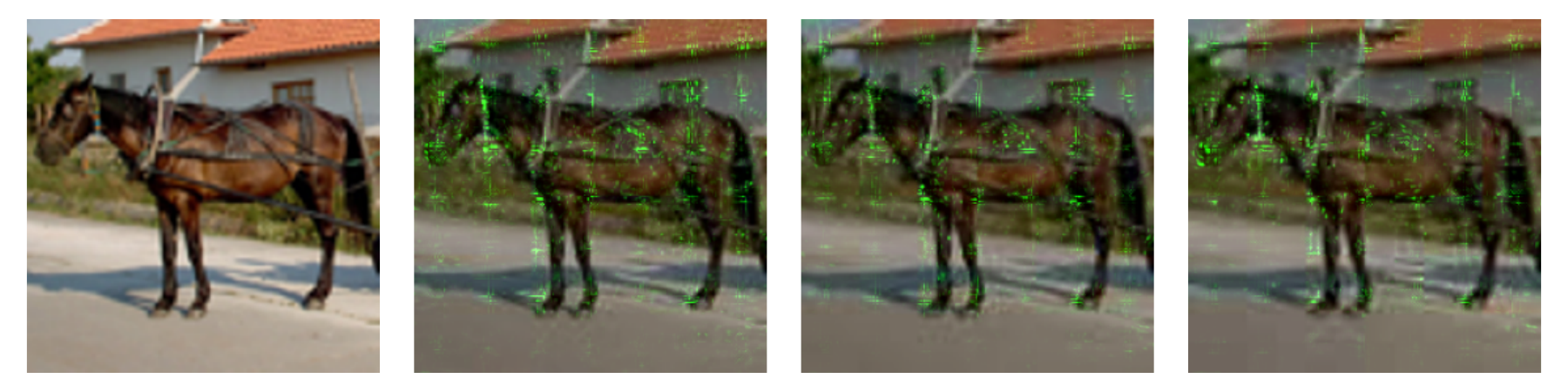}\\
			\includegraphics[width=\linewidth ]{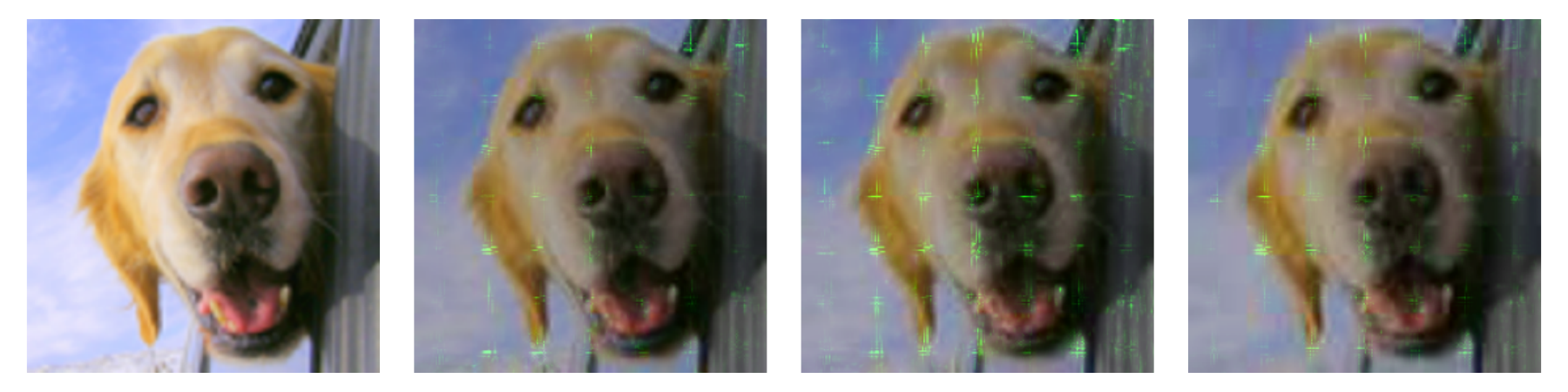}\\
			\includegraphics[width=\linewidth ]{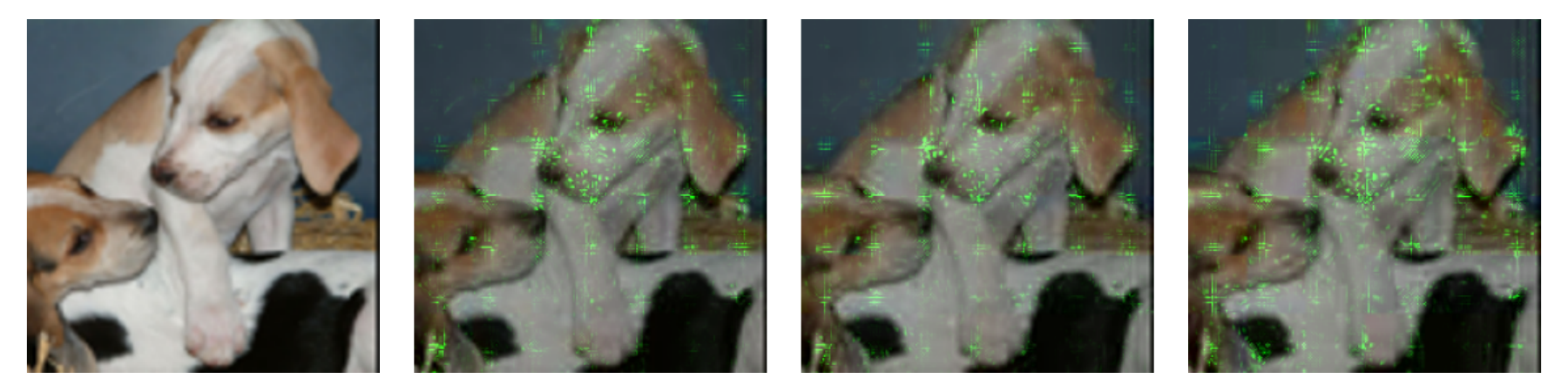}\\
			\includegraphics[width=\linewidth ]{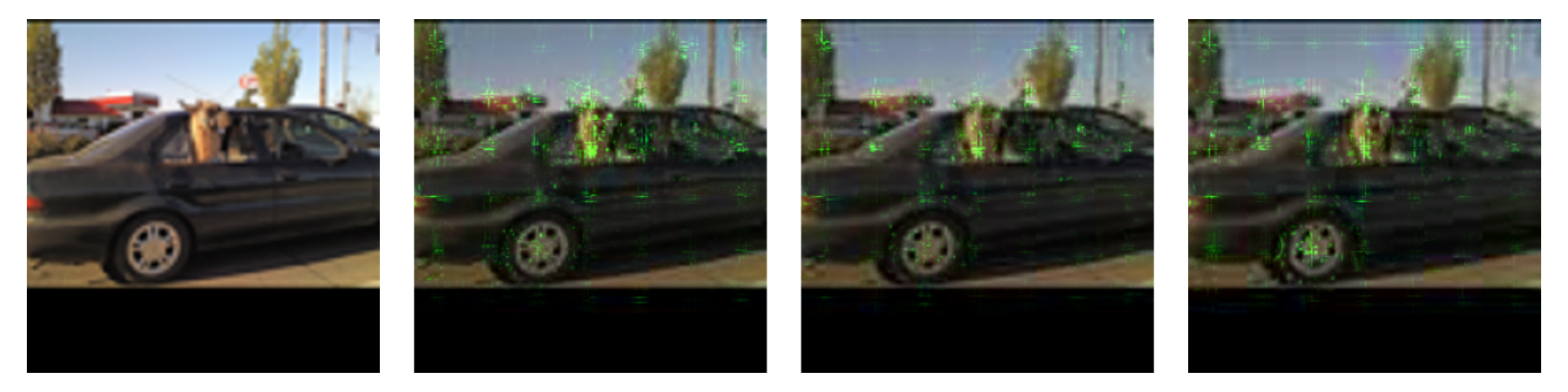}\\
			\includegraphics[width=\linewidth ]{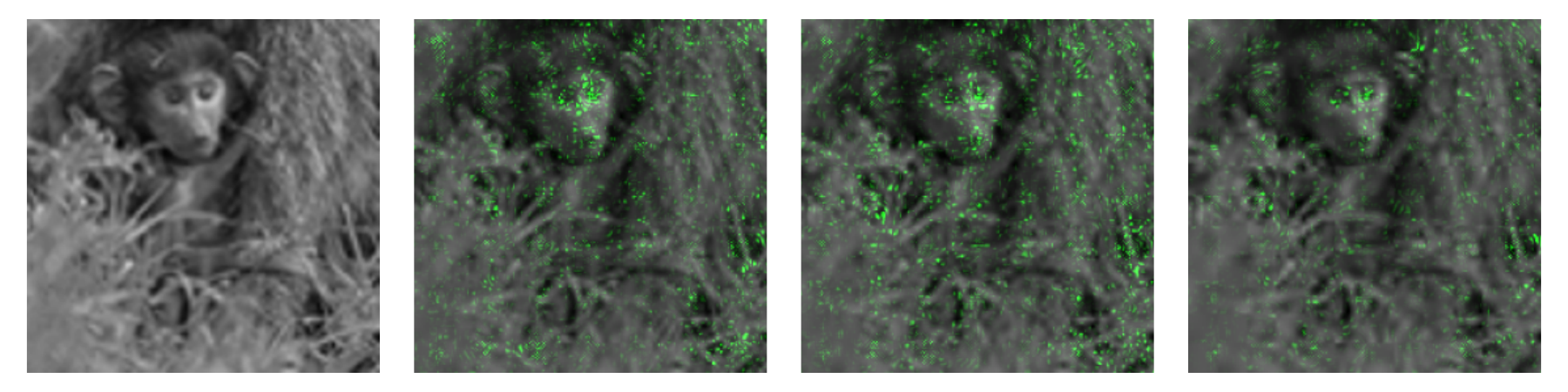}\\
			\caption{ViT positive}
		\end{subfigure} \\
		\caption{Examples on visualisation of integrated gradients for ViT-B/32.} \label{fig:ViT-B-32 supp stl-10}
\end{figure}
\begin{table*}[!ht]
		\centering
		\scalebox{0.85}{%
		\begin{tabular}{lccc}
		\toprule
		      True label & Predicted label & Predicted score & IG \\
		     \midrule
		     dog & cat, cat, monkey, monkey & 0.2313, 0.2036, 0.1284, 0.1312 & 0.0995, 0.4628, 0.5523 \\
             cat & cat, cat, cat, cat & 0.9843, 0.9695, 0.8641, 0.7563 & 0.0163, 0.1273, 0.2588 \\
             bird  & car, car, car, car & 0.0022, 0.0050, 0.0272, 0.0445 & -0.8398, -2.4107, -3.0264 \\
             cat & cat, cat, cat, cat & 0.9774, 0.9806, 0.8440, 0.8783 & -0.0035, 0.1523, 0.1018 \\
             car & bird, bird, bird, bird & 0.0012, 0.0004, 0.0003, 0.0006 & 1.0531, 1.4599, 0.7195 \\
             horse & monkey, monkey, monkey, monkey & 0.0000, 0.0001, 0.0002, 0.0017 & -0.8399, -1.3930, -4.0528 \\
             dog & dog, dog, dog, dog & 0.9456, 0.8902, 0.8897, 0.9540 & 0.0583, 0.0586, -0.0075 \\
             dog & dog, dog, dog, cat & 0.7744, 0.2991, 0.2231, 0.2275 & 0.9604, 1.2202, 1.2141 \\
             car & bird, bird, bird, bird & 0.0037, 0.0026, 0.0012, 0.0018 & 0.3116, 1.1126, 0.6517 \\
             monkey & cat, horse, dog, cat & 0.0374, 0.0150, 0.0405, 0.0550 & 0.9117, -0.0521, -0.3912 \\

		     \bottomrule
		\end{tabular}
		}
        \caption{Detailed information on the visual outputs in Figure \ref{fig:RN50 supp stl-10} and \ref{fig:Supp RN50 both stl-10}. }\label{tbl:Supp stl-10 RN50 comparisons}
\end{table*}
\begin{table*}[!ht]
		\centering
		\scalebox{0.85}{%
		\begin{tabular}{lccc}
		\toprule
		      True label & Predicted label & Predicted score & IG \\
		     \midrule
		     dog & dog, dog, dog, dog & 0.9568, 0.9236, 0.9114, 0.4917 & 0.0362, 0.0486, 0.6657 \\
             cat & cat, cat, cat, cat & 0.9737, 0.9691, 0.9508, 0.9525 & 0.0050, 0.0238, 0.0219 \\
             bird  & car, car, car, car & 0.0003, 0.0002, 0.0005, 0.0016 & 0.2784, -0.4080, -1.6100 \\
             cat & cat, cat, cat, cat & 0.9730, 0.9490, 0.9504, 0.9408 & 0.0256, 0.0238, 0.0336 \\
             car & bird, bird, bird, bird & 0.0009, 0.0020, 0.0017, 0.0013 & -0.7868, -0.6456, -0.3844 \\
             horse & monkey, monkey, monkey, monkey & 0.0000, 0.0000, 0.0000, 0.0000 & -0.4918, -0.3042, 1.2674 \\
             dog & dog, dog, dog, dog & 0.9581, 0.9655, 0.9656, 0.9810 & -0.0067, -0.0067, -0.0228 \\
             dog & dog, dog, dog, dog & 0.9404, 0.9317, 0.9191, 0.8953 & 0.0099, 0.0235, 0.0494 \\
             car & bird, bird, bird, bird & 0.0048, 0.0097, 0.0089, 0.0026 & -0.7108, -0.6097, 0.6313 \\
             monkey & car, deer, car, monkey & 0.0855, 0.0896, 0.1559, 0.1979 & -0.0584, -0.6076, -0.8505 \\
		     \bottomrule
		\end{tabular}
		}
        \caption{Detailed information on the visual outputs in Figure \ref{fig:ViT-B-32 supp stl-10} and \ref{fig:Supp ViT both stl-10}. }\label{tbl:Supp stl-10 ViT comparisons}
\end{table*}
\end{document}